\documentclass[11pt]{article}

\usepackage{microtype} 
\usepackage{booktabs}  
\usepackage{url}  

\usepackage{amsmath,amsfonts,bm}

\def\eqref#1{equation~\ref{#1}}

\def\floor#1{\lfloor #1 \rfloor}
\def\1{\bm{1}}

\DeclareMathAlphabet{\mathsfit}{\encodingdefault}{\sfdefault}{m}{sl}
\SetMathAlphabet{\mathsfit}{bold}{\encodingdefault}{\sfdefault}{bx}{n}

\newcommand{\E}{\mathbb{E}}

\DeclareMathOperator*{\argmax}{arg\,max}
\DeclareMathOperator*{\argmin}{arg\,min}

\usepackage{amsmath}
\usepackage{amsthm}
\usepackage{graphicx}
\usepackage{amsmath}
\usepackage{amssymb}
\usepackage{mathtools}
\usepackage{amsthm}
\usepackage{todonotes}
\usepackage{multirow}
\usepackage{nicefrac}
\theoremstyle{plain}
\newtheorem{theorem}{Theorem}[section]

\newtheorem{corollary}[theorem]{Corollary}
\theoremstyle{definition}

\theoremstyle{remark}

\usepackage[inline]{enumitem}
\usepackage{array}
\usepackage[table]{xcolor}
\usepackage[final]{automl}
\usepackage{wrapfig}
\usepackage[capitalize,noabbrev]{cleveref}

\usepackage{natbib}
\title{Dynamic Priors in Bayesian Optimization for Hyperparameter Optimization}

\author[1]{\nameemail{Lukas Fehring}{l.fehring@ai.uni-hannover.de}}
\author[1,2]{\nameemail{Marcel Wever}{m.wever@ai.uni-hannover.de}}
\author[1]{\nameemail{Maximilian Spliethöver}{m.spliethoever@ai.uni-hannover.de}}
\author[1]{\nameemail{Leona Hennig}{l.hennig@ai.uni-hannover.de}}
\author[1,2]{\nameemail{Henning Wachsmuth}{h.wachsmuth@ai.uni-hannover.de}}
\author[1,2]{\nameemail{Marius Lindauer}{m.lindauer@ai.uni-hannover.de}}

\affil[1]{Institute of Artificial Intelligence, Leibniz University Hannover}
\affil[2]{L3S Research Center}
\hypersetup{%
  pdfauthor={Lukas Fehring, Marcel Wever, Maximilian Splieth\"over, Leona Hennig, Henning Wachsmuth, Marius Lindauer},
  pdftitle={Dynamic Priors in Bayesian Optimization for Hyperparameter Optimization},
  pdfsubject={DynaBO is a Bayesian HPO method that lets users dynamically inject priors mid-run, safeguards against misleading ones, and achieves provable convergence plus SOTA results.},
  pdfkeywords={Bayesian optimization, hyperparameter optimization, AutoML, dynamic priors, user priors, prior rejection, human-in-the-loop, interactive machine learning}
}

\begin{document}

\maketitle


\newcommand{\cA}{\mathcal{A}}
\newcommand{\cL}{\mathcal{L}}
\newcommand{\cD}{\mathcal{D}}
\newcommand{\cX}{\mathcal{X}}
\newcommand{\cY}{\mathcal{Y}}

\newcommand{\perf}{f} 
\newcommand{\conf}{\lambda} 
\newcommand{\confs}{\Lambda} 
\newcommand{\inc}{\hat{\conf}} 
\newcommand{\confopt}{\conf^\ast} 
\newcommand{\af}{\alpha} 
\newcommand{\EI}{\af_{EI}}
\newcommand{\fun}{f}
\newcommand{\surrogate}{\hat{\fun}}
\newcommand{\prior}{\pi}

\newcommand{\tool}{{\texttt{DynaBO}}\xspace}
\newcommand{\pibo}{$\pi$BO\xspace}

\renewcommand\vec{\mathbf}
\newcommand{\yahpo}{YAHPO Gym\xspace}
\newcommand{\inctuple}[1]{(\conf_{#1}, \perf_{\conf_{#1}})}
\definecolor{mydarkblue}{rgb}{0.0,0.0,0.6} 
\newcommand{\change}[1]{\textcolor{mydarkblue}{#1}}

\begin{abstract}
    Bayesian optimization (BO) is a widely used approach to hyperparameter optimization (HPO). However, most existing HPO methods only incorporate expert knowledge during initialization, limiting practitioners' ability to influence the optimization process as new insights emerge. This limits the applicability of BO in iterative machine learning development workflows.
    We propose \tool, a BO framework that enables continuous user control of the optimization process. Over time, \tool leverages provided user priors by augmenting the acquisition function with decaying, prior-weighted preferences while preserving asymptotic convergence guarantees. To enhance robustness, we introduce a surrogate-model-based safeguard that detects and, possibly, rejects misleading priors. We prove theoretical results on near-certain convergence, robustness to deceptive priors, and accelerated convergence when informative priors are provided.
    Extensive experiments across various HPO benchmarks show that \tool consistently outperforms state-of-the-art competitors across all benchmarks and for all prior kinds. Our results demonstrate that \tool enables reliable and efficient collaborative BO, bridging automated and manually controlled model development.
\end{abstract}

\section{Introduction}
\label{sec:introduction}

Automated Hyperparameter optimization (HPO) is essential for the effective deployment of machine learning models. 
Its effectiveness was shown on classical machine learning \citep{eggensperger-neuripsdbt21a,bansal-neurips22a,pfisterer-automl22a} as well as on deep neural networks and transformers for computer-vision and natural language processing~\citep{muller-icml23a,wang-jmlr24, rakotoarison-icml24a, dblp-pineda-iclr24a}. Nevertheless, experts often prefer manual over automated HPO~\citep{bouthillier-hal20a,vanderbloom-automlws21a, kannengiesser-acm25a}. 

To strengthen the acceptance of automated HPO, in view of a collaborative, human-centered approach~\citep{lindauer-icml24a}, two main lines of research exist:
(i) explainability methods for HPO provide users with insights into the optimization process; (ii)  including experts in the optimization process, addressing the perceived lack of control~\citep{kannengiesser-acm25a}. While explainability has made significant progress in recent years~\citep{wang-chi19a,zoller-tiis23a,segel-automl23a,segel-jmlr25a, wever-aaai26a}, control over HPO remains quite limited, and is focused on initialization.

Specifically for effectively interacting with HPO, \citet{hvarfner-iclr22a} and \citet{mallik-neurips23a} propose novel user-centric interfaces for Bayesian optimization (BO) \citep{jones-jgo98a} and Hyperband \citep{li-iclr17a}, respectively. By explicitly specifying initial user priors on the location of an optimal configuration, these interfaces enable users to bias the optimization toward desired regions. 
\citet{dblp-hvarfner-iclr24a} extend this idea by building a general framework that enables users to encode additional target-function properties, such as achievable performance, into the optimization process.
These approaches show that informative priors improve performance all the while misleading ones do not break the optimization process. Additionally, any associated performance deterioration can be recovered from. While advancing human-centered HPO, these methods rely exclusively on static prior user input and lack the online control required for dynamic, interactive adjustments.

\begin{figure}[t]
    \centering
    \begin{minipage}[c]{0.5\linewidth}
        \centering
        \includegraphics[width=\linewidth,trim={0cm 0cm 0cm 0.1cm},clip]{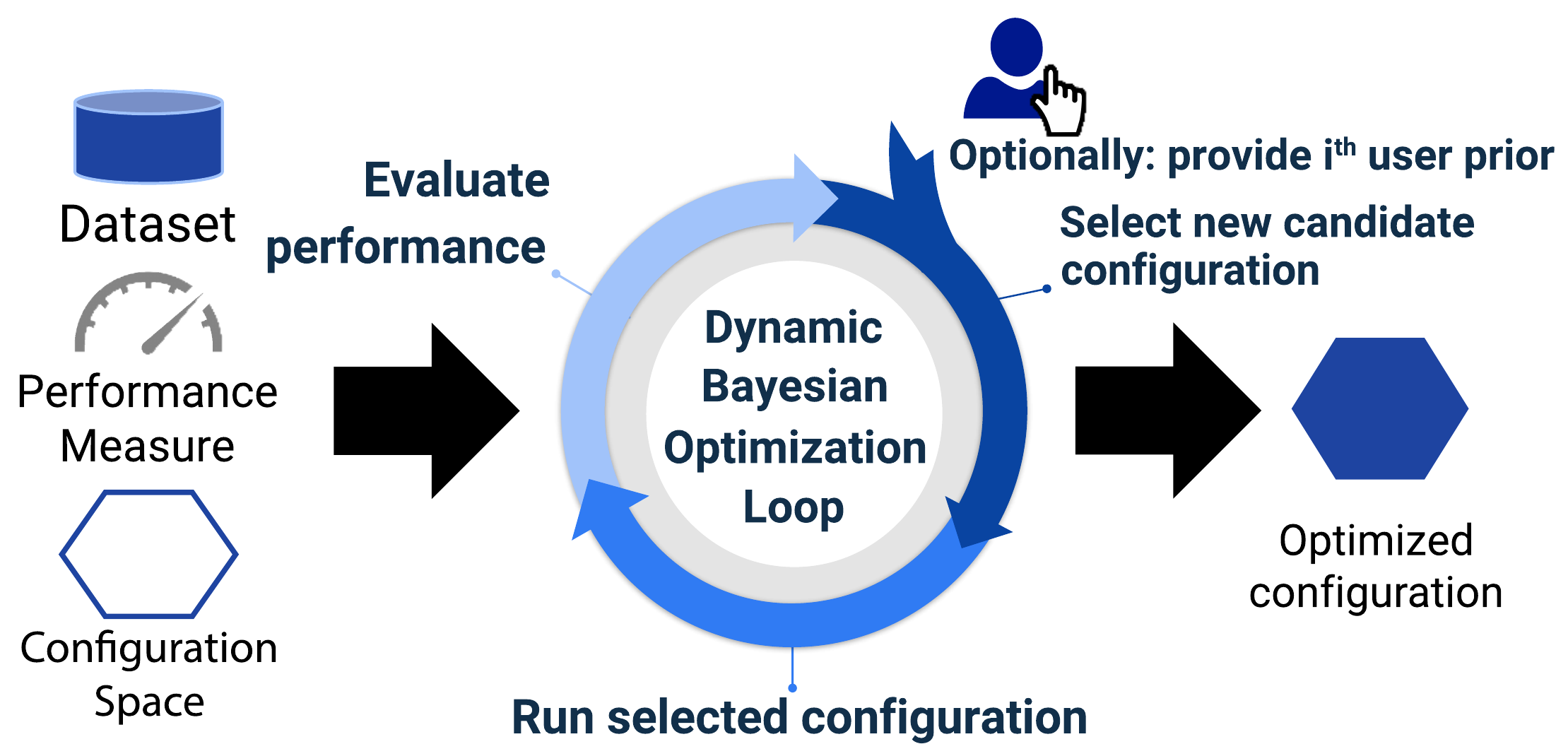}
    \end{minipage}%
    \begin{minipage}[c]{0.45\linewidth}
        \centering
        \vspace{0.55cm}
        \includegraphics[width=\linewidth,trim={.2cm 0cm 0cm 0cm},clip]{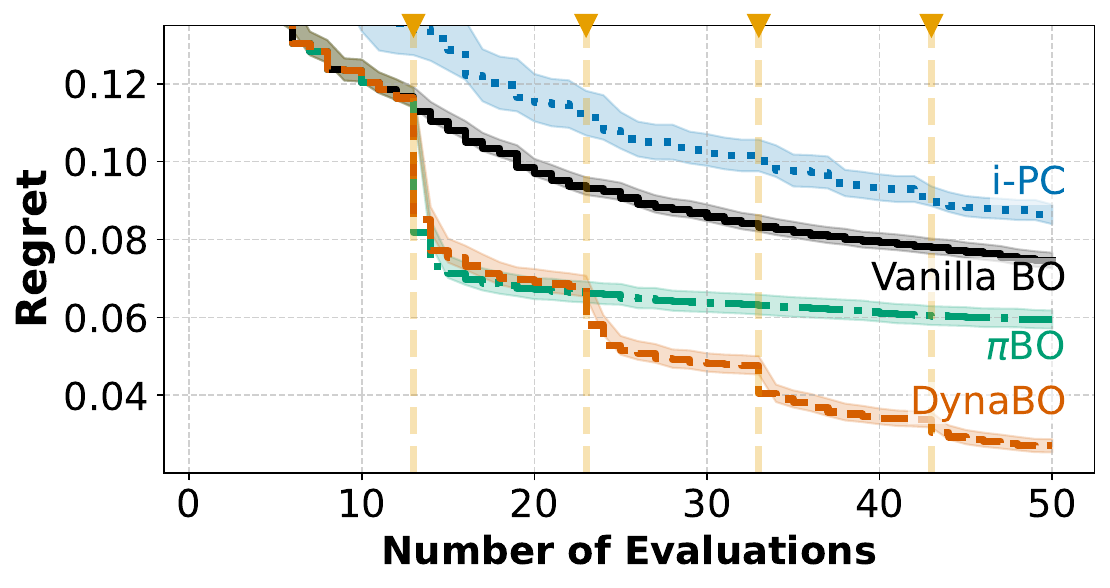}
    \end{minipage}
    \caption{Left: Overview of the proposed dynamic Bayesian optimization \tool, which allows steering the optimization process by dynamically adding priors at runtime. Right: Comparison of \tool with Vanilla BO, $\pi$BO, and interactive Probabilistic Circuits (i-PC) on the PD1 benchmarks with \texttt{Expert} priors provided at vertical lines and standard error uncertainties.}
    \label{fig:intro-figure}
\end{figure}

We focus on settings where users repeatedly want to inject prior knowledge into BO. To that end, we propose a new state-of-the-art HPO approach, dubbed \tool, and visualized in \cref{fig:intro-figure}. \tool (i) allows users to continuously steer optimization, and (ii) enables them to use HPO in the typical rapid-prototyping workflows of machine learning practitioners~\citep{studer-mdp21a} where users oversee the construction of the continuously-updated model. 
In contrast to \citet{seng-automlconf25a}, who operate in a similar setup, we focus on combining the surrogate-based acquisition function with the prior, effectively synergizing the user's prior and the surrogate model's belief. \tool allows priors on all hyperparameters simultaneously, and
when supplied with priors in close succession, prior effects are combined. Concretely, \tool generalizes the work of \citet{hvarfner-iclr22a} from a single to \textit{multiple} priors, retaining \pibo's speedup for informative priors. To address the remaining performance deterioration caused by misleading priors, we propose a data-driven detection mechanism with minimal overhead. Lastly, beyond maintaining \pibo's convergence guarantees, we prove acceleration for informative priors even for finite-time settings.\\
\textbf{Contributions.}\,\,\, We make the following contributions:
\begin{enumerate*}
    \item[(i)] We propose \tool, the first BO framework that supports continuous user steering through dynamic priors. \tool further includes a safeguard to detect and, if necessary, reject misleading user priors.
    \item[(ii)] We establish convergence, robustness, and acceleration guarantees.
    \item[(iii)] We demonstrate strong empirical performance across diverse HPO benchmarks, outperforming \pibo and probabilistic circuits over all prior kinds.
\end{enumerate*}

\section{Hyperparameter Optimization}\label{sec:hyperparameter_optimization}

Hyperparameter optimization (HPO) aims to find a suitable learner and task-dependent hyperparameter configuration $\conf \in \confs$ \citep{bischl-dmkd23a} 
with the objective of minimizing the cost 
\[
\confopt \in {\arg\min}_{\conf \in \confs} \, f(\conf).
\]
While HPO can be tackled via grid or random search \citep{bergstra-jmlr12a}, sophisticated methods, such as Bayesian optimization, are often more efficient and effective \citep{turner-neuripscomp21a}.

\paragraph{Bayesian Optimization (BO)} BO~\citep{mockus-ottc75a,jones-jgo98a,shahriari-ieee16a} is a model-based sequential optimization technique widely-used for sample-efficient HPO~\citep{snoek-nips12a, falkner-icml18a,  cowenrivers-jair22a, makarova-automlconf22a, bischl-dmkd23a}. BO is particularly well-suited for optimizing black-box functions $\fun$, which are costly to evaluate and have neither a closed-form solution nor gradient information available.

The BO process begins with an initial design that aims to provide diverse coverage of the hyperparameter configuration space. It then proceeds, alternating between two main steps:
\begin{enumerate*}
    \item[(i)] fitting a probabilistic surrogate model $\surrogate$ to the $t$ observations, and
    \item[(ii)] selecting the next configuration.
\end{enumerate*}

New configurations $\lambda_{t+1}$ are selected such that they maximize the acquisition function $\af_{\surrogate}:\Lambda\rightarrow\mathbb{R}$ that quantifies the utility of candidates by balancing configuration space exploration and exploitation of well-performing regions. Formally, this means:
\begin{equation}
    \conf^{t+1} \in \argmax_{\conf \in \confs} \alpha_{\surrogate}(\conf) \label{eq:acq_opt}
\end{equation}

A prominent acquisition function example is Expected Improvement (EI) \citep{jones-jgo98a}, which selects points expected to yield improvements over the best configuration observed so far, also called the incumbent. 
Formally, let $\inc$ denote the current incumbent of the black-box function, then EI is defined as
\[
\af^{EI}_{\surrogate}(\conf) := \E \left[ \max \left( \fun(\inc) - \surrogate(\conf), 0 \right) \right] ,
\]
where $\surrogate(\conf)$ is treated as a random variable representing the probability distribution at $\conf$ as modeled by $\surrogate$.
In HPO, the black-box function $\fun$ typically corresponds to the empirical generalization error, estimating the expected loss, for example, via hold-out validation or cross-validation.

\section{Related Work}

Approaches related to \tool can be categorized into three groups based on their usage of
\begin{enumerate*}[label=(\arabic*)]
    \item data-driven priors,
    \item explicit user-generated priors, and
    \item interactive user steering mechanisms.
\end{enumerate*}

\textbf{Data-Driven Priors}\,\,\,
A substantial body of research leverages prior experience to configure the HPO process. This includes transfer learning across tasks \citep{swersky-nips13a,feurer-aaai15a,rijn-kdd18a,feurer-arxiv22a}, prior extraction and transfer from low to high fidelity landscapes \citep{li-aaai26a}, configuration space design \citep{perrone-neurips19a}, 
and surrogate model configuration \citep{feurer-automl18c}. While effective, such approaches do not enable direct user interaction. Instead, they require users to trust the automated knowledge transfer mechanisms. 

\textbf{Learning from User Priors}\,\,\,
Another line of research explicitly incorporates user-specified beliefs.
\citet{bergstra-nips11a} introduce fixed priors over the configuration space, whereas \citet{souza-ecml21a} estimate posterior-driven models, though both approaches are limited by their acquisition function compatibility. \citet{ramachandran-kbs20a} warp the configuration space to emphasize promising regions, but this requires invertible priors and struggles with misleading ones. More recently, \citet{hvarfner-iclr22a} propose \pibo, which utilizes user beliefs on the position of the target function optimum to augment the acquisition functions. Similarly, \citet{mallik-neurips23a} extend Hyperband \citep{li-iclr17a} to balance sampling of configurations in a user-defined region with randomly sampled ones. However, none of these methods considers continuous control.

\textbf{Interactive Hyperparameter Optimization}\,\,\,
Beyond initialization, recent work integrates users more directly into the optimization process. \citet{xu-arxiv24b} allow users to reject candidate evaluations, while \citet{adachi-aistats24} let users choose among proposed alternatives.
\citet{seng-automlconf25a} propose an alternative to the common BO framework based on probabilistic circuits. They focus on optimizing with the prior as evidence; i.e., priors are not combined with the surrogate model's belief. Complementarily, \citet{chang-arxiv25a} introduce LLINBO working with LLM-generated candidate configurations to augment BO, with rejection schemes for robustness. Likewise, cooperative design optimization \citep{niwa-arxiv25a} explores natural language interfaces where LLMs propose configurations, optionally guided by users. Their user studies indicate that such interaction mitigates overtuning to local optima, while maintaining user agency.

In contrast to learning from data-driven or user-provided priors, our proposed method \tool enables users to provide priors at any time during the optimization process, while remaining compatible with standard BO and model- and acquisition-function-agnostic. Unlike previous methods for interactive HPO, \tool not only natively synergizes the acquisition function and prior to jointly select configurations but also provides safeguards to detect and possibly reject misleading priors, thereby advancing toward a collaborative HPO paradigm.

\section{Dynamic Priors in Bayesian Optimization}\label{sec:dynamic-priors}

In this section, we show how \tool leverages at runtime-provided (or dynamic) priors in BO. To this end, we first generalize prior-weighted acquisition functions from a single prior provided at initialization to multiple dynamic priors (\cref{sec:prior-weighted-af}), and we then devise a method to detect misleading priors and safeguard against them (\cref{sec:rejecting-priors}).

\subsection{Prior-Weighted Acquisition Function}\label{sec:prior-weighted-af}

Following \citet{hvarfner-iclr22a}, we integrate user-provided prior information on the location of the optimum by weighting the original acquisition function with a prior distribution.
Unlike their approach, however, we dynamically incorporate this external knowledge during optimization.

\newcommand{\priortuple}[1]{(t^{(#1)}, \prior^{(#1)})}

Given an acquisition function $\af$, and a user-specified prior $\prior: \confs \rightarrow \mathbb{R}^+$, \citet{hvarfner-iclr22a} select the next point to be evaluated with respect to $\fun$ at time $t$ as follows:
\[
\af^\text{\pibo}_{\surrogate}(\conf) := \af_{\surrogate}(\conf) \cdot \prior(\conf)^{\beta/t}\, ,
\]
where $\beta \in \mathbb{R}^+$ is a scaling hyperparameter. For $t\rightarrow \infty$, the weight induced by $\prior$ converges to 1 independent of the configuration $\conf$ and $\beta$, that is, the effect of the prior diminishes over time.

\begin{figure}[t]
    \centering
    \begin{subfigure}{0.5\linewidth}
        \centering
        \includegraphics[width=\linewidth,trim={0cm 0.6cm 0cm 0cm},clip]{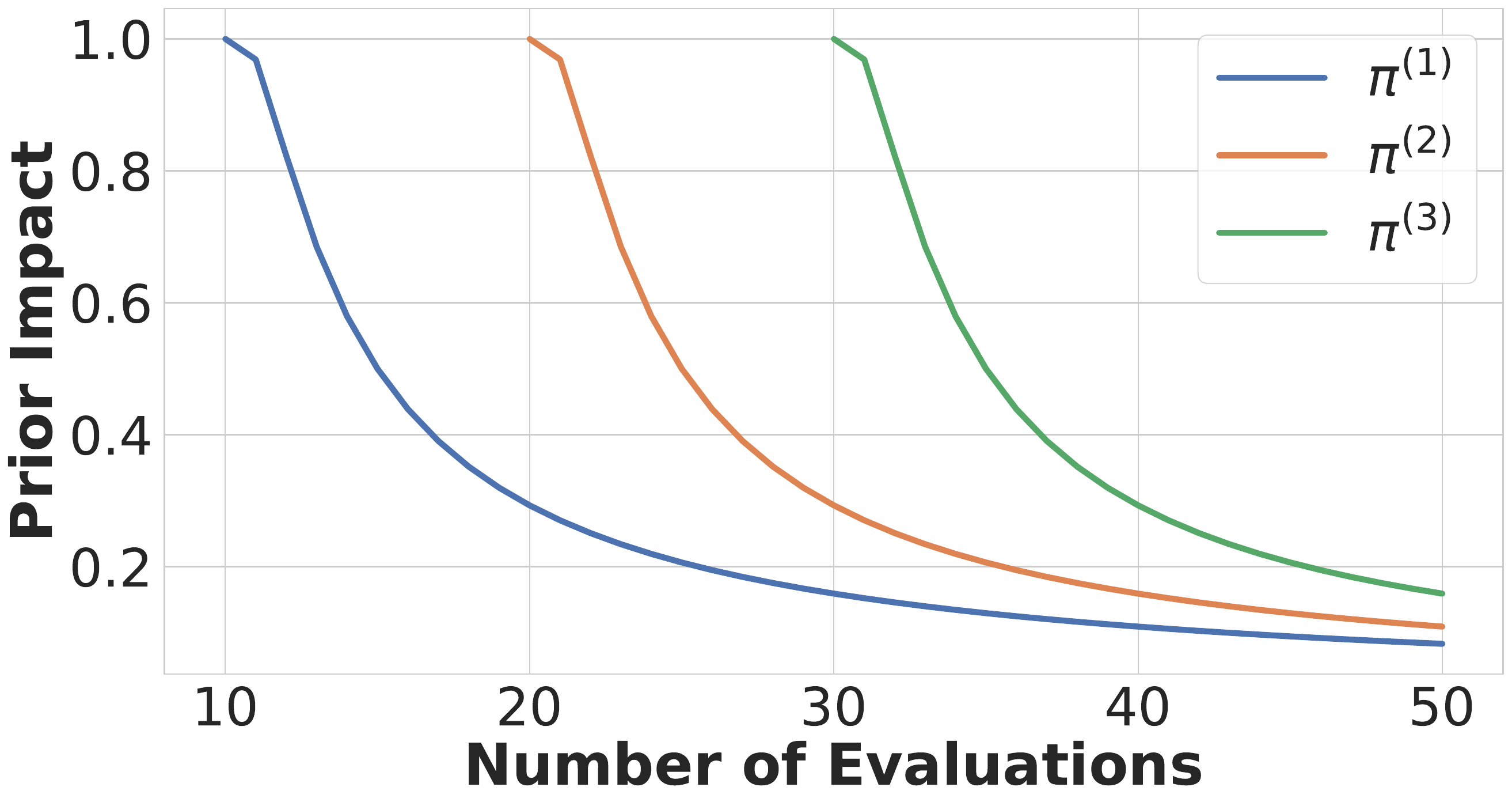}
        \caption{Acquisition function impact of priors $\pi^1,\pi^2,\pi^3$, provided at $t=10, 20, \text{and } 30$. We plot the prior impact as $1-\pi(\conf)^{\beta/t}$ for  $\pi(\lambda)=0.5$.}
        \label{fig:prior-impact}
    \end{subfigure}%
    \begin{subfigure}{0.5\linewidth}
        \centering
        \includegraphics[width=\linewidth,trim={0.35cm 19cm 2.5cm 0.8cm},clip]{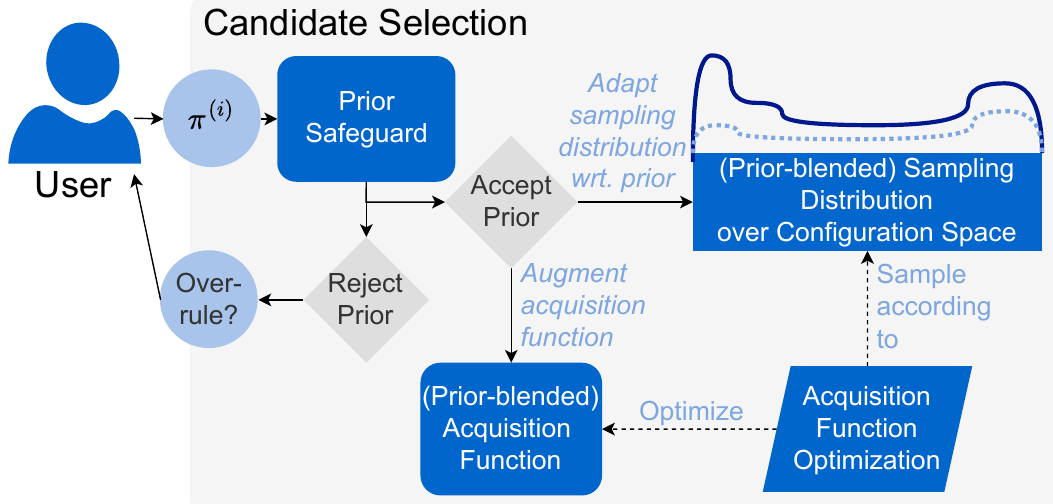}
        \caption{Illustration of the candidate selection process in \tool. A safeguard mechanism evaluates provided priors, determining their acceptance. 
        }
        \label{fig:example-prior-af}
    \end{subfigure}
    \caption{Visualization of prior decay (left) and candidate selection process (right).}
\end{figure}

In contrast, we suppose a finite sequence of user-specified priors $\{\pi^{(m)}\}_{m=1}^{M}$ provided at times $\{t^{(m)}\}_{m=1}^{M}$, $t^{(1)}<\ldots<t^{(M)}\leq T$. We define a dynamically-adapted acquisition function $\alpha_\text{dyna}:\Lambda\rightarrow\mathbb{R}$ by multiplying the sum of the priors (for a comparison with multiplying the product of priors, refer to \cref{app:sum-vs-product}):
\[
\af^\text{dyna}_{\surrogate}(\conf) := \af_{\surrogate}(\conf)\, \cdot \sum_{m=1}^M \prior^{(m)}(\conf)^{\beta/(t-t^{(m)})}\,
\]
By stacking priors, we can dynamically incorporate information provided at different time steps. Priors are faded individually based on their age; that is, older priors are considered less important, as shown in \cref{fig:prior-impact}.
The flexibility in incorporating multiple priors sets \tool apart from \pibo \citep{hvarfner-iclr22a} and Priorband \citep{mallik-neurips23a}, which consider a single initial prior, as well as \citet{seng-automlconf25a}'s approach, which considers only a single prior at a time. 

To optimize the acquisition function, we build on \citet{hutter-lion11a}'s combined local and random search, which first samples a set of candidate starting configurations and then performs hill climbing. We extend this approach by including configurations drawn from the provided prior distribution as starting points. This ensures that regions favored by the prior are explored explicitly during acquisition function optimization. Further details are deferred to \cref{app:facilitating_prior_behavior}.

\subsection{Detecting and  Rejecting Priors}
\label{sec:rejecting-priors}

To safeguard against misleading priors, which are known to slow down optimization processes \citep{hvarfner-iclr22a, mallik-neurips23a, seng-automlconf25a}, we propose a mechanism to detect and, if necessary, reject poor priors. To this end, we utilize the surrogate model $\surrogate$ and an acquisition function $\xi:\confs \to \mathbb{R}$. $\xi$ can but need not be the same as $\alpha$. For an overview of how the safeguard is embedded in \tool's candidate selection routine, we refer to \cref{fig:example-prior-af}.

Specifically, we assess the promise of a prior $\prior^{(m)}$ by comparing the potential of the suggested region with that of the region around the best found configuration, the current incumbent $\inc$. 

The potential of regions is assessed by sampling a number of configurations according to the provided prior $\prior^{(m)}$, as well as a normal distribution around the current incumbent $\inc$. These distributions are denoted by $\mathcal{F}_{\prior^{(m)}}$ and $\mathcal{N}_{\inc}$, respectively. We then compare  both samples based on  $\xi$ and $\surrogate$. We accept the prior if and only if the quality of the prior exceeds the quality of the incumbent area by a given threshold $\tau$:
\begin{equation}\label{eq:threshold}
\E_{\conf \sim \mathcal{F}_{\prior^{(m)}}} \left[ \xi_{\surrogate}(\lambda) \right] - \E_{\conf \sim \mathcal{N}_{\inc}}  \left[ \xi_{\surrogate}(\conf) \right] \geq \tau \,\, .
\end{equation}
Higher values of $\tau$ reject more priors, filtering out misleading ones at the cost of discarding helpful ones; lower values embrace user priors more readily but admit misleading ones more easily.  

In practice, $\tau$ should be chosen with respect to the acquisition function $\xi$, e.g., for Lower Confidence Bound (LCB) or EI \citep{snoek-nips12a} in the space of loss values. Since we want to reject priors based on their potential, we recommend utilizing LCB and setting $\tau$ with respect to the user's beliefs on the remaining optimization potential $f({\inc}) - f({\confopt})$. 
We further suggest using the prior rejection to warn against priors, thus allowing users to overrule the mechanism, which further fosters a human-in-the-loop optimization workflow. For implementation details, a sensitivity analysis of $\tau$, refer to \cref{sec:results:reject}, and \cref{app:further-results:rejection_ablation}, respectively.

\section{Theoretical Analysis}\label{sec:theoretical-analysis}

In the theoretical analysis of \tool, we establish convergence and robustness properties under multiple dynamically provided user priors, and quantify the conditions under which informative priors yield accelerated convergence. These results are stated for the unsafeguarded algorithm; since the safeguard of \cref{sec:rejecting-priors} only filters out priors before they enter the acquisition function, the guarantees carry over directly. Extending the convergence results of \pibo \citep{hvarfner-iclr22a}, we demonstrate that our approach retains the almost sure convergence behavior of vanilla BO in the limit even under misleading priors. At the same time, our method can effectively leverage informative priors to accelerate convergence. Note that we assume a \textit{finite} prior set, and the utilization of UCB as an acquisition function. All proofs are provided in \cref{app:proofs}. We follow standard convergence results for BO \citep{srinivas-it12a} augmented by dynamic prior influence.

\subsection{Almost Sure Convergence of \tool}
We analyze the asymptotic behavior of \tool given assumptions on objective function regularity, finiteness, and vanishing influence of priors, introduced formally in Appendix~\ref{app:proofs}. This allows priors to be selected dynamically and to vary in quality.

\begin{theorem}[Almost Sure Convergence of \tool]\label{thm-dynbo-convergence}
Under the assumptions in \cref{app:proofs}, let $\lambda^* \in \arg\max_{\lambda \in \Lambda} f(\lambda)$ be a global maximizer of $f$. The sequence of query points selected by {\upshape \tool}, $\{\lambda_t\}_{t=1}^{\infty}\subset\Lambda$, satisfies almost sure convergence to a global optimum; that is, irrespective of the variation in priors, the method converges to $\lambda^*$ with probability one:
\begin{equation*}
    \lim_{T\to\infty} \min_{t \le T} \bigl(f(\lambda^*) - f(\lambda_t)\bigr) = 0 \qquad \text{a.s.}
\end{equation*}
\end{theorem}

\subsection{Robustness to Misleading Priors}
Asymptotically, the algorithm does not suffer degradation in performance even when faced with misleading priors. Formally, the best objective value found after $t$ iterations is, in the limit, no worse than vanilla BO's. This is a safeguard against noise or misguidance.

\begin{corollary}[Robustness to Misleading Priors]\label{corr:misleading priors}
Let $s_T^\mathrm{Dyna}$ and $s_T^\mathrm{UCB}$ denote the simple regret of \tool and vanilla GP-UCB, respectively. Then the asymptotic simple regret of \tool matches that of vanilla GP-UCB, even when priors are misleading:
\begin{equation*}
\lim_{T\to\infty}(s_T^\mathrm{Dyna} - s_T^\mathrm{UCB})=0 \quad\mathrm{a.s.}
\end{equation*}
That is, the influence of misleading priors vanishes in the limit.
\end{corollary}

\subsection{Accelerated Convergence with Informative Priors}
While \tool is robust to misleading priors, it is also designed to leverage informative information to accelerate optimization. To demonstrate that user-provided priors can accelerate the optimization process, we first formally characterize the adapted search behavior in \cref{theorem:adapted_search}. From this, we conclude that when informative priors are provided, \tool exhibits accelerated optimization behavior without compromising almost sure convergence or robustness to misleading priors.


\begin{theorem}[Adaptation of the Optimization Behavior through Priors]\label{theorem:adapted_search}
Let Assumptions A1--A3 in Appendix~\ref{app:proofs} hold. Suppose there exists a set $U_\epsilon \subset \Lambda$ of diameter 
$\epsilon > 0$ and $\delta \in (0,1)$ such that the prior-induced weighting $P(\lambda, t)$ 
concentrates \tool's sampling:
\begin{equation}
    P(\lambda_t \in U_\epsilon) \geq 1 - \delta, \quad \forall t \in \{1, \ldots, T\}.
\end{equation}
Then, the expected cumulative regret $\mathbb{E}[R_T]$ of \tool after \(T\) iterations satisfies, with high probability,
\begin{equation}
\mathbb{E}[R_T] \leq C\sqrt{T \beta_T^{\mathrm{UCB}}\, \gamma_T(U_\epsilon)} + \delta T B,
\end{equation}
with $C > 0$ from \citet{srinivas-it12a}, $B \coloneqq \sup_{\lambda \in \Lambda}(f(\lambda^*) - f(\lambda))$, 
confidence parameter $\beta_T^{\mathrm{UCB}}$, and information gain $\gamma_T(U_\epsilon) < \gamma_T(\Lambda)$ 
restricted to $U_\epsilon$.
\end{theorem}

\paragraph{Acceleration through Adapted Search Behavior}
Theorem \ref{theorem:adapted_search} formalizes the intuition that any prior reduces the search domain from $\Lambda$ to some $U_\epsilon$. This is always the case, especially of interest, however, if the prior emphasizes a local or, ideally, global optimum, hence \tool guiding the search towards improved configurations. The concentration condition $P(\lambda_t \in U_\epsilon) \geq 1 - \delta$ is a statement about this effect of the prior on \tool's search behavior.

Consider the special case of an informative prior, where $\lambda^* \in U_\epsilon$.
A prior that focuses sampling on any sufficiently small region around an optimum $U_\epsilon$ yields acceleration in the finite-horizon setting.
The algorithm is efficiently resolving $f$ near $\lambda^*$, which translates directly into faster identification of well-performing configurations.
Points within $U_\epsilon$ are more correlated under the kernel, so the posterior variance decreases more rapidly over $U_\epsilon$ than over the full domain. 
Nevertheless, Theorem~\ref{theorem:adapted_search} holds independent of whether $U_\epsilon$ contains $\lambda^*$, since the information gain $\gamma_T(U_\epsilon)$ over a smaller region is strictly less than $\gamma_T(\Lambda)$. However, if $U_\epsilon$ does not contain $\lambda^*$, the convergence to $\lambda^*$ relies on the prior decay, as guaranteed asymptotically by Theorem~\ref{thm-dynbo-convergence}. 

Our almost-sure convergence as well as asymptotic robustness~\ref{corr:misleading priors} claims show \tool's quality of recovering from uninformative or misleading priors. The sufficiency of asymptotic recovery in prior-guided BO has been established by \cite{hvarfner-iclr22a}.

\newpage
\section{Empirical Evaluation}\label{sec:empirical-evaluation}

In this section, we present an empirical evaluation of \tool, analyzing its anytime performance across various black-box benchmark scenarios and for different qualities of priors.
Concretely, we describe how priors are constructed (\cref{sec:prior-construct}), and our experimental setup (\cref{sec:experiment-setup}). We then discuss the results of a comparison with the PCs~\citep{seng-automlconf25a} and \pibo~\citep{hvarfner-iclr22a} baselines (\cref{sec:results:circuits} and \cref{sec:results:pibo}).
Additional experiments can be found in \cref{app:further-results}.

\subsection{Prior Construction: Expert, Advanced, Local, Deceptive}\label{sec:prior-construct}
Inspired by the evaluation protocols of \citet{souza-ecml21a,hvarfner-iclr22a, mallik-neurips23a}, and \citet{seng-automlconf25a}, we construct artificial, data-driven priors. \texttt{Expert} and \texttt{Advanced} priors bias \tool towards significantly better-performing regions, while \texttt{Local} priors favor well-performing regions similar to the current incumbent configuration. We hypothesize that such local priors are more similar to human behavior. \texttt{Deceptive} priors, in contrast, deliberately sample from poorly performing regions to simulate a worst-case scenario where a user provides misguided priors. In line with \citep{hvarfner-iclr22a}, we set $\beta$ to $N/10$ in our experimental evaluation, with $N$ as the total number of trials. Details on prior construction are discussed in \cref{app:priors}.

\newlength{\figheight}
\setlength{\figheight}{0.19\textwidth}
\subsection{Experimental Setup}\label{sec:experiment-setup}
In our experiments, we explore different configurations of \tool. We focus on random forests as surrogate models, and EI as the optimization acquisition function (theory-aligned results with LCB and GPs are provided in \cref{app:further-results:lcb_results} and \cref{app:further-results:gp-normal}). For prior rejection, we utilize LCB with $\tau = -0.15$ (the corresponding sensitivity analysis is provided in \cref{sec:results:reject}).

As baselines, we consider \citep{seng-automlconf25a}'s probabilistic circuits%
\footnote{A comparison to PCs required minor adaptations of their methodology and source code. All adaptations were discussed with \citet{seng-automlconf25a}. More information is provided in \cref{app:detailed_experimental_setup:competitors}\label{fn:pc}}
for interactive hyperparameter optimization, both with distribution priors and pointwise priors on the prior center.
Additionally, we consider the state-of-the-art approach \pibo as well as vanilla Bayesian optimization (vanilla BO), as implemented in the SMAC3 library \citep{lindauer-jmlr22a}. \pibo allows the user to provide prior information before the optimization process, whereas vanilla BO utilizes no user guidance. Given the same seed, \tool and \pibo will sample identical (first) priors.   
For each model and dataset, we run $30$ different seeds and plot the mean regret of the best-found incumbent, wrt.~the preceding exploration detailed in \cref{app:priors}. Shaded areas indicate the standard error. 
Further details can be found in \cref{app:detailed_experimental_setup}.


\subsection{Comparison to Probabilistic Circuits}\label{sec:results:circuits}

\begin{wrapfigure}{r}{0.5\textwidth}  
  \centering
  \vspace{-1.7\intextsep}  
  \begin{tabular}{@{}cc@{}}
    \hspace{.6cm}\texttt{Informative} & \texttt{Deceptive}\\ 
    \includegraphics[height=\figheight,trim={0.35cm 0.35cm 0.35cm 0.35cm},clip]{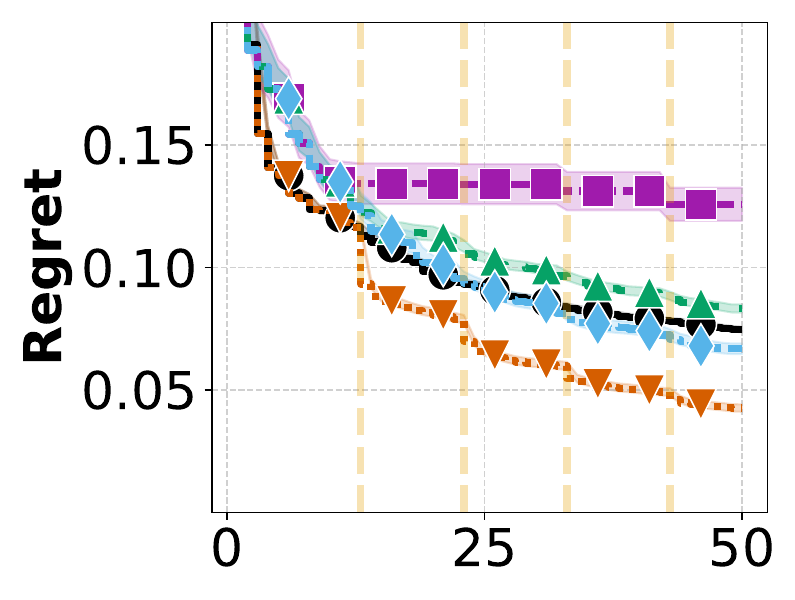} &
    \includegraphics[height=\figheight,trim={3.35cm 0.35cm 0.35cm 0.35cm},clip]{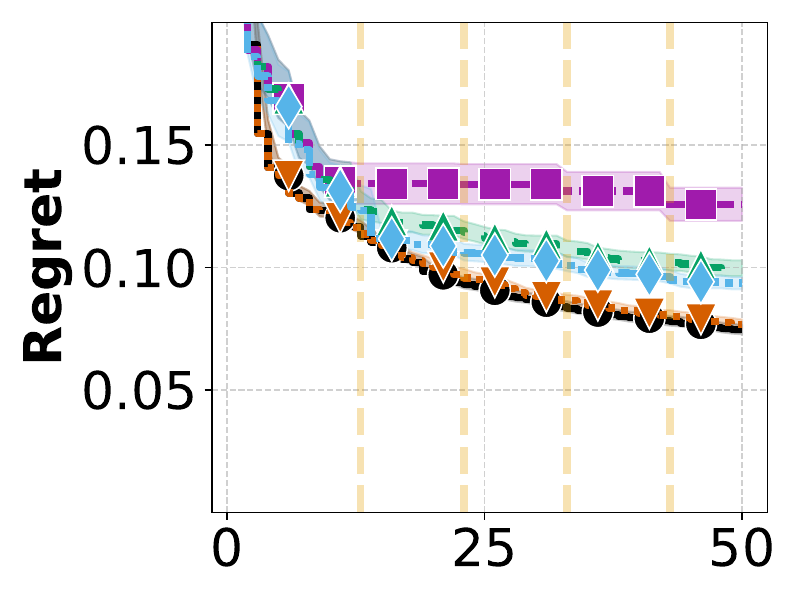} 
  \end{tabular}
  \begin{minipage}{\linewidth}
    \centering
    \includegraphics[width=\textwidth,trim={0cm 0cm 0cm 10cm},clip]{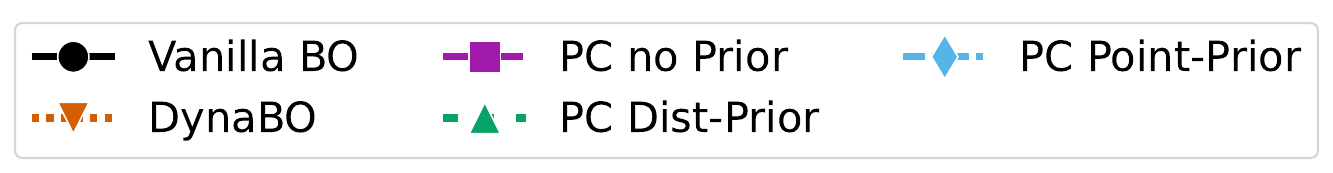}
  \end{minipage}
  \caption{Comparison of \tool and PC variants, with vanilla BO as reference. \texttt{Informative} priors include \texttt{Expert}, \texttt{Advanced}, and \texttt{Local} Priors.}
  \label{fig:pc_comparison}
  \vspace{-0.386cm}
\end{wrapfigure}

A comparison of the optimization behavior of \tool, and~\citet{seng-automlconf25a}'s probabilistic circuits (PCs) without priors, with distribution priors, and with pointwise priors is shown in \cref{fig:pc_comparison}\footref{fn:pc}. We note that the point-wise priors provide an advantage for PCs, since there is no uncertainty on well-performing configurations.
Nevertheless, \tool outperforms all PCs across all prior kinds and scenarios, motivating us to consider \texttt{Expert}, \texttt{Advanced}, and \texttt{Local} grouped as \texttt{Informative} priors. While these priors benefit PCs, they only outperform \texttt{vanilla-BO} if provided with pointwise priors.

\noindent
Interestingly, in the deceptive setting, PCs appear to benefit from priors but are not competitive with vanilla-BO or \tool. A detailed discussion of PCs' optimization performance is provided in \cref{app:further-results:pc}. We conjecture that \tool's superior performance stems from combining priors with the surrogate model, rather than using them as evidence. We therefore focus on the stronger competitors, vanilla BO and \pibo.

\subsection{Comparison to Vanilla BO and \pibo}\label{sec:results:pibo}

\begin{figure*}[t]
  \centering
  \setlength{\figheight}{0.132\textwidth}
  \begin{tabular}{@{}c@{}c@{}c@{}c@{}c@{}c@{}c@{}}
    & \hspace{.6cm}\texttt{widernet} & \texttt{resnet} & \texttt{transf} & \texttt{xformer} & \texttt{lcbench} & \texttt{xgboost}\\ 
    \rotatebox{90}{\hspace{.5cm}\texttt{Expert}} &
    \includegraphics[height=\figheight,trim={0.35cm 1.45cm 0cm 0.35cm},clip]{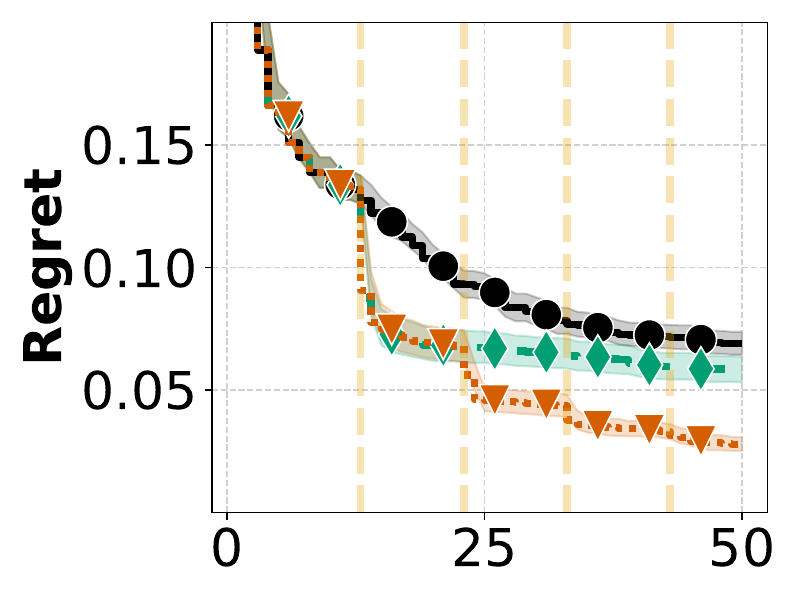} &
    \includegraphics[height=\figheight,trim={3.55cm 1.45cm 0cm 0.35cm},clip]{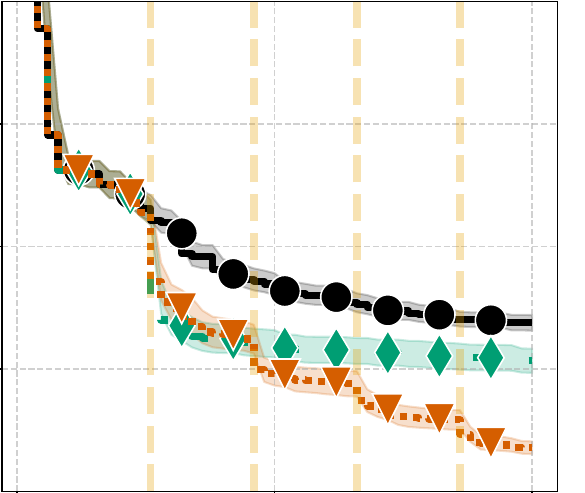} &
    \includegraphics[height=\figheight,trim={3.55cm 1.45cm 0cm 0.35cm},clip]{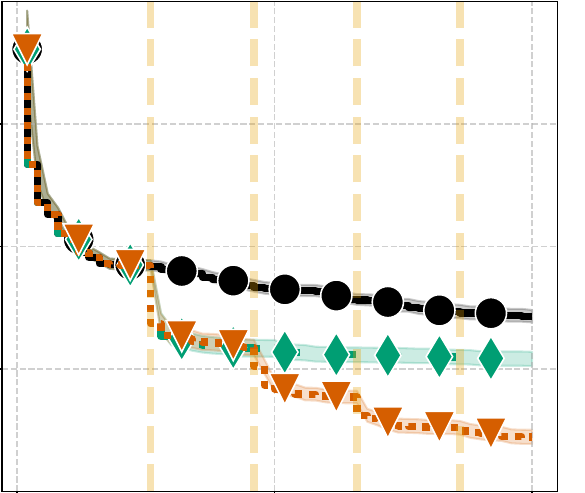} &
    \includegraphics[height=\figheight,trim={3.55cm 1.45cm 0cm 0.35cm},clip]{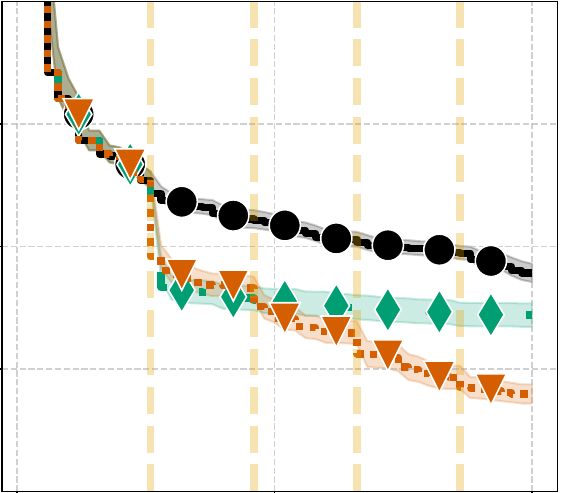} &
    \includegraphics[height=\figheight,trim={3.55cm 1.45cm 0cm 0.35cm},clip]{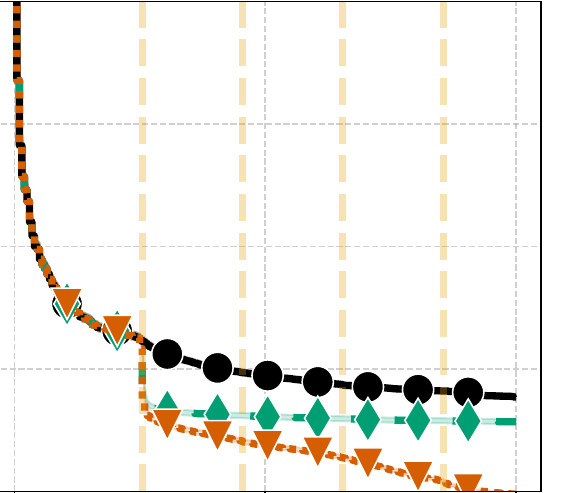} &
    \includegraphics[height=\figheight,trim={3.55cm 1.45cm 0cm 0.35cm},clip]{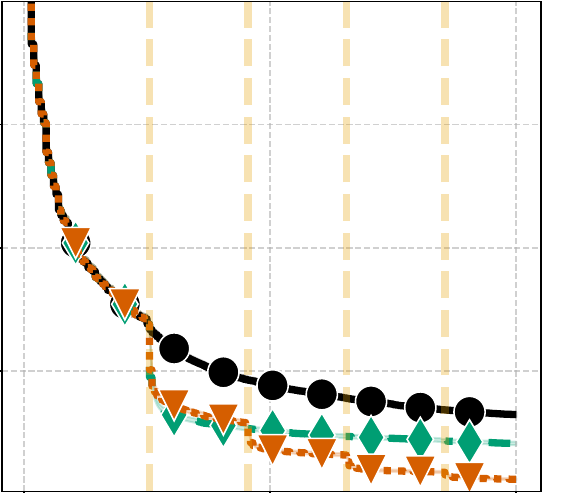}\\

    \rotatebox{90}{\hspace{.3cm}\texttt{Advanced}} &
    \includegraphics[height=\figheight,trim={0.35cm 1.45cm 0cm 0.35cm},clip]{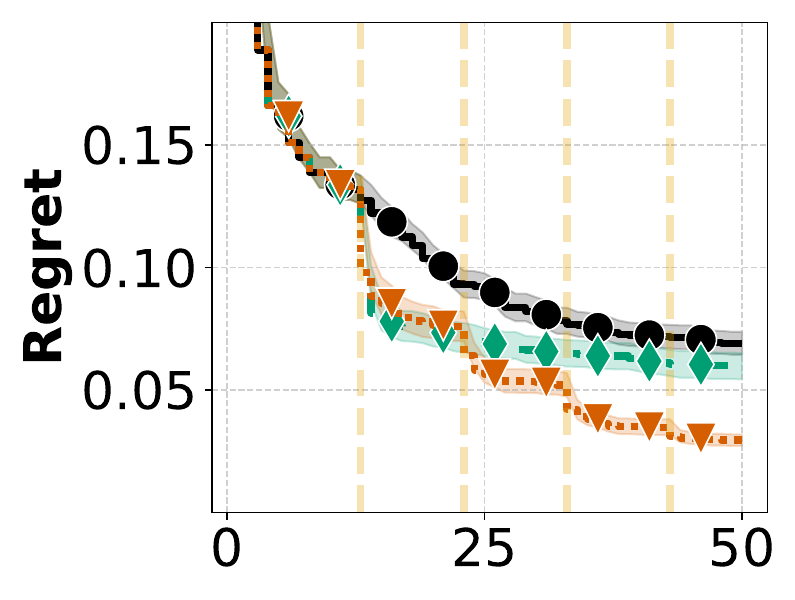} &
    \includegraphics[height=\figheight,trim={3.55cm 1.45cm 0cm 0.35cm},clip]{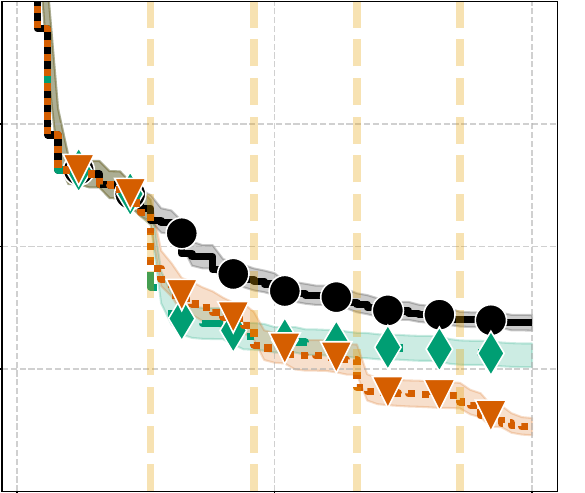} &
    \includegraphics[height=\figheight,trim={3.55cm 1.45cm 0cm 0.35cm},clip]{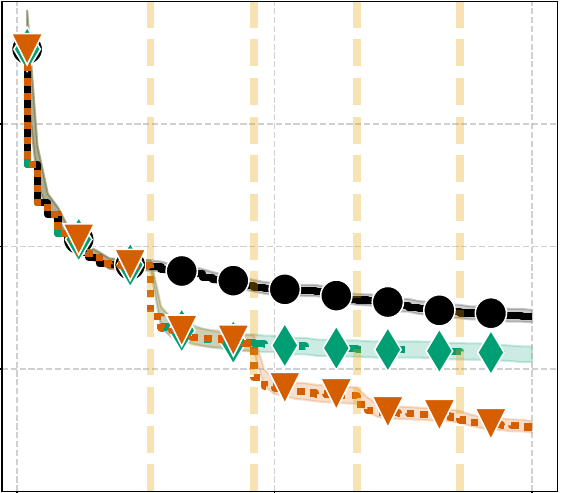} &
    \includegraphics[height=\figheight,trim={3.55cm 1.45cm 0cm 0.35cm},clip]{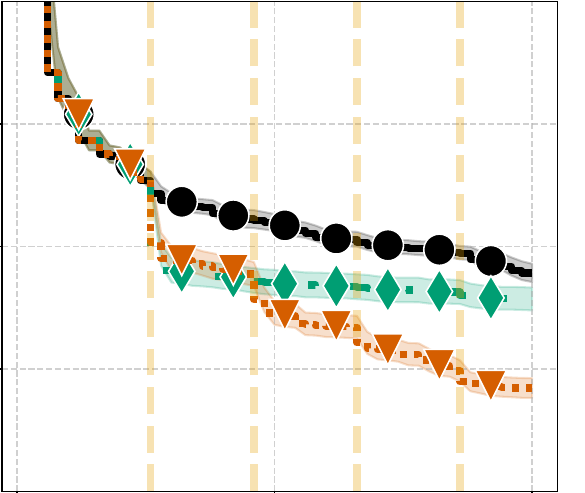} &
    \includegraphics[height=\figheight,trim={3.55cm 1.45cm 0cm 0.35cm},clip]{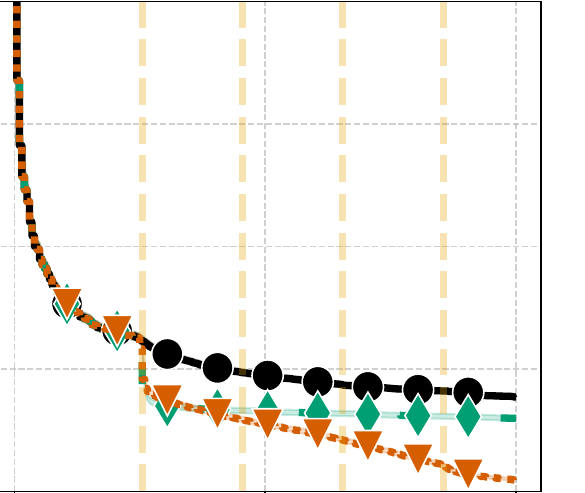}  &
    \includegraphics[height=\figheight,trim={3.55cm 1.45cm 0cm 0.35cm},clip]{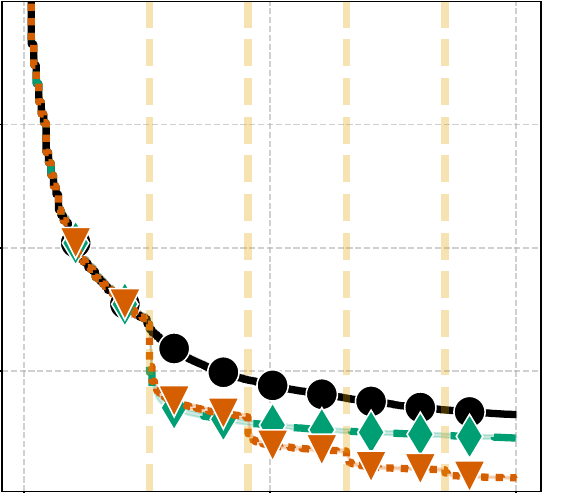}\\

    \rotatebox{90}{\hspace{.55cm}\texttt{Local}} &
    \includegraphics[height=\figheight,trim={0.35cm 1.45cm 0cm 0.35cm},clip]{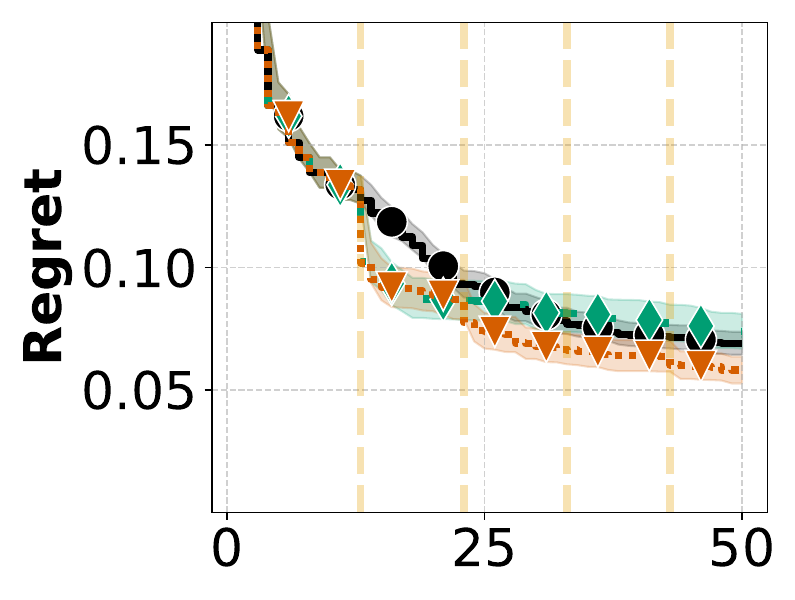} &
    \includegraphics[height=\figheight,trim={3.55cm 1.45cm 0cm 0.35cm},clip]{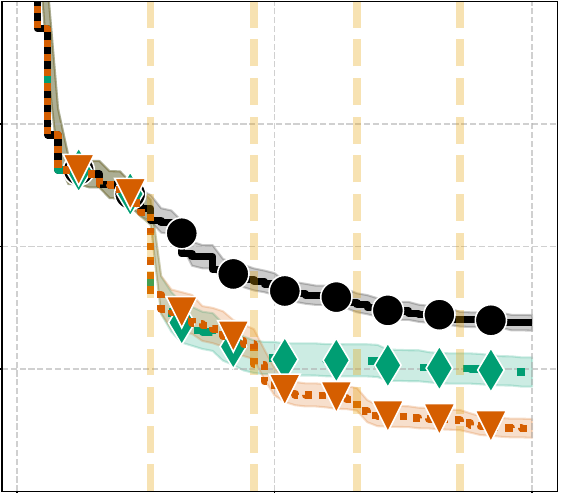} &
    \includegraphics[height=\figheight,trim={3.55cm 1.45cm 0cm 0.35cm},clip]{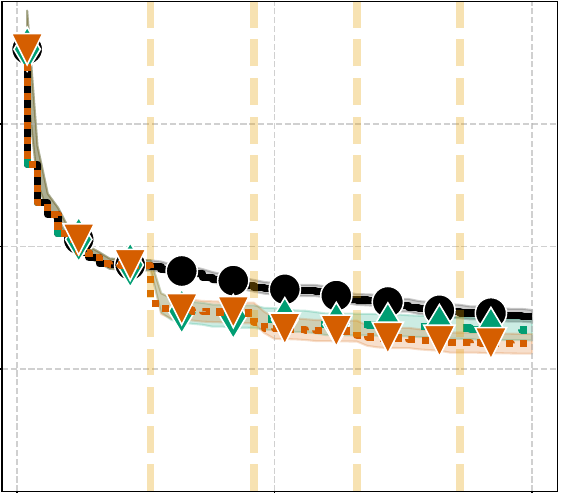} &
    \includegraphics[height=\figheight,trim={3.55cm 1.45cm 0cm 0.35cm},clip]{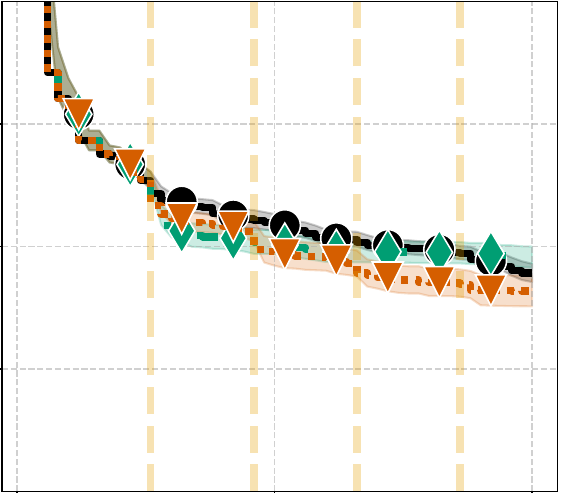} &
    \includegraphics[height=\figheight,trim={3.55cm 1.45cm 0cm 0.35cm},clip]{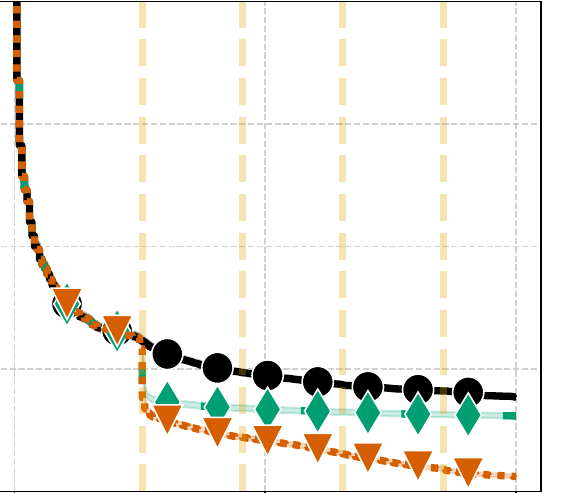}  &
    \includegraphics[height=\figheight,trim={3.55cm 1.45cm 0cm 0.35cm},clip]{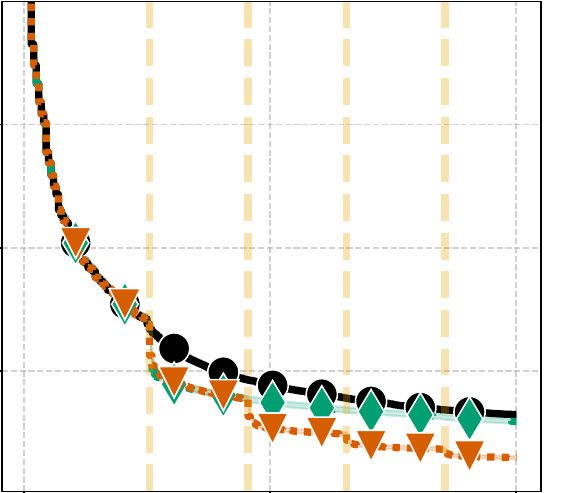}\\

    \rotatebox{90}{\hspace{.45cm}\texttt{Deceptive}} &
    \includegraphics[height=\figheight+0.13\figheight,trim={0.35cm 0.35cm 0cm 0.35cm},clip]{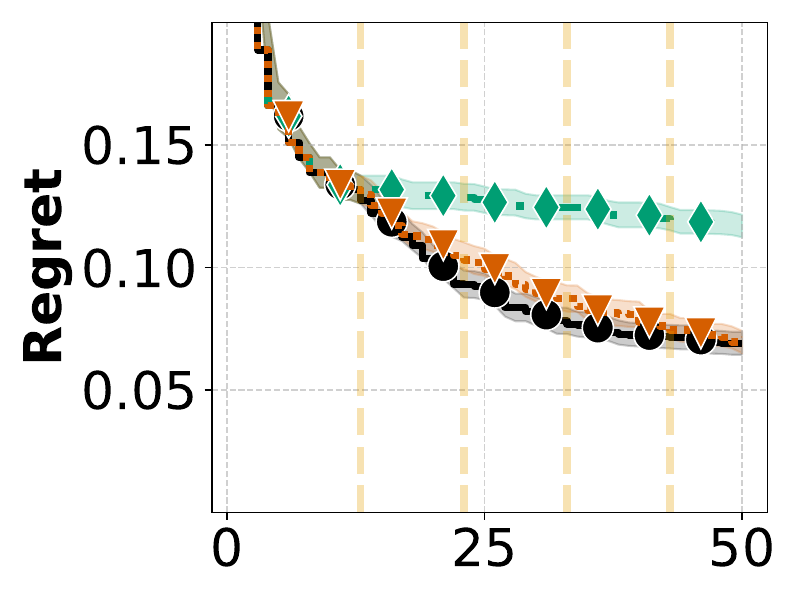} &
    \includegraphics[height=\figheight+0.13\figheight,trim={3.55cm 0.35cm 0cm 0.35cm},clip]{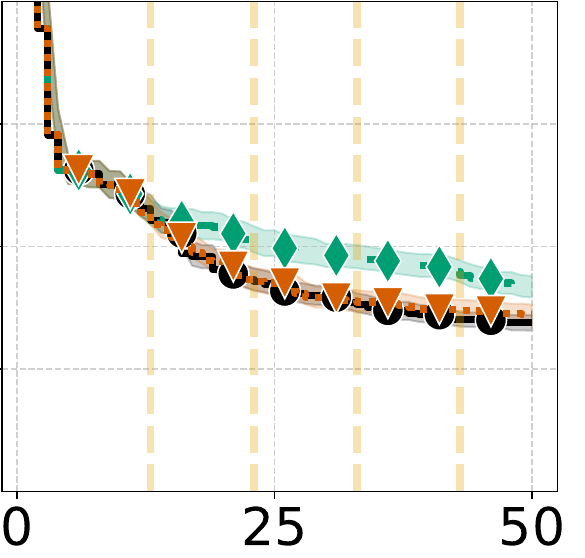} &
    \includegraphics[height=\figheight+0.13\figheight,trim={3.55cm 0.35cm 0cm 0.35cm},clip]{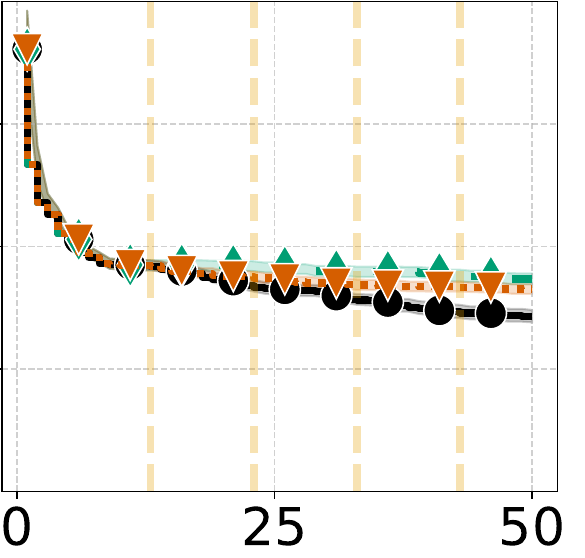} &
    \includegraphics[height=\figheight+0.13\figheight,trim={3.55cm 0.35cm 0cm 0.35cm},clip]{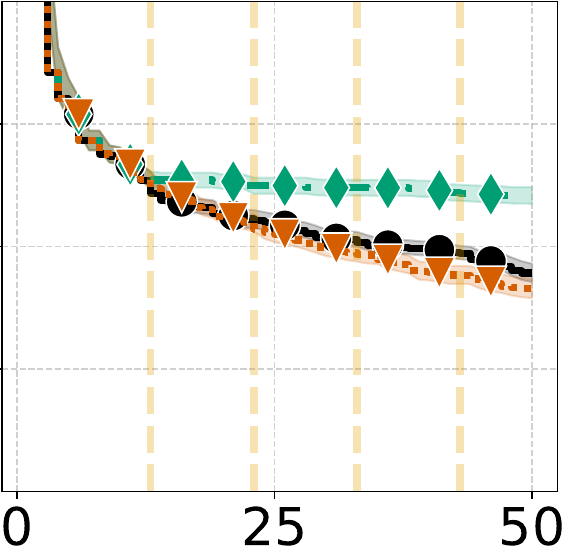} &
    \includegraphics[height=\figheight+0.13\figheight,trim={3.55cm 0.35cm 0cm 0.35cm},clip]{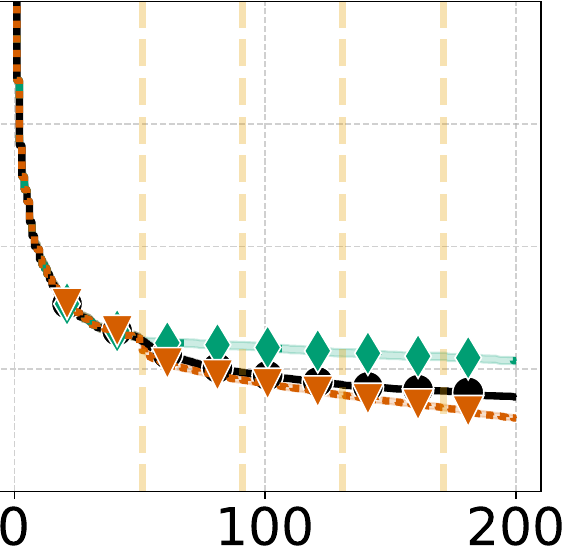}  &
    \includegraphics[height=\figheight+0.13\figheight,trim={3.55cm 0.35cm 0cm 0.35cm},clip]{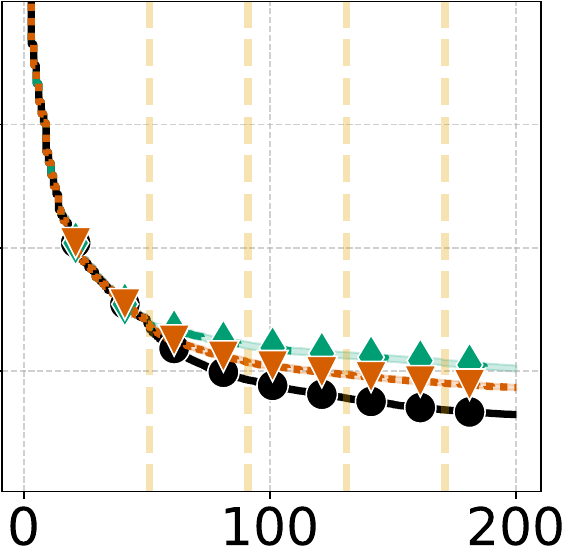}
  \end{tabular}

  \begin{minipage}{\textwidth}
    \centering
    \vspace{-.2cm}
    \includegraphics[width=.3\textwidth,trim={3cm 0cm .3cm 14cm},clip]{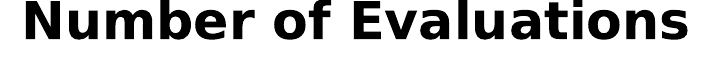}
  \end{minipage}

  \begin{minipage}{\textwidth}
    \centering
    \vspace{-.3cm}
    \includegraphics[width=.5\textwidth,trim={0cm .2cm 0cm 10.3cm},clip]{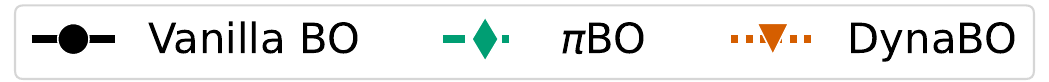}
  \end{minipage}
  \caption{Mean regret for \texttt{PD1}, \texttt{lcbench}, and \texttt{xgboost} using \texttt{Expert}, \texttt{Advanced}, \texttt{Local}, and \texttt{Deceptive} priors, provided at vertical lines. For \texttt{lcbench} and \texttt{xgboost}, the plots average all datasets.}
  \label{fig:main-results}
\end{figure*}

In \cref{fig:main-results}, we present anytime performance plots comparing Vanilla BO, \pibo, and \tool  \footnote{Note that vanilla BO does not accept any priors, and \pibo only accepts one prior at the beginning of optimization.}.

\paragraph{Expert and Advanced Priors}
Generally speaking, \tool outperforms vanilla BO, and \pibo significantly, in both anytime and final performance. On \texttt{widernet} and \texttt{xformer}, \tool is predominated by \pibo until the second prior is provided, due to overly-cautious rejection of priors. Overall, providing expert priors yields a slightly larger performance boost than providing advanced priors. 

\paragraph{Local Priors}
For local priors, we see similar results, with a reduced gain through the provided priors. In the case of \texttt{resnet}, local priors reduce rather than improve performance for both \pibo and \tool. However, this impacts \pibo more than \tool. Overall, local priors accelerate \tool.

\paragraph{Deceptive Priors}
Deceptive priors degrade performance for \pibo and \tool. However, our rejection mechanism results in a significant performance boost on all scenarios but \texttt{lm1b\_transformer}, where deceptive priors are often not recognized, further discussed in \cref{app:further-results:rejection_ablation}. The recovery from deceptive priors is analyzed in \cref{fig:increased_budget} using a $500$ trial budget. Here, \tool with rejection outperforms \pibo in anytime performance. If equipped with Gaussian processes, \pibo and \tool recover from deceptive priors; with random forests, \tool's prior rejection scheme is necessary.

\noindent
In general, \tool performs better than \pibo. Additionally, \tool's rejection mechanism boosts performance for both local and deceptive priors, while sacrificing only a little solution quality for expert and advanced priors. More results provided in the appendix reinforce this conclusion. For example,  \cref{app:further-results:gp-normal} presents theory aligned results with GPs and LCB. The additional runtime required by our computations adds roughly 12 seconds (including time spent sampling priors) to the overall virtual runtime of approximately 26 hours for an optimization run. 

\subsection{Prior Rejection Sensitivity Analysis}\label{sec:results:reject}

\begin{figure*}[t]
  \centering
  \begin{minipage}{.48\textwidth}
        \centering
        \hspace{1.3cm}{\texttt{Random Forest Surrogate}}
        \includegraphics[height=.295\textwidth,trim={.0cm .35cm .0cm 0cm}, clip]{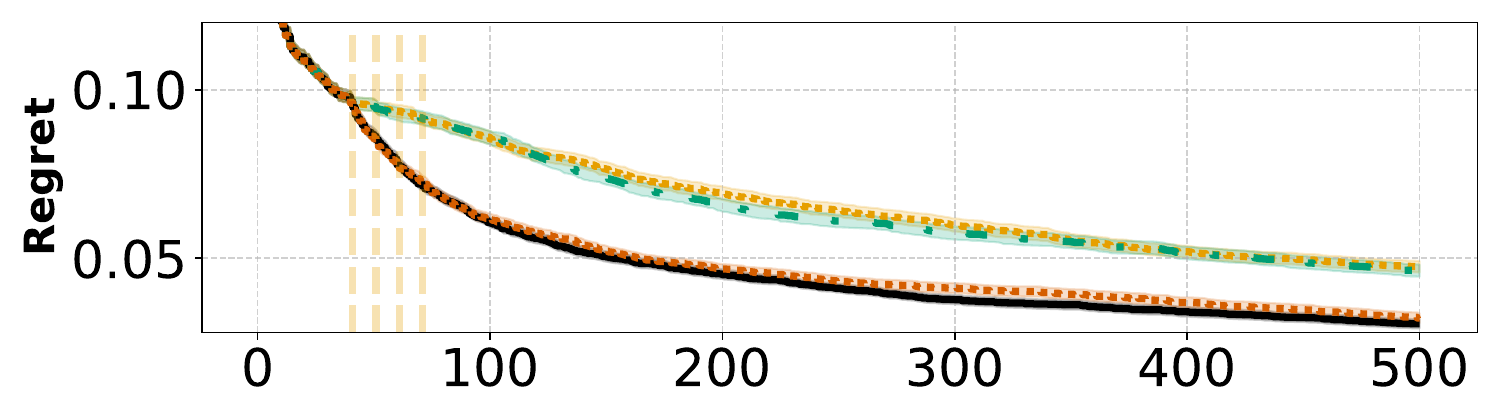}
    \end{minipage}
    \hfill
    \begin{minipage}{.48\textwidth}
        \centering
        {\texttt{Gaussian Process Surrogate}}
        \includegraphics[height=.295\textwidth,trim={3.35cm .35cm .35cm 0cm}, clip]{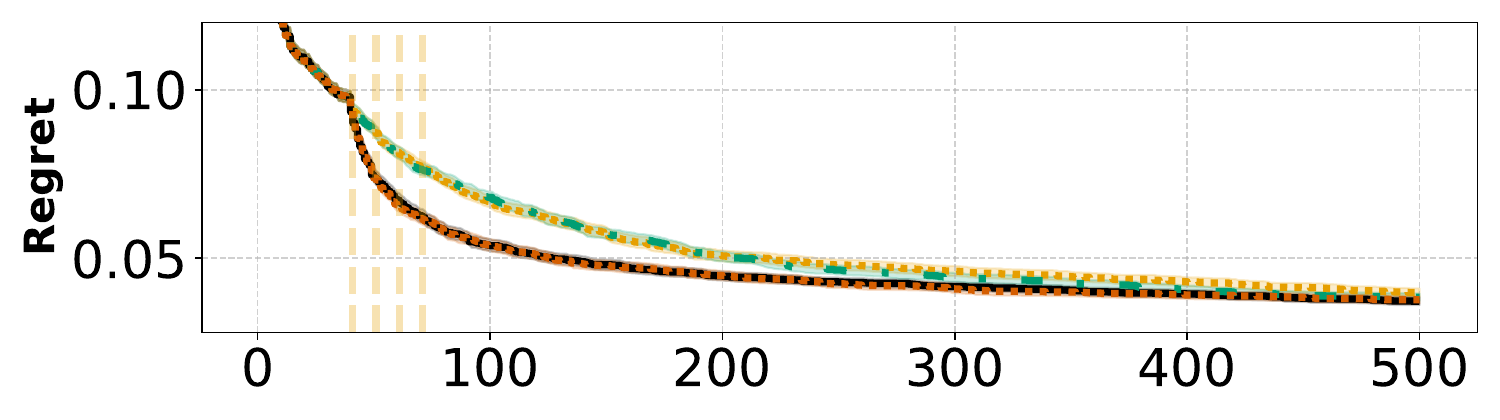}
    \end{minipage}
    \includegraphics[width=.7\textwidth,trim={.0cm 0cm 0cm 7.5cm},clip]{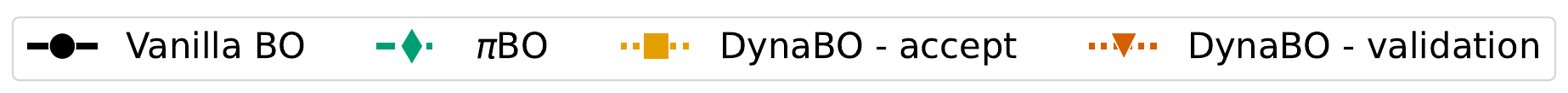}
    \caption{Anytime regret for \texttt{PD1} averaged over $30$ seeds, and scenarios comparing \texttt{vanilla BO}, \pibo, \tool accept all priors (\texttt{\tool-accept}), and \tool with validation (\texttt{\tool-validation}).}
    \label{fig:increased_budget}

  \vspace{2mm} 

  \begin{minipage}{.016\linewidth}
    \centering
    \includegraphics[width=\linewidth,trim={0.2cm .35cm 48.2cm 0cm},clip]{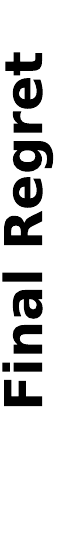}
  \end{minipage}
  \begin{minipage}{.65\linewidth}
    \centering
    \includegraphics[width=\linewidth,trim={1.12cm .35cm 16.5cm 0cm},clip]{figures/ICML_26_submission/final_results/prior_rejection_ablation/final_cost_barplot_overall_final_cost_barplot.pdf}
  \end{minipage}
  \begin{minipage}{.3\linewidth}
    \centering
    \vspace{.2cm}
    \includegraphics[width=\linewidth,trim={37cm .35cm 0cm 0cm},clip]{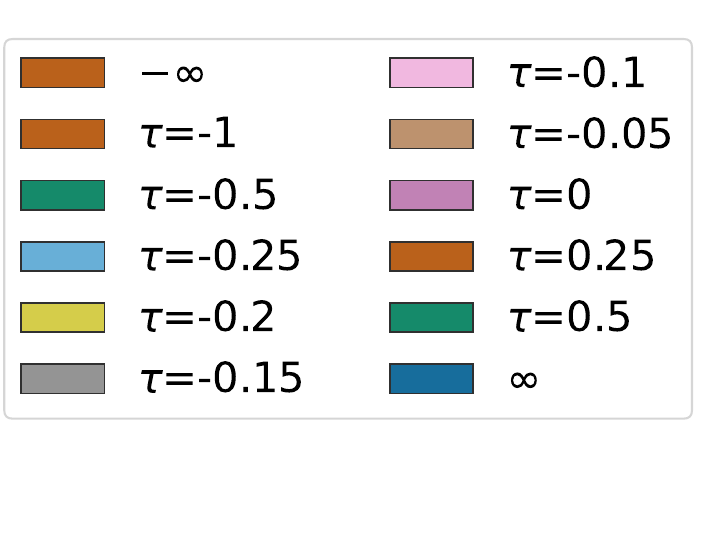}
  \end{minipage}
  
  \caption{Sensitivity analysis of different thresholds $\tau$. Setting $\tau = -\infty$ accepts all, while setting $\tau = \infty$ rejects all priors. The box plots contain the merged results from all PD1 scenarios.}
  \label{fig:ablation-results}
\end{figure*}

We also conduct a sensitivity analysis of the prior rejection criterion in \cref{fig:ablation-results}, evaluating the regret achieved at the end of the optimization process (scenario-specific plots and a detailed analysis are provided in \cref{app:further-results:rejection_ablation}). This analysis shows that $\tau =- 0.15$ achieves a good balance, accepting informative priors and rejecting misleading ones, while showing potential for further improvement by tailoring the rejection to the conducted experiment.

\section{Conclusion}\label{sec:conclusion}
In this work, we present \tool, allowing seamless integration of user beliefs into Bayesian optimization (BO) for hyperparameter optimization. \tool adapts acquisition functions by incorporating priors during optimization, while ensuring robustness by safeguarding against misleading priors. 
To validate our approach, we provide theoretical guarantees of convergence and robustness. Additionally, we are the first to prove finite-time acceleration if provided with informative priors. All of these theoretical guarantees hold even without the prior-rejection safeguard. 

In our detailed experimental comparison with PCs and \pibo, \tool outperforms PCs and \pibo irrespective of prior kind across different surrogate models and acquisition functions. Importantly, our evaluation moves beyond oracle priors on the optimal configuration by creating imperfect priors based on current performance. Additionally, even when provided with deceptive priors multiple times, \tool retains vanilla-BO's optimization qualities.  
This empirically demonstrates that \tool yields a considerable optimization speedup with minimal risk and overhead.

\section{Limitations and Future Work}
As in the related literature \citep{hvarfner-iclr22a, mallik-neurips23a, dblp-hvarfner-iclr24a, seng-automlconf25a}, while our empirical evaluation considers different kinds of user priors, they remain syntactically generated. Future work should focus on expanding the types of priors supported and on considering priors from different sources. Additionally, analyzing the effect of different surrogate models on prior handling in detail and extending \tool to multi-fidelity optimization, similar to the surrogate-free approach by~\citet{mallik-neurips23a}, is worth exploring. Lastly, directions for future work include enhancing \tool with explainable AI (XAI) and large language models (LLMs) to automatically create priors and support more transparent, user-friendly, low/no-code interfaces, fostering more intuitive collaboration between users and HPO approaches.

\paragraph{Impact Statement}
By incorporating human insights directly into the optimization loop, \tool promotes greater accessibility and democratization of machine learning tools. At the same time, our safeguard mechanism ensures that user priors do not compromise performance. This balance between human intuition and algorithmic rigor has the potential to accelerate the development of new AI applications. Still, careful consideration must be given to the provenance and quality of user priors to mitigate risks, such as encoding bias or the amplification of misconceptions. Additionally, as with any optimization method, \tool inherits the risks associated with the underlying problem formulation. In particular, in ML, there is a risk of implicitly exploiting undesirable data properties to optimize predictive accuracy. 

\begin{acknowledgements}
Lukas Fehring, Marcel Wever, and Marius Lindauer acknowledge funding by the European Union (ERC, ``ixAutoML'', grant no. 101041029). Views and opinions expressed are, however, those of the author(s) only and do not necessarily reflect those of the European Union or the European Research Council Executive Agency. Neither the European Union nor the granting authority can be held responsible for them. 
\begin{center}
\includegraphics[height=3cm]{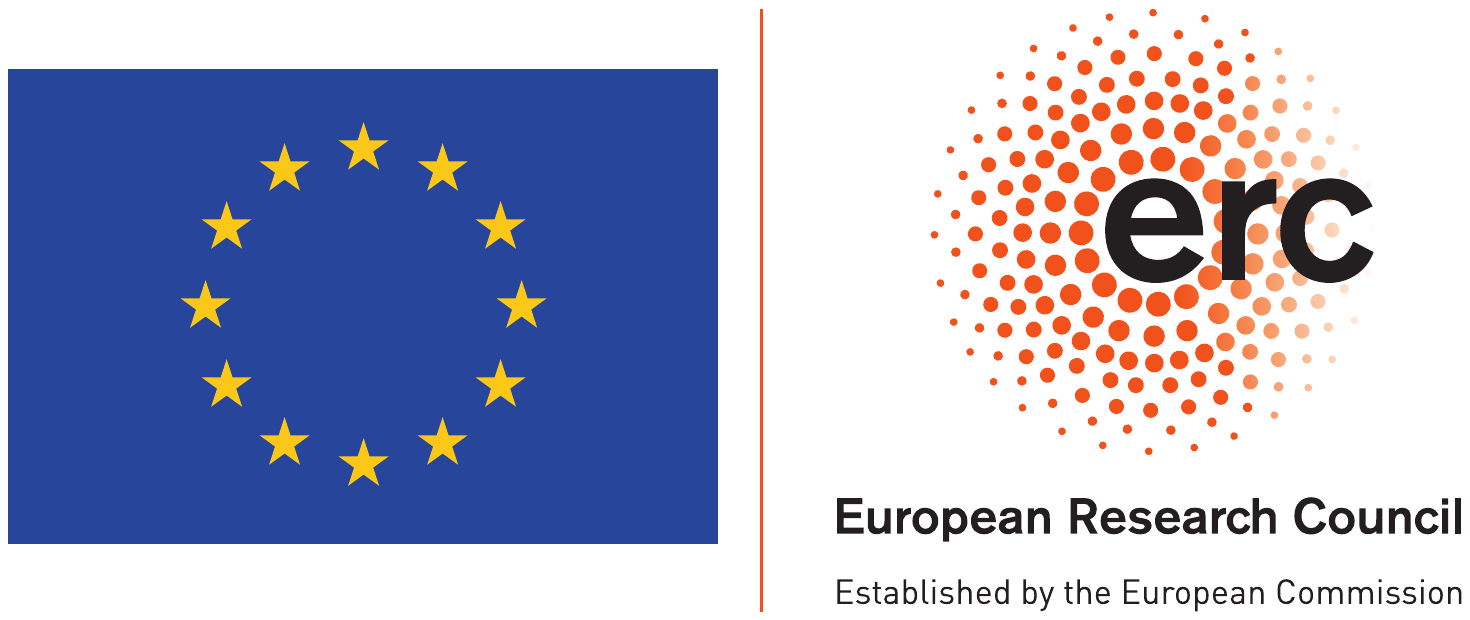}
\end{center}
Maximilian Spliethöver and Henning Wachsmuth have been supported by the Deutsche Forschungsgemeinschaft (DFG, German Research Foundation) under project number TRR 318/3 2026 – 438445824.

The authors gratefully acknowledge the computing time provided to them on the high-performance computers Noctua2 at the NHR Center PC2. These are funded by the Federal Ministry of Education and Research and the state governments participating based on the resolutions of the GWK for the national high-performance computing at universities (\url{www.nhr-verein.de/unsere-partner}). This work was supported by the Federal Ministry of Education and Research (BMBF), Germany, under the AI service center KISSKI (grant no. 01IS22093C).

We further thank our LUH-AI research team and the broader AutoML Research Community for their advice and discussions during presentations of preliminary work, especially Difan Deng and Carolin Benjamins, who proofread an early version of the manuscript. We also want to thank all anonymous peer reviewers, ACs, PCS, and the countless others behind the scenes whom we may fail to name, for their input, cooperation, consideration, and time investment.

\end{acknowledgements}


\bibliography{strings,lib,lib_ext,proc,proc_ext}




\newpage
\section*{Submission Checklist}


\begin{enumerate}
\item For all authors\dots
  \begin{enumerate}
  \item Do the main claims made in the abstract and introduction accurately
    reflect the paper's contributions and scope?
    \answerYes{}
  \item Did you describe the limitations of your work?
    \answerYes{}
  \item Did you discuss any potential negative societal impacts of your work?
    \answerNA{}
  \item Did you read the ethics review guidelines and ensure that your paper
    conforms to them? (see \url{https://2022.automl.cc/ethics-accessibility/})
    \answerYes{}
  \end{enumerate}
\item If you ran experiments\dots
  \begin{enumerate}
  \item Did you use the same evaluation protocol for all methods being compared (e.g.,
    same benchmarks, data (sub)sets, available resources, etc.)?
    \answerYes{For PCs, we choose to follow their initial design, after conferring with the main authors. This was an intentional choice on their part. We also adapted their source code to support priors for all hyperparameters.}
  \item Did you specify all the necessary details of your evaluation (e.g., data splits,
    pre-processing, search spaces, hyperparameter tuning details and results, etc.)?
    \answerYes{}{}
  \item Did you repeat your experiments (e.g., across multiple random seeds or
    splits) to account for the impact of randomness in your methods or data?
    \answerYes{}
  \item Did you report the uncertainty of your results (e.g., the standard error
    across random seeds or splits)?
    \answerYes{}
  \item Did you report the statistical significance of your results?
    \answerNo{}
  \item Did you use enough repetitions, datasets, and/or benchmarks to support
    your claims?
    \answerYes{}
  \item Did you compare performance over time and describe how you selected the
    maximum runtime?
    \answerYes{}
  \item Did you include the total amount of compute and the type of resources
    used (e.g., type of \textsc{gpu}s, internal cluster, or cloud provider)?
    \answerYes{}
  \item Did you run ablation studies to assess the impact of different
    components of your approach?
    \answerYes{}
  \end{enumerate}
\item With respect to the code used to obtain your results\dots
  \begin{enumerate}
\item Did you include the code, data, and instructions needed to reproduce the
    main experimental results, including all dependencies (e.g.,
    \texttt{requirements.txt} with explicit versions), random seeds, an instructive
    \texttt{README} with installation instructions, and execution commands
    (either in the supplemental material or as a \textsc{url})?
    \answerYes{}
  \item Did you include a minimal example to replicate results on a small subset
    of the experiments or on toy data?
    \answerYes{However, our experiments depend on prior data-generation runs. As a result, they can not be executed without the preceding data exploration runs. We therefore provide two minimal examples: One for the baseline (without priors) and one for \tool.}
  \item Did you ensure sufficient code quality and documentation so that someone else
    can execute and understand your code?
    \answerYes{}
  \item Did you include the raw results of running your experiments with the given
    code, data, and instructions?
    \answerNo{The raw results would exceed the common file limit. We are happy to share them if requested. However, all plots used in this paper are based on the raw data, and all plotting and data creation scripts are provided.}
  \item Did you include the code, additional data, and instructions needed to generate
    the figures and tables in your paper based on the raw results?
    \answerYes{}
  \end{enumerate}
\item If you used existing assets (e.g., code, data, models)\dots
  \begin{enumerate}
  \item Did you cite the creators of used assets?
    \answerYes{}
  \item Did you discuss whether and how consent was obtained from people whose
    data you're using/curating if the license requires it?
    \answerNA{We believe that this is not relevant to our conducted experiments.}
  \item Did you discuss whether the data you are using/curating contains
    personally identifiable information or offensive content?
    \answerNA{We believe that this is not relevant to our conducted experiments.}
  \end{enumerate}
\item If you created/released new assets (e.g., code, data, models)\dots
  \begin{enumerate}
    \item Did you mention the license of the new assets (e.g., as part of your
    code submission)?
    \answerYes{}{}
    \item Did you include the new assets either in the supplemental material or as
    a \textsc{url} (to, e.g., GitHub or Hugging Face)?
    \answerYes{}{}
  \end{enumerate}
\item If you used crowdsourcing or conducted research with human subjects\dots
  \begin{enumerate}
  \item Did you include the full text of instructions given to participants and
    screenshots, if applicable?
    \answerNA{}{}
  \item Did you describe any potential participant risks, with links to
    institutional review board (\textsc{irb}) approvals, if applicable?
    \answerNA{}{}
  \item Did you include the estimated hourly wage paid to participants and the
    total amount spent on participant compensation?
    \answerNA{}{}
  \end{enumerate}
\item If you included theoretical results\dots
  \begin{enumerate}
  \item Did you state the full set of assumptions of all theoretical results?
    \answerYes{}{}
  \item Did you include complete proofs of all theoretical results?
    \answerYes{}{}
  \end{enumerate}
\end{enumerate}

\newpage
\appendix


\newpage
\appendix
\crefalias{section}{appendix}
\crefalias{subsection}{appendix}
\section*{Organization of Technical Appendices and Supplementary Material}
The appendix is split into four main parts. In \cref{app:proofs}, we prove our theoretical guarantees. In \cref{app:priors}, we provide a detailed discussion of our prior construction and selection scheme. In \cref{app:detailed_experimental_setup}, we discuss the experimental setup. Lastly, in \cref{app:further-results}, we provide additional evaluation results.

\contentsline {section}{\numberline{A} Theoretical Guarantees}{\pageref{app:proofs}}{}
\contentsline {subsection}{\numberline{A.1} Theorem - Almost Sure Convergence of \tool}{\pageref{theorem convergence dynabo}}{}
\contentsline {subsection}{\numberline{A.2} Corollary - Robustness to Misleading Priors}{\pageref{corollary robustness dynabo}}{}
\contentsline {subsection}{\numberline{A.3} Theorem - Acceleration of Convergence with Informative Priors}{\pageref{app:proof-acceleration}}{}

\contentsline {section}{\numberline{B} Aggregation of Priors: Summation vs. Multiplication }{\pageref{app:sum-vs-product}}{}
\contentsline {subsection}{\numberline{B.1} Conceptual Argument: Union vs.\ Intersection of Beliefs}{\pageref{app:cancellation}}{}
\contentsline {subsection}{\numberline{B.2} Numerical Argument: Stability Under Additional Priors }{\pageref{app:numerical}}{}
\contentsline {subsection}{\numberline{B.3} Probabilistic Argument: When Is Multiplication Justified? }{\pageref{app:probabilistic}}{}
\contentsline {subsection}{\numberline{B.4} Experimental Results }{\pageref{app:sum_vs_product_experiments}}{}

\contentsline {section}{\numberline{C} Additional \tool Details}{\pageref{app:additional_details}}{}
\contentsline {subsection}{\numberline{C.1} Facilitating Prior Behavior}{\pageref{app:facilitating_prior_behavior}}{}
\contentsline {subsection}{\numberline{C.2} Further Details on the Rejection Criterion}{\pageref{app:further_details_prior_rejection}}{}

\contentsline {section}{\numberline{D} Artificial Prior Generation}{\pageref{app:priors}}{}
\contentsline {subsection}{\numberline{D.1} Different Prior Kinds}{\pageref{app:different_prior_kinds}}{}
\contentsline {subsection}{\numberline{D.2} Data Generation Runs}{\pageref{app:data-generation}}{}
\contentsline {subsection}{\numberline{D.3} Prior Construction}{\pageref{app:prior_construction}}{}

\contentsline {section}{\numberline{E} Detailed Experimental Setup}{\pageref{app:detailed_experimental_setup}}{}
\contentsline {subsection}{\numberline{E.1}Summary of the Benchmark}{\pageref{app:detailed_experimental_setup:benchmarks}}{}
\contentsline {subsection}{\numberline{E.2}Cluster Setup}{\pageref{app:detailed_experimental_setup:cluster_setup}}{}
\contentsline {subsection}{\numberline{E.3}Competitor Setup}{\pageref{app:detailed_experimental_setup:competitors}}{}

\contentsline {section}{\numberline{F} Additional Empirical Results}{\pageref{app:further-results}}{}
\contentsline {subsection}{\numberline{F.1} Increased Budget Results}{\pageref{app:further-results:increased-budget}}{}
\contentsline {subsection}{\numberline{F.2} LCB Results}{\pageref{app:further-results:lcb_results}}{}
\contentsline {subsection}{\numberline{F.3} Sensitivity Analysis of the Prior Rejection Criterion}{\pageref{app:further-results:rejection_ablation}}{}
\contentsline {subsection}{\numberline{F.4} Detailed Comparison with Probabilistic Circuits}{\pageref{app:further-results:pc}}{}
\contentsline {subsection}{\numberline{F.5} Gaussian Process Main Results}{\pageref{app:further-results:gp-normal}}{}
\contentsline {subsection}{\numberline{F.6} Random Prior Location}{\pageref{app:further-results:dynamic}}{}
\contentsline {subsection}{\numberline{F.7}Na{\"i}ve Dynamic Extension for $\pi$BO}{\pageref{pibo_extension}}{}
\contentsline {subsection}{\numberline{F.8} Prior Decay Ablation}{\pageref{app:further-results:prior-decay}}{}
\contentsline {subsection}{\numberline{F.9} Investigating the Impact of $\beta$}{\pageref{beta}}{}
\contentsline {subsection}{\numberline{F.10} Prior Rejection Sampling Budget Ablation}{\pageref{app:further-results:rior_rejection_sampling_budget_ablation}}{}

\contentsline {section}{\numberline{G} Declaration of LLM Usage}{\pageref{app:llm-usage}}{}
\newpage

\section{Theoretical Guarantees}\label{app:proofs}
For all the subsequent proofs, we make the following assumptions:

\textbf{A1 - Objective Function Regularity:} The objective function $f_\text{obj}:\Lambda\rightarrow\mathbb{R}$ is a sample from a Gaussian process with mean  $m:\Lambda\rightarrow\mathbb{R}$ and positive definite kernel $k(\lambda, \lambda')$. Its Reproducing kernel Hilbert space (RKHS) norm is bounded almost surely by $\mid\mid f\mid\mid_{\mathcal{H}_k}\leq B$. The domain $\Lambda\subset\mathbb{R}^d$  is compact.

W.l.o.g., we minimize the objective function. However, for the theoretical analysis, we consider the maximization of the equivalent utility function $f(\lambda) \coloneqq K - f_\text{obj}(\lambda)$, where the constant $K$ is chosen to ensure $f(\lambda) > 0$ for all $\lambda \in \Lambda$. This affine transformation preserves the global optimum and guarantees the strict positivity required for multiplicative re-weighting via the sum of priors.

\textbf{A2 - Finiteness of User Priors:} Let $\{(t^{(m)}, \pi^{(m)})\}_{m=1}^M$ be a finite series of user-specified prior functions $\pi ^{(m)}:\Lambda\rightarrow (0,1]$ with iteration indices $t^{(m)}\in\mathbb{N}$ such that $t^{(1)} <...<t^{(M)}$. Moreover, we assume that $\Lambda\subset\mathbb{R}^d$ is compact and that each prior $\pi^{(m)}$ is continuous on $\Lambda$.

\textbf{A3 - Vanishing Influence of Priors:} There exists $\beta\in\mathbb{R^+}$ such that for all indices $m$ and $\lambda\in\Lambda$ $$\lim_{t \rightarrow \infty} \pi^{(m)}(\lambda)^{\beta/(t-t^{(m)})} = 1.$$ That is, the multiplicative influence of the prior function on the acquisition criterion diminishes to unity with increasing number of iterations. The function $t\mapsto \pi^{(m)}(\lambda)^{\beta/(t-t^{(m)})}$ is monotone in $t$ for every $m$ and $\lambda\in\Lambda$.

Additionally, we assume Upper Confidence Bound (UCB) as an acquisition function for maximization of the utility function.

\subsection{Theorem - Almost Sure Convergence of \tool}\label{theorem convergence dynabo}
Let $\lambda^* \in \arg\max_{\lambda \in \Lambda} f(\lambda)$ be a global maximizer of $f$.
Given Assumptions A1, A2 and A3, the sequence of points ${(\lambda_t)}_{t\geq 1}$ selected by \tool satisfies
\begin{equation}
    \lim_{T\to\infty} \min_{t \le T} \bigl(f(\lambda^*) - f(\lambda_t)\bigr) = 0 \qquad \text{a.s.}
\end{equation}

\begin{proof}
The proof follows standard convergence results for Bayesian optimization (BO)~\citep{srinivas-it12a}, augmented by the dynamic influence of priors. At iteration $t>t^{(M)}$, the \tool acquisition function is defined as
\begin{equation}
    \alpha_\text{dyna}(\lambda,t)\coloneqq\alpha(\lambda,t)\sum_{m=1}^{M}\pi^{(m)}(\lambda)^{\beta/ (t-t^{(m)})},
\end{equation}
where $\alpha(\lambda,t)$ denotes a standard GP-UCB acquisition function.

First, we establish uniform convergence. Since $\pi^{(m)}$ is strictly positive and continuous on the compact set $\Lambda$, the log-prior is bounded. Let $L_m \coloneqq \sup_{\lambda\in\Lambda}|\log\pi^{(m)}(\lambda)| < \infty$. Then for all $\lambda\in\Lambda$, defining $h_t^{(m)}(\lambda) = \exp(\tfrac{\beta}{t - t^{(m)}} \log\pi^{(m)}(\lambda))$, we have the bound $|h_t^{(m)}(\lambda) - 1| \le \exp(\tfrac{\beta}{t - t^{(m)}} L_m) - 1$. The right-hand side converges to $0$ uniformly as $t\to\infty$. Summing over $m$ implies that the weighting term $P(\lambda,t)$ converges uniformly to $M$:
\begin{equation}
    P(\lambda,t) \xrightarrow[t\to\infty]{\text{uniform}} M.
\end{equation}

Next, we bound the deviation $\Delta_t \coloneqq \sup_{\lambda\in\Lambda} |\alpha_{\mathrm{dyna}}(\lambda,t) - M\alpha(\lambda,t)|$. Noting that $\alpha_{\mathrm{dyna}}(\lambda,t) - M\alpha(\lambda,t) = \alpha(\lambda,t)(P(\lambda,t)-M)$, we have
\begin{equation}
    \Delta_t \le \left(\sup_{\lambda\in\Lambda}|\alpha(\lambda,t)|\right) \cdot \sup_{\lambda\in\Lambda}|P(\lambda,t)-M|.
\end{equation}
Under standard GP-UCB assumptions, $\sup_{\lambda}|\alpha(\lambda,t)| = O(\sqrt{\log t})$. Together with $\sup_{\lambda}|P(\lambda,t)-M| \to 0$, this implies $\Delta_t \to 0$.

Let $\lambda_t$ and $\lambda_t^{\mathrm{UCB}}$ be maximizers of $\alpha_{\mathrm{dyna}}(\cdot,t)$ and $\alpha(\cdot,t)$, respectively. Using the maximizing property $\alpha_{\mathrm{dyna}}(\lambda_t^{\mathrm{UCB}},t) \le \alpha_{\mathrm{dyna}}(\lambda_t,t)$ and the definition of $\Delta_t$, we derive
\begin{equation}
    \alpha(\lambda_t^{\mathrm{UCB}},t) - \alpha(\lambda_t,t) \le \frac{2\Delta_t}{M} \eqqcolon \varepsilon_t, \quad \text{where } \varepsilon_t \xrightarrow[t\to\infty]{} 0.
\end{equation}
Thus, $\lambda_t$ is an asymptotically exact maximizer of $\alpha(\cdot,t)$.

\citet{srinivas-it12a} show that GP-UCB with an approximate maximizer satisfying $\alpha(\lambda_t,t) \ge \max_{\lambda}\alpha(\lambda,t) - \varepsilon_t$ incurs cumulative regret bounded by $R_T \le C\sqrt{T\beta_T\gamma_T} + \sum_{t=1}^T \varepsilon_t$. Since $\varepsilon_t \to 0$, the average error vanishes, i.e., $\frac{1}{T}\sum_{t=1}^T \varepsilon_t \to 0$. Therefore, $R_T/T \to 0$. Finally, since $\min_{t\le T}(f(\lambda^*) - f(\lambda_t)) \le R_T/T$, we conclude
\begin{equation}
    \lim_{T\to\infty} \min_{t \le T} \bigl(f(\lambda^*) - f(\lambda_t)\bigr) = 0 \qquad \text{a.s.}
\end{equation}
\end{proof}

\subsection{Corollary - Robustness to misleading priors}
\label{corollary robustness dynabo}
Under Assumptions~A1--A3, \tool is robust to misleading priors: such priors may cause a finite amount of additional exploration, but they do not affect the asymptotic simple regret compared to vanilla GP-UCB.
Let 
\begin{equation*}
    A_T^{\mathrm{UCB}} := \max_{1 \le t \le T} f(\lambda_t^{\mathrm{UCB}}),
\qquad
A_T^{\mathrm{Dyna}} := \max_{1 \le t \le T} f(\lambda_t^{\mathrm{DynaBO}}).
\end{equation*}
Then
\begin{equation*}
    \lim_{T\to\infty} \bigl(A_T^{\mathrm{Dyna}} - A_T^{\mathrm{UCB}}\bigr) = 0
\quad \text{almost surely}.
\end{equation*}
In particular,
\begin{equation*}
    \liminf_{T\to\infty} \bigl(A_T^{\mathrm{Dyna}} - A_T^{\mathrm{UCB}}\bigr) \ge 0.
\end{equation*}
Let $s_T^{\mathrm{UCB}} := f(\lambda^*) - A_T^{\mathrm{UCB}}$ and
$s_T^{\mathrm{Dyna}} := f(\lambda^*) - A_T^{\mathrm{Dyna}}$ be the simple
regrets. Vanilla GP-UCB satisfies $s_T^{\mathrm{UCB}} \to 0$
\citep{srinivas-it12a}, and Theorem~\ref{theorem convergence dynabo}
gives $s_T^{\mathrm{Dyna}} \to 0$. For every $T$,
\begin{equation*}
    A_T^{\mathrm{Dyna}} - A_T^{\mathrm{UCB}}
= s_T^{\mathrm{UCB}} - s_T^{\mathrm{Dyna}},
\end{equation*}
which converges to $0$ almost surely. Assumption~A3 ensures that
misleading priors only influence a finite number of iterations.

\subsection{Theorem - Acceleration of Convergence with Informative Priors}
\label{app:proof-acceleration}

While Theorem~\ref{theorem convergence dynabo} establishes asymptotic consistency via prior decay, we now characterize the performance gain during the finite horizon $T$ where the prior remains influential.

Let Assumptions A1--A3 hold. Suppose there exists a set $U_\epsilon \subset \Lambda$ of diameter 
$\epsilon > 0$ and $\delta \in (0,1)$ such that the prior-induced weighting $P(\lambda, t)$ 
concentrates \tool's sampling:
\begin{equation}
    P(\lambda_t \in U_\epsilon) \geq 1 - \delta, \quad \forall t \in \{1, \ldots, T\}.
\end{equation}
Then the expected cumulative regret of \tool after $T$ iterations satisfies
\begin{equation}
    \mathbb{E}[R_T] \leq C\sqrt{T \beta_T^{\mathrm{UCB}} \gamma_T(U_\epsilon)} + \delta T B,
\end{equation}
where $B$ is the maximum instantaneous regret and $\gamma_T(U_\epsilon)$ is the information gain restricted to $U_\epsilon$. Since $\gamma_T(U_\epsilon) < \gamma_T(\Lambda)$, this bound demonstrates an acceleration over standard BO provided $\delta$ is sufficiently small.

\begin{proof}
We partition the time steps into two sets: those where the selected point $\lambda_t$ falls within the informative region $U_\epsilon$, and those where it falls outside. Define the index sets:
\begin{equation}
    \mathcal{I}_{U_\epsilon}\coloneqq\{1\le t\le T : \lambda_t\in U_\epsilon\}, \qquad \mathcal{I}_{\text{out}}\coloneqq\{1\le t\le T : \lambda_t\notin U_\epsilon\}.
\end{equation}

For iterations in $\mathcal{I}_{U_\epsilon}$, the queried points lie in $U_\epsilon$. Therefore, the regret decomposition of \citet{srinivas-it12a} applies with the information gain restricted to $U_\epsilon$. Following \citet{srinivas-it12a}, the sum of squares of instantaneous regrets is bounded by the mutual information. Applying the Cauchy–Schwarz inequality yields the standard regret bound scaled to the effective dimension of $U_\epsilon$:
\begin{equation}
    \sum_{t\in\mathcal{I}_{U_\epsilon}} (f(\lambda^*) - f(\lambda_t)) 
    \le C\sqrt{|\mathcal{I}_{U_\epsilon}|\,\beta_T^{\mathrm{UCB}}\,\gamma_{|\mathcal{I}_{U_\epsilon}|}(U_\epsilon)}
    \le C\sqrt{T\,\beta_T^{\mathrm{UCB}}\,\gamma_T(U_\epsilon)}.
\end{equation}
For iterations in $\mathcal{I}_{\text{out}}$, we bound the regret by the worst-case instantaneous regret $B \coloneqq \sup_{\lambda\in\Lambda}(f(\lambda^*) - f(\lambda))$. Using the concentration assumption $\mathbb{P}(\lambda_t\in U_\epsilon) \geq 1-\delta$,
we have $\mathbb{P}(\lambda_t\notin U_\epsilon)\leq\delta$ for all $t$.
Thus,
\begin{equation}
    \mathbb{E}[|\mathcal{I}_{\text{out}}|] 
    = \sum_{t=1}^T \mathbb{P}(\lambda_t\notin U_\epsilon)
    \le \delta T.
\end{equation}
Consequently, the expected regret contribution from these outliers is:
\begin{equation}
    \mathbb{E}\!\left[\sum_{t\in\mathcal{I}_{\text{out}}} (f(\lambda^*) - f(\lambda_t))\right] \le B \cdot \mathbb{E}[|\mathcal{I}_{\text{out}}|] \le \delta T B.
\end{equation}
Combining the two terms by linearity of expectation:
\begin{equation}
    \mathbb{E}[R_T] 
    = \mathbb{E}[R_{T, \text{in}}] + \mathbb{E}[R_{T, \text{out}}]
    \le C\sqrt{T\,\beta_T^{\mathrm{UCB}}\,\gamma_T(U_\epsilon)} + \delta T B.
\end{equation}
\end{proof}

\newpage

\section{Aggregation of Priors: Summation vs. Multiplication}
\label{app:sum-vs-product}

\begin{wrapfigure}{r}{0.45\linewidth}
    \centering
    \includegraphics[width=\linewidth]{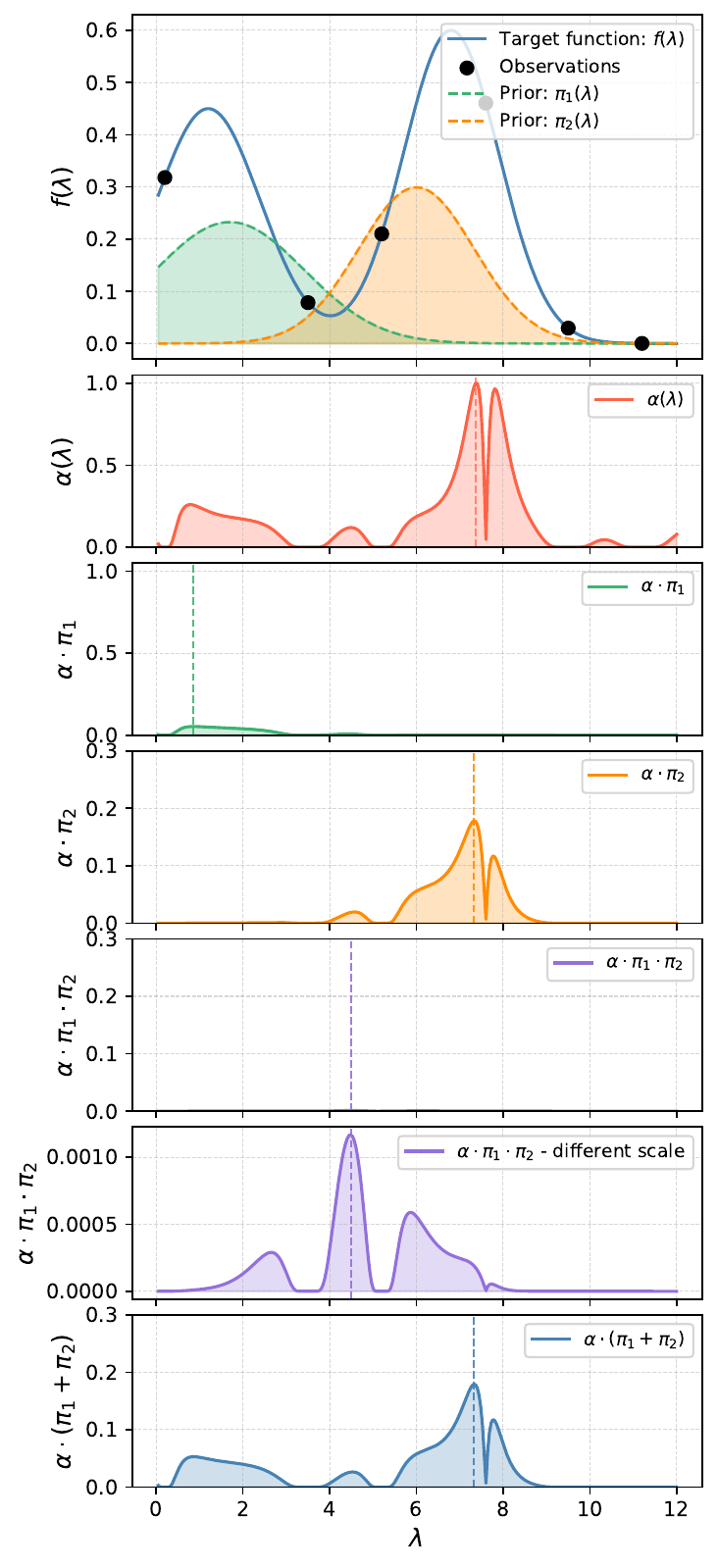}
    \caption{Comparison of strategies for adapting the acquisition function. From top to bottom: the optimization landscape and priors; the unmodified acquisition function; the acquisition function multiplied with the priors individually; the acquisition function multiplied with both priors (raw and rescaled); and the acquisition function multiplied with the sum of the priors.}
    \label{fig:acquisition_function_comparison}
\end{wrapfigure}

When multiple priors are available over the search space, two natural candidates for combining them into a single acquisition signal are summation and multiplication. In this work, we aggregate priors by summation. Below we give three reasons---conceptual (\cref{app:cancellation}), numerical (\cref{app:numerical}), and probabilistic (\cref{app:probabilistic})---for this choice, and provide supporting experimental evidence in \cref{app:sum_vs_product_experiments}. \cref{fig:acquisition_function_comparison} illustrates the qualitative differences between the two operators and serves as a visual reference for the discussion that follows.

\subsection{Conceptual Argument: Union vs.\ Intersection of Beliefs}
\label{app:cancellation}

Multiplying priors behaves like an \emph{intersection} of beliefs: probability mass survives only in regions where all priors simultaneously assign non-negligible density. Consequently, any configuration that scores poorly under even a single prior is effectively penalized, regardless of how strongly it is supported by the remaining priors. When priors target different regions of the search space---as is typical in our setting---this causes unintended cancellation, concentrating mass only in the (often small or empty) intersection. We then also provide an experimental evaluation in \cref{app:sum_vs_product_experiments} further strengthening our arguments.

Summation, by contrast, behaves like a \emph{union} of beliefs and preserves the multi-modal structure induced by priors that disagree. Each prior contributes meaningfully to the acquisition function, so that regions favored by any individual prior remain candidates for exploration. This contrast is visible in \cref{fig:acquisition_function_comparison}, where the product collapses to the overlap of the input priors while the sum retains each of their modes.

\subsection{Numerical Argument: Stability Under Additional Priors}
\label{app:numerical}

Beyond the conceptual issue discussed in \cref{app:cancellation}, the product formulation suffers from a structural numerical problem. Each prior contributes a factor, so the aggregated acquisition function tends towards zero as additional priors are incorporated. In the limit, the acquisition function becomes numerically uninformative, undermining the optimization procedure regardless of the quality of the priors. Summation does not exhibit this pathology, as its magnitude scales gracefully with the number of priors. This is visually represented by $\alpha \cdot \pi_1 \cdot \pi_2$ in \cref{fig:acquisition_function_comparison}.

\subsection{Probabilistic Argument: When Is Multiplication Justified?}
\label{app:probabilistic}

The product of priors admits a clean Bayesian interpretation, but this interpretation relies on a specific assumption that does not hold in our setting: namely, that the priors are conditionally independent observations of a shared latent quantity. Under this assumption, multiplication corresponds to sequential belief updating and is the correct operation.

In our setting, priors do not encode independent observations of a shared objective. They encode \emph{distinct user preferences} over different regions of the search space---heterogeneous signals that must be \emph{aggregated} rather than \emph{combined as evidence}. Multiplying them therefore conflates aggregation with updating, which is precisely the wrong operation here. Summation is the natural aggregation operator in this context, as it respects the union of preferences without treating the priors as redundant observations of the same underlying truth.

\subsection{Experimental Results}
\label{app:sum_vs_product_experiments}

\begin{figure}[H]
  \centering
  \setlength{\figheight}{0.18\textwidth}
  
  \resizebox{\textwidth}{!}{
  \begin{tabular}{@{}c@{}c@{}c@{}c@{}c@{}c}
    & \hspace{.6cm}\texttt{widernet} & \texttt{resnet} & \texttt{transf} & \texttt{xformer}\\ 
    \rotatebox{90}{\texttt{\hspace{.4cm} Expert}} &
    \includegraphics[height=\figheight,trim={0.35cm 1.45cm 0cm 0.35cm},clip]{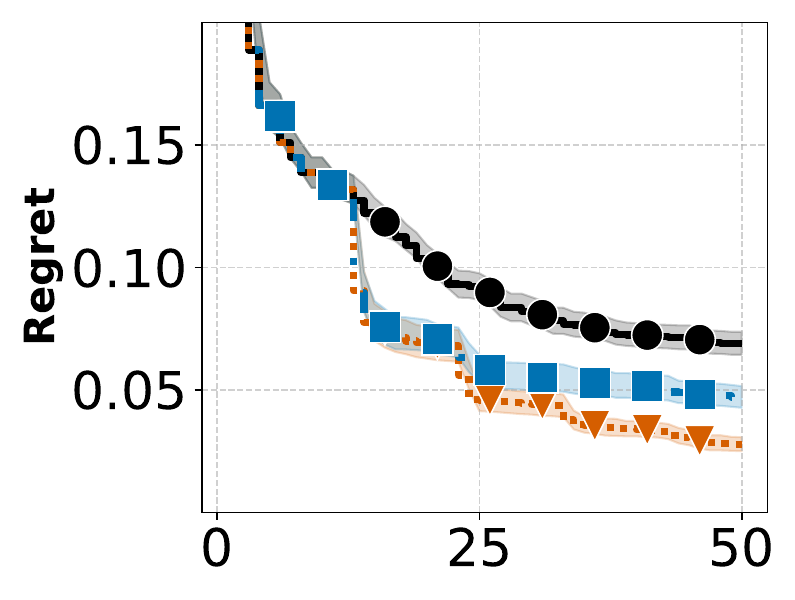} &
    \includegraphics[height=\figheight,trim={3.25cm 1.45cm 0cm 0.35cm},clip]{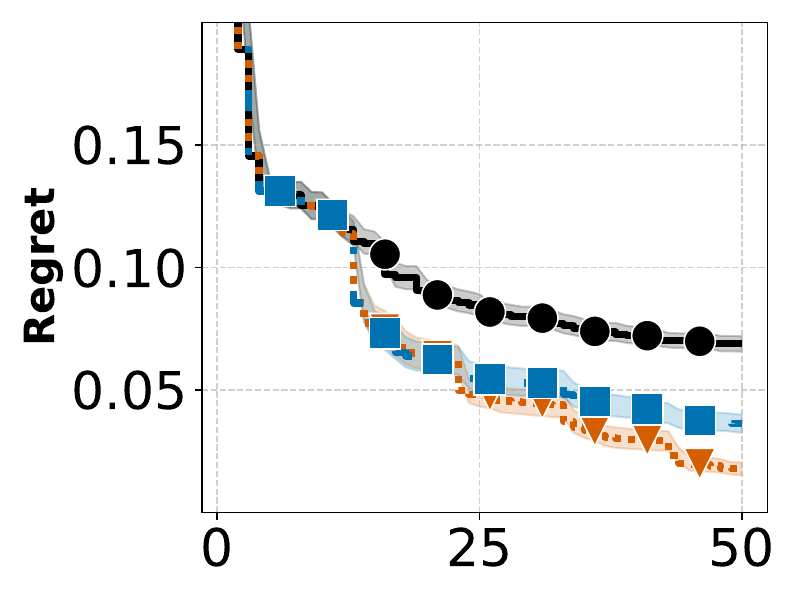} &
    \includegraphics[height=\figheight,trim={3.25cm 1.45cm 0cm 0.35cm},clip]{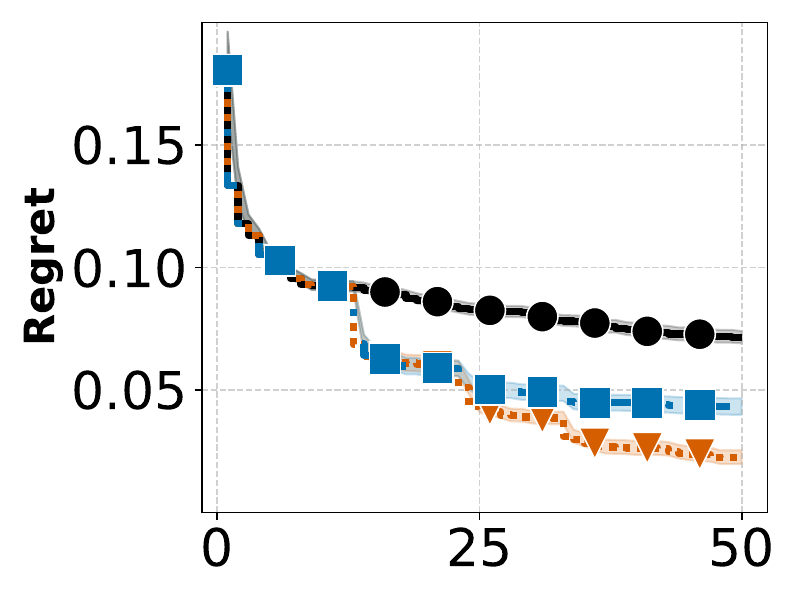} &
    \includegraphics[height=\figheight,trim={3.25cm 1.45cm 0cm 0.35cm},clip]{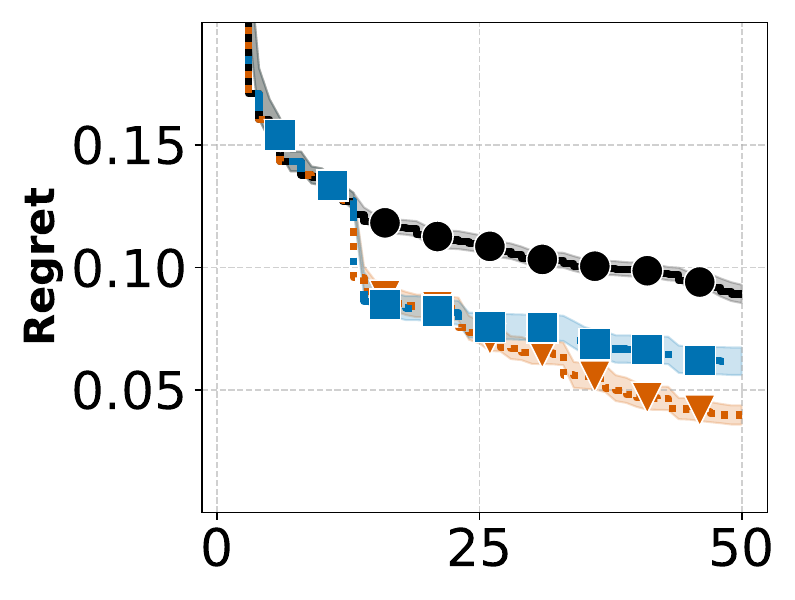} &
    \\

    \rotatebox{90}{\hspace{.2cm}\texttt{Advanced}} &
    \includegraphics[height=\figheight,trim={0.35cm 1.45cm 0cm 0.35cm},clip]{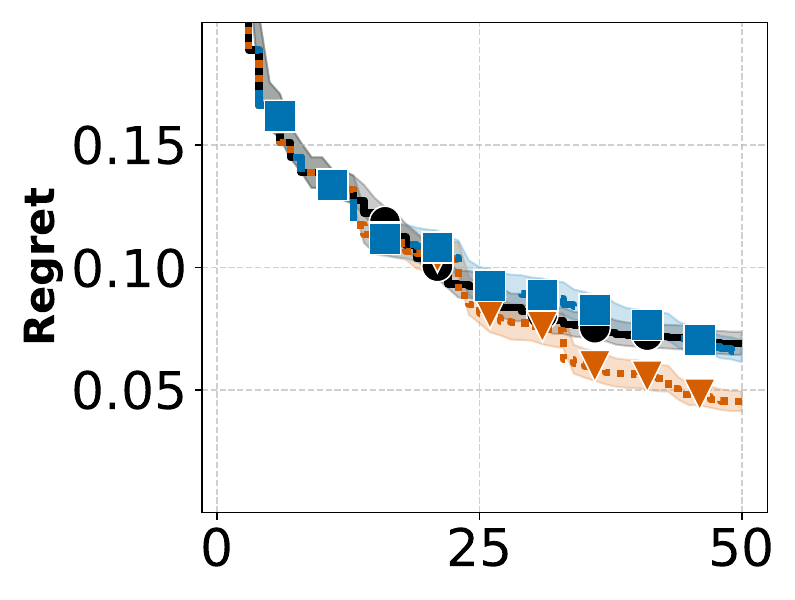} &
    \includegraphics[height=\figheight,trim={3.25cm 1.45cm 0cm 0.35cm},clip]{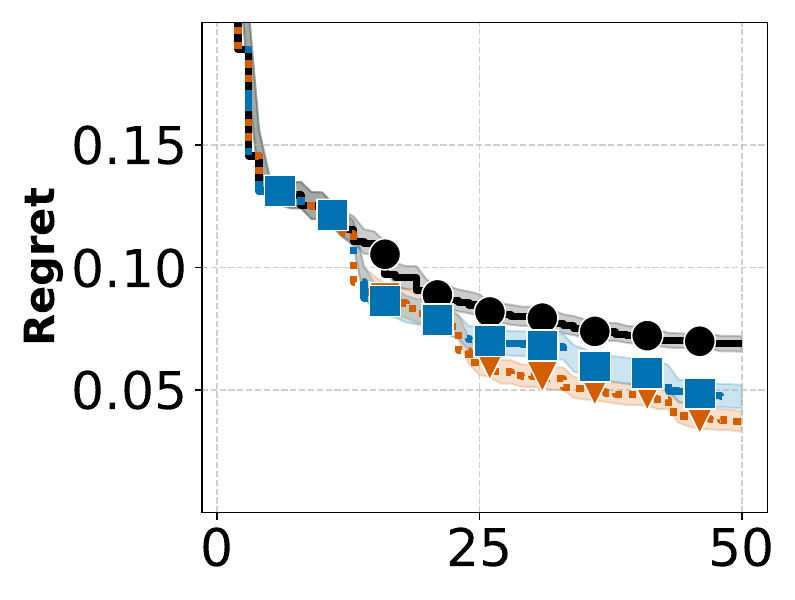} &
    \includegraphics[height=\figheight,trim={3.25cm 1.45cm 0cm 0.35cm},clip]{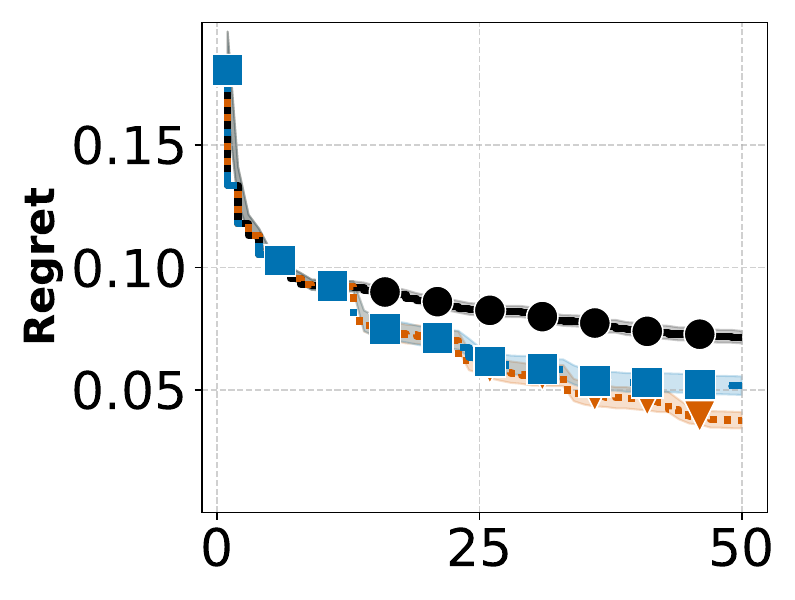} &
    \includegraphics[height=\figheight,trim={3.25cm 1.45cm 0cm 0.35cm},clip]{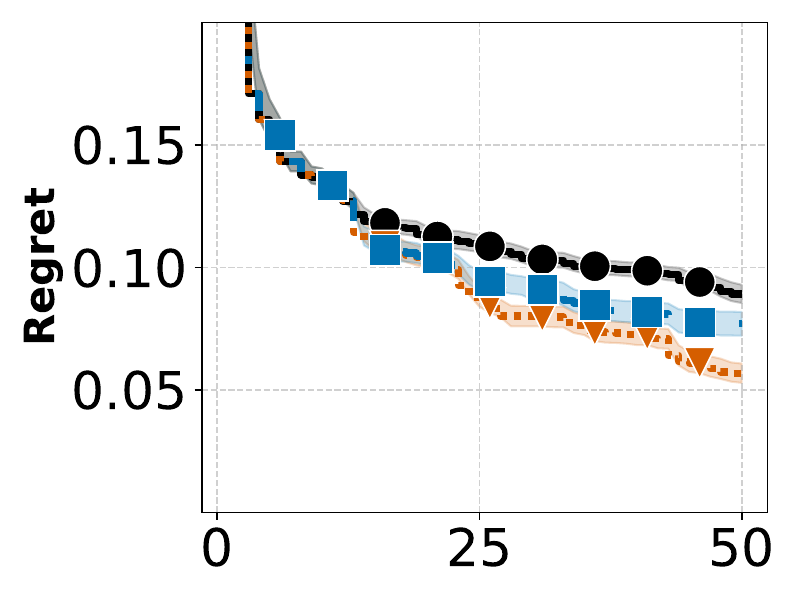} &
    \\

    \rotatebox{90}{\hspace{.75cm}\texttt{Local}} &
    \includegraphics[height=\figheight,trim={0.35cm 1.45cm 0cm 0.35cm},clip]{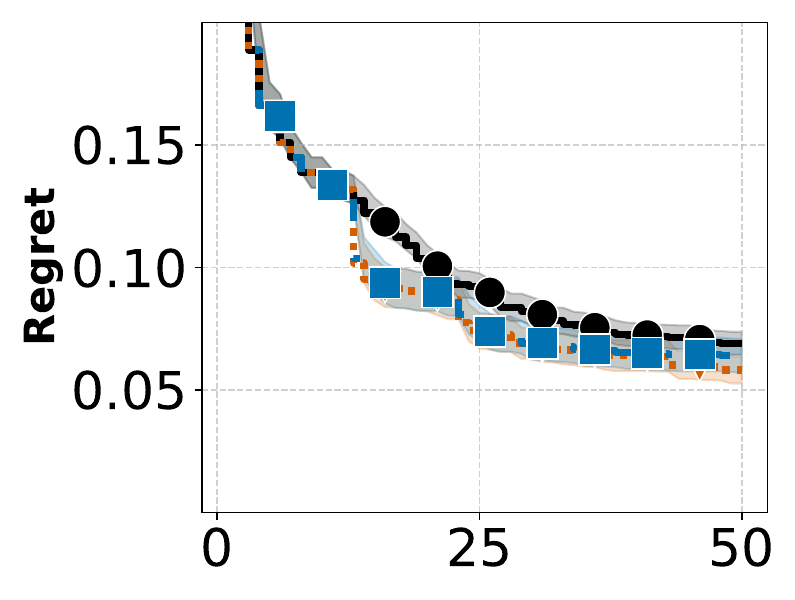} &
    \includegraphics[height=\figheight,trim={3.25cm 1.45cm 0cm 0.35cm},clip]{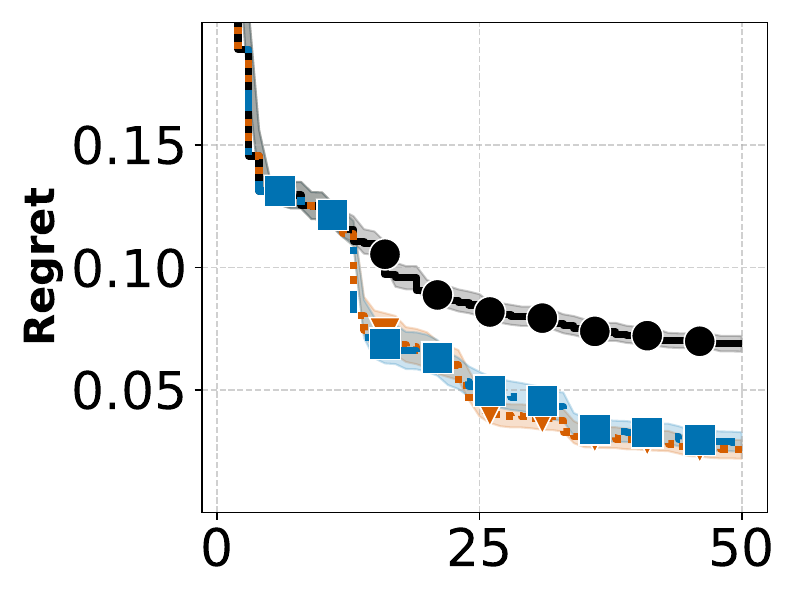} &
    \includegraphics[height=\figheight,trim={3.25cm 1.45cm 0cm 0.35cm},clip]{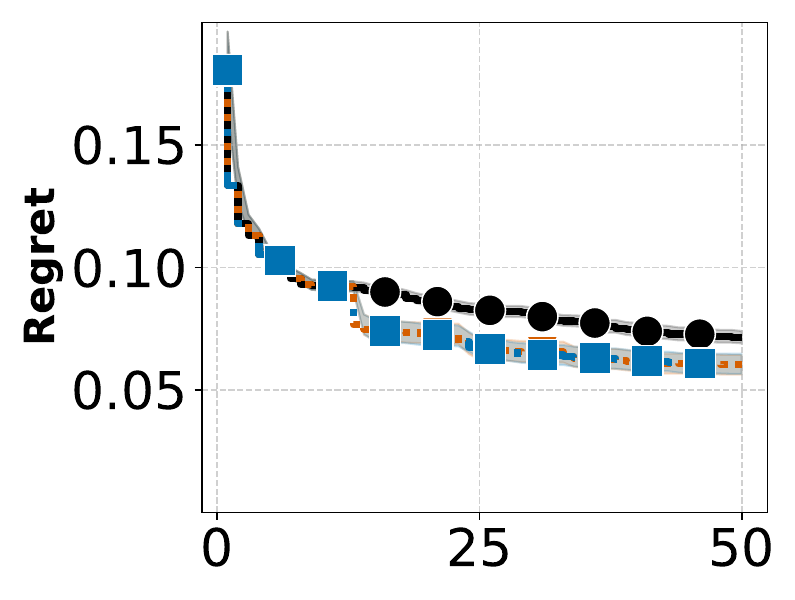} &
    \includegraphics[height=\figheight,trim={3.25cm 1.45cm 0cm 0.35cm},clip]{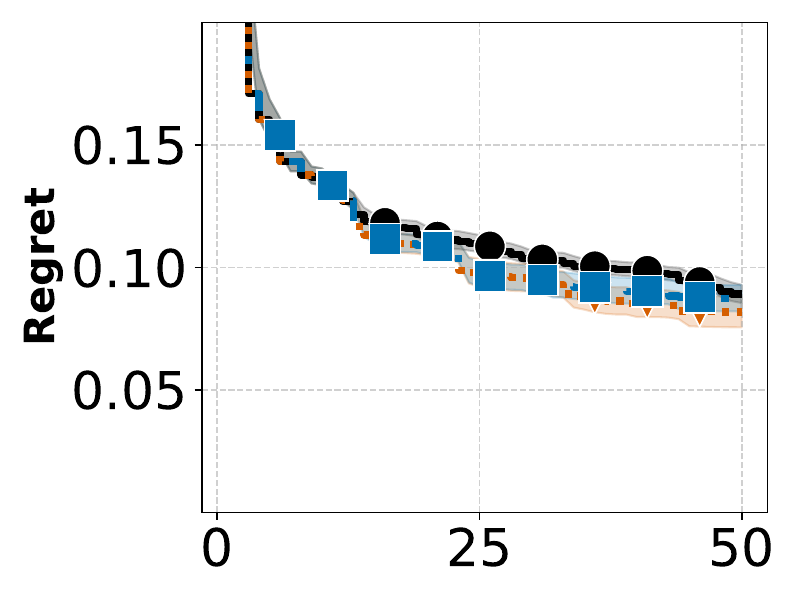} &
    \\

    \rotatebox{90}{\hspace{.3cm}\texttt{Deceptive}} &
    \includegraphics[height=\figheight+0.13\figheight,trim={0.35cm 0.35cm 0cm 0.35cm},clip]{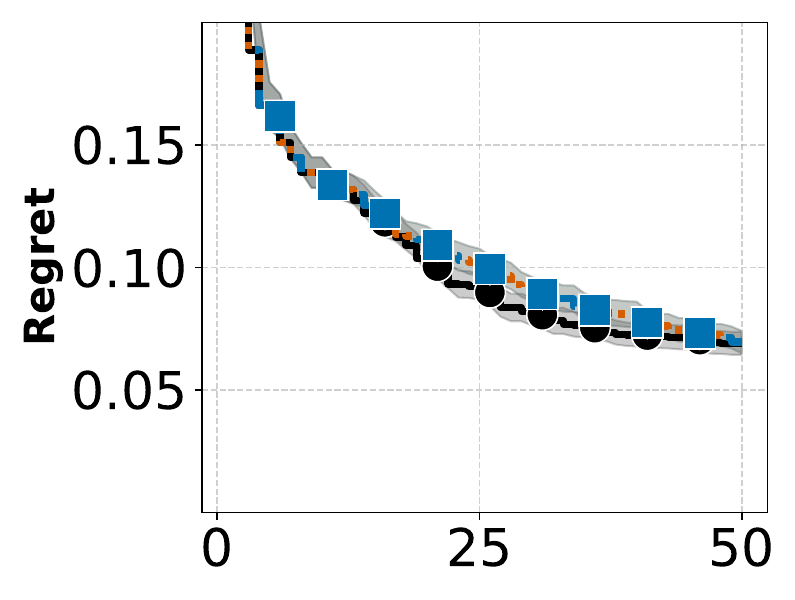} &
    \includegraphics[height=\figheight+0.13\figheight,trim={3.25cm 0.35cm 0cm 0.35cm},clip]{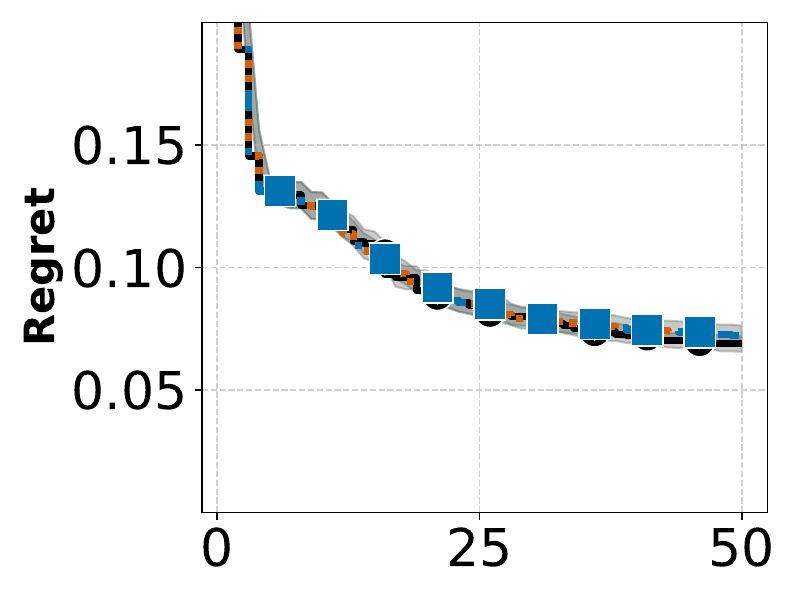} &
    \includegraphics[height=\figheight+0.13\figheight,trim={3.25cm 0.35cm 0cm 0.35cm},clip]{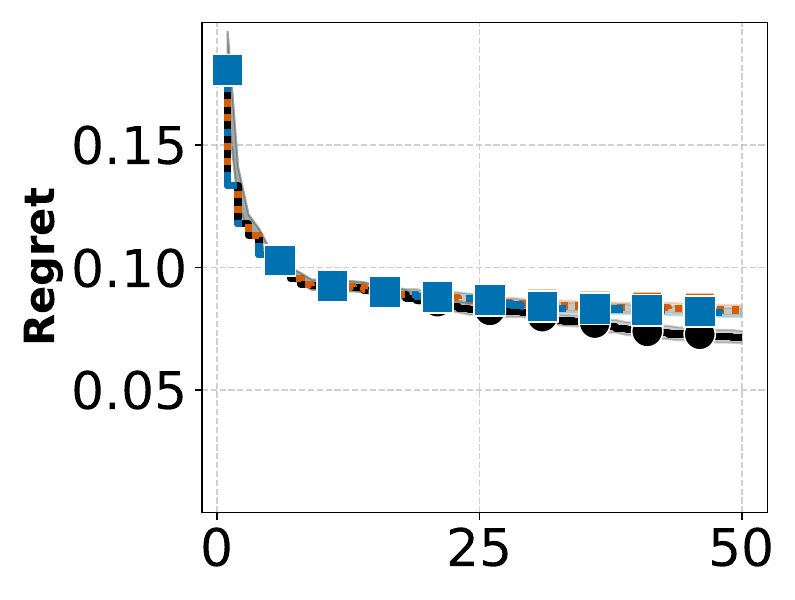} &
    \includegraphics[height=\figheight+0.13\figheight,trim={3.25cm 0.35cm 0cm 0.35cm},clip]{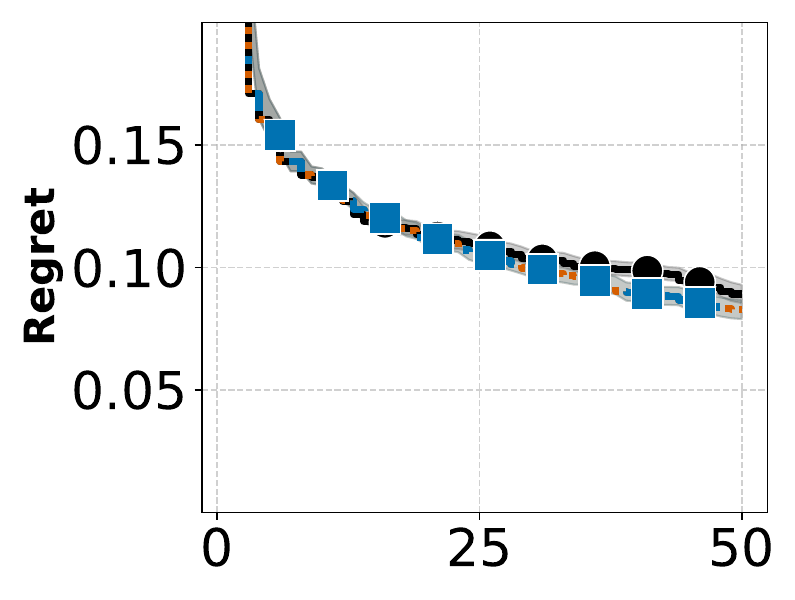} &
  \end{tabular}}

  \begin{minipage}{\textwidth}
    \centering
    \vspace{-.2cm}
    \includegraphics[width=.3\textwidth,trim={3cm 0cm .3cm 14cm},clip]{figures/iclr_submission/final_results/number_of_evaluations.pdf}
  \end{minipage}

    \begin{minipage}{\textwidth}
    \centering
    \includegraphics[width=.5\textwidth,trim={0cm 0cm 0cm 0cm},clip]{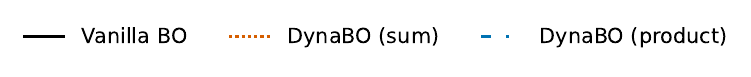}
    \end{minipage}
  \caption{Mean regret for \texttt{PD1} using Expert, Advanced, Local, and Deceptive priors comparing Vanilla BO, \tool with summed priors, and \tool with multiplied priors.}
  \label{fig:sum_vs_product}
\end{figure}

In \cref{fig:sum_vs_product}, we show the comparison of \tool with sum and product to add priors on the PD1 benchmark with the setup of the main experiments (see \cref{sec:empirical-evaluation}). The sum outperforms the product by a wide margin for informative prior kinds. For uninformative prior kinds, there are no differences. This further strengthens the arguments made in above and indicates that the sum is the better choice.

\newpage
\section{Additional \tool Details}\label{app:additional_details}
\subsection{Facilitating Prior Behavior} \label{app:facilitating_prior_behavior}

\paragraph{Adapting Candidate Sampling}
If user priors are peaked in a small area of the configuration space, the resulting acquisition function $\af_\text{dyna}$ could be harder to optimize, see \cref{eq:acq_opt}.
To ensure that the acquisition function optimizer covers both user-suggested regions and generally unexplored areas of the configuration space adequately, we adopt a modified version of \citet{hutter-lion11a}'s combined local and random search. 
They propose sampling a mix of candidate configurations in the vicinity of the best previously found configuration, and additional candidates are drawn from more distant regions of the configuration space. The highest potential candidates, with respect to the acquisition function $\af$, serve as a starting point for hill climbing. 
To ensure that a sufficient number of candidates are sampled close to the peaked prior, we adapt the sampling of the starting points of the local search to the prior distributions. To this end, each prior $\pi^{(m)}$ is assigned a weight $\omega_m = e^{\phi(t-t^{(m)})}$, which also decays with time. Then, a fraction $\omega_m \cdot \min\{\sum_{i=1}^m \omega_i, 0.9\}$ of candidate configurations are sampled according to prior $\pi^{(m)}$.\\

Suppose a finite sequence of user-specified priors $\{\pi^{(m)}\}_{m=1}^{M}$ is provided at times $\{t^{(m)}\}_{m=1}^{M}$, with $t^{(1)}<...<t^{(M)}\leq T$, then the random configurations are replaced as follows:
\renewcommand{\floor}[1]{\lfloor #1 \rfloor}
\begin{itemize}
    \item At iteration $t$ prior $\pi^{(m)}$ is associated with a weight $\omega_m = e^{-0.126\cdot (t - t^{(m)})}$.
    \item If $\sum\limits_{m=1}^M  \omega_m\leq 0.9$, $\pi^m$ is used to sample $\floor{\omega_m \cdot 5000 + 0.5}$ configurations. The rest are sampled uniformly at random.
    \item If $\sum\limits_{m=1}^M \omega_m > 0.9$,  $\pi^m$ is used to sample 
    $\left\lfloor \dfrac{\omega_m}{\sum_{j=1}^M \omega_j} \cdot 5000 + 0.5\right\rfloor$
    configurations. The rest are sampled uniformly at random.
\end{itemize}

\paragraph{Numerical Stability}
To ensure numerical stability and to maintain the impact of the initial acquisition function, we clip priors at 1e-12 and thereby ensure that priors take values in $\mathbb{R}^+$. Furthermore, due to the decaying mechanism, priors converge to $1$ with increasing $t$.

\subsection{Further Details on the Rejection Criterion}\label{app:further_details_prior_rejection}
Following the principle of optimism in the face of uncertainty, we assess the potential of both regions in terms of their lower confidence bounds (LCBs) \citep{agrawal-aap1995a}:
\begin{equation*}
    LCB(\lambda) = (-1) \cdot (\mu(\lambda) - \kappa \sigma(\lambda))\, ,
\end{equation*}
where $\mu(\cdot)$ denotes the mean predicted with $\surrogate$ for $\conf$, and $\sigma(\cdot)$ the uncertainty in terms of standard deviation. This way, we interpret all uncertainty the surrogate model may have at a configuration $\conf$ as the potential for its performance to achieve an improvement at all. Intuitively, the LCB criterion allows us to quantify the explorative potential of the prior region as opposed to the exploitative potential close to the current incumbent, since uncertainty close to the incumbent is typically low.

Provided a prior $\pi^{(m)}$ with mean $\mu^{(m)}$, and standard deviation $\sigma^{(m)}$ as 
in \cref{eq:prior_construction}, our initialization of the rejection criterion
\begin{equation}
\E_{\conf \sim \mathcal{F}_{\prior^{(m)}}} \left[ \xi_{\surrogate}(\lambda) \right] - \E_{\conf \sim \mathcal{N}_{\inc}}  \left[ \xi_{\surrogate}(\conf) \right] \geq \tau \,\, 
\end{equation}
utilizes $500$ configurations from normal distributions for both. The quality of the prior $\mathcal{F}_{\prior^{(m)}}$is assessed using $\mathcal{N}_{\prior^{(m)}}\sim(\mu^{(m)},\sigma^{(m)})$. The quality of the area around the incumbent $\inc$ is assessed using a normal distribution with the incumbent in the center, and the provided priors standard deviation $\mathcal{N}_{\inc} \sim (\inc, \sigma^{(m)})$. Our main experiments utilize $\tau=-0.15$.

\paragraph{Categorical Hyperparameters}
To apply our prior rejection scheme on categorical hyperparameters, we conduct the following adaptations: For the configurations sampled according to the prior $\conf \sim \mathcal{N}_{\prior^{(m)}}$, categorical values are sampled according to the provided weights. For the configurations sampled in the area around the incumbent $\conf^+\sim \mathcal{F}_{\inc}$, the incumbent's configuration is utilized.

\newpage

\section{Artificial Prior Generation}\label{app:priors}
In the following paragraphs, we discuss additional details of our benchmarking setup with a focus on the generation of artificial priors. Firstly, in \cref{app:different_prior_kinds} we discuss the four different kinds of priors considered here. In \cref{app:data-generation} we discuss how we generate the data used for the artificial priors. In \cref{app:prior_construction} we discuss how priors are then constructed. For simplicity, we refer to $\perf{\conf}$ as $\perf_{\conf}$ here.
 
\subsection{Different Prior Kinds}\label{app:different_prior_kinds}
Inspired by the evaluation protocols of \citet{souza-ecml21a,hvarfner-iclr22a, mallik-neurips23a}, and \citet{seng-automlconf25a}, we construct artificial, data-driven priors. To ground our investigation in an analysis of well and poorly performing areas of the configuration space, we conduct an extensive search for every benchmark scenario and cluster the found configuration-loss pairs $\inctuple{}:=(\conf, f(\conf))$ hierarchically via Gower's distance \citep{gower-biometrics1971a} into $n$ clusters. For each of the $n$ clusters, $c_1,\ldots,c_n$, we compute its centroid $\overline{c_i}$ and the median loss $\overline{\perf_{c_i}}$ of configurations contained. In the following, we assume the clusters to be ordered according to their median loss, that is, $\overline{\perf_{c_i}} \leq \overline{\perf_{c_j}}$ for $ i<j$. 

To obtain dynamic priors, considering the current incumbent $\inc$ and its loss $\perf_{\inc}$, we select the cluster $c^+$ and configuration $\conf^+$ according to the following four prior policies simulating different aspects and levels of informativeness. The chosen configuration $\conf^+$ is then used as the center of a normal distribution over the configuration space to guide the optimization process toward its broader region.

\textbf{Expert Priors} bias \tool toward clusters spanning significantly-better regions of the configuration space, that is, $\perf_{\inc} \geq \overline{\perf_{c_i}}$. A cluster $c^+$ is sampled with probability $\mathbb{P}(c_i) \propto e^{0.1 \cdot i}$. From this cluster, we choose the best configuration $\conf^+\in\arg\min_{\conf\in c^+}\perf_\conf$ as prior center.
        
\textbf{Advanced Priors}\,\, bias \tool toward clusters spanning better-performing regions of the configuration space, that is, $\perf_{\inc} \geq \overline{\perf_{c_i}}$. A cluster $c^+$ is sampled with probability $\mathbb{P}(c_i) \propto e^{0.15 \cdot i}$. From this cluster, we sample a configuration $\conf^+\in c^+$ randomly as the prior center. 

\textbf{Local Priors}\,\, bias \tool toward well-performing clusters close to the current incumbent. To this end, the incumbents' Gower's distance \citep{gower-biometrics1971a} to each cluster $D^{gower}(\inc, \overline{c_i})$ is utilized to select the $10$ closest, later considered clusters $C^+$. The cluster with the lowest median loss $c^+ \in \argmin_{c \in C^+} \overline{\perf_c}$ is selected, and the prior center $\lambda^+ \in c^+$ is sampled randomly. 

\textbf{Deceptive Priors}  bias \tool toward sampling configurations in poorly performing regions of the configuration space. For that, $c^+$ is randomly sampled from the five clusters with the worst median loss. The center of the prior is set to $\lambda^+ \in \argmax_{\conf \in c^+} \perf_\conf$. This is meant to simulate the worst case in which a human user provides priors based on wrong assumptions.

\subsection{Data Generation Runs}
\label{app:data-generation}

As mentioned in \cref{sec:hyperparameter_optimization}, HPO aims to find a well-performing hyperparameter configuration $\inc$ according to the cost function $f$.

As discussed in \cref{sec:prior-construct}, we construct priors based on data collected through Bayesian optimization runs. These data generation runs are conducted as follows:

\begin{enumerate}
    \item Generate prior data: For each learner $A$, dataset $D$ combination, execute explorative Bayesian optimization runs with both the more greedy Expected Improvement (EI) and more explorative Lower Confidence Bounds (LCB) \citep{papenmeier-corr25a}. For each acquisition function, run 10 seeds for a budget of $5,000$ iterations. Then, for every algorithm, dataset combination, concatenate the lists of preliminary incumbents and assemble a joint list sorted by losses $f_\lambda$:
    \[
        I_{A, D} = [\inctuple{1}, \inctuple{2}, \ldots ,  \inctuple{n}].
    \]
    \item To ensure that also non-well-performing areas of the configuration space are covered, $I_{A, D}$ is supplemented with $n$ non-incumbent configurations and their loss.
    \item Due to structured configuration spaces, some hyperparameters may not be active. In the case of our experiments, this only occurs for numeric hyperparameters. These values are filled with $-1$.
    \item Create clusters of configurations of preliminary incumbents in the configuration space: For each learner $A$, dataset $D$ combination, cluster the incumbent configurations into $100$ clusters 
    \[
        C_{A,D} = \{c_1, c_2,...,c_{100}\} \quad \text{with} \quad c_i=\{\inctuple{c_i^1}, \inctuple{c_i^2},...\}
    \]
    using Agglomerative Clustering with Gower's Distance \citep{gower-biometrics1971a} and Ward Linkage. For each cluster, compute a centroid $\overline{c_i}$ and the median performance $\overline{\perf_{c_i}}$.
\end{enumerate}

\subsection{Prior Construction}\label{app:prior_construction}

During optimization of $A$ on dataset $D$, priors are generated dynamically.
\begin{enumerate}
    \item Sample prior configuration $\lambda^+$
    \item Build prior: We hypothesize that with each prior provided to \tool, the confidence of a user would grow. In our synthetic prior generation, we therefore build the $k$-th prior $\prior^k$ as follows: For each numerical hyperparameter $\lambda_j$ with lower bounds $\lambda_1^l, \lambda_2^l, ..., \lambda_d^l$ and upper bounds $\lambda_1^u, \lambda_2^u, ..., \lambda_d^u$, we set
    \begin{equation}
        \prior^k = [\mu_j, \sigma_j]_{j=1}^{d} = \bigg[\bigg(\lambda^+_j, \frac{|\lambda_j^u - \lambda_j^l|}{k \cdot 5}\bigg)\bigg]_{j=1}^{d}. \label{eq:prior_construction}
    \end{equation}
    \item For each categorical hyperparameter $\lambda_j = \lambda^+_j$.
\end{enumerate}

As mentioned in \cref{sec:experiment-setup}, we provide four priors for evaluations on \yahpo, and four priors $\prior^1, \prior^2, \prior^3, \prior^4$ for evaluations on PD1, respectively.

\

\newpage

\section{Detailed Experimental Setup}\label{app:detailed_experimental_setup}
The implementation of \tool  is available at \href{https://github.com/automl/DynaBO}{https://github.com/automl/DynaBO}.\\
The adapted implementation of PCs is available at \href{https://github.com/LUH-AI/ibo-hpc}{https://github.com/LUH-AI/ibo-hpc}.

\subsection{Summary of the Benchmark}\label{app:detailed_experimental_setup:benchmarks}
Rather than training many models with different hyperparameter configurations, we use surrogate models provided by \citep{pfisterer-automl22a} for XGBoost (\texttt{xgboost}) \citep{chen-kdd16a} and multi-layer perceptrons (\texttt{lcbench}) dubbed traditional machine learning. For complex architectures, we utilize surrogates trained for \citet{mallik-neurips23a} based on data collected by \citet{wang-jmlr24}. We consider a wide ResNet \citep{he-cvpr16a} (\texttt{widernet}) trained on CIFAR100 \citep{krizhevsky-tech09a}, a ResNet \citep{he-cvpr16a} (\texttt{resnet}) trained on ImageNet \citep{deng-cvpr09a}, a transformer \citep{vaswani-neurips17a} (\texttt{transf}) trained on LM1B \citep{chelba-is13a}, and a transformer (\texttt{xformer}) trained on WMT15 \citep{bojar-acl15a}. 
 A learner, searchspace, and dataset overview is provided in \cref{tab:beautiful_rbv2_lcbench}.
\definecolor{headerblue}{RGB}{70,130,180}
\definecolor{headertext}{RGB}{255,255,255}
\definecolor{rowgray}{gray}{0.95}

\begin{table}[h]
    \centering
    \caption{An overview of the evaluated \textit{scenarios}, each with the considered configuration \textit{configuration space} type and number of \textit{datasets}, with which each scenario was evaluated.}
    \label{tab:beautiful_rbv2_lcbench}
    \renewcommand{\arraystretch}{1.3} 
    \setlength{\tabcolsep}{12pt}      
    \begin{tabular}{>{\raggedright\arraybackslash}p{4cm} >{\raggedright\arraybackslash}p{4cm} >{\raggedleft\arraybackslash}p{2cm}}
        \rowcolor{headerblue} 
        \textbf{\textcolor{white}{Scenario}} & \textbf{\textcolor{white}{Configuration Space}} & \textbf{\textcolor{white}{\# Datasets}} \\
        \rowcolor{rowgray} rbv2\_xgboost & 14D: Mixed & 119 \\
        lcbench & 7D: Numeric & 34 \\
        \rowcolor{rowgray} cifar100\_wideresnet\_2048 & 4D: Numeric & 1 \\
        imagenet\_resnet\_512 & 4D: Numeric & 1\\
        \rowcolor{rowgray} lm1b\_transformer\_2048 & 4D: Numeric & 1\\
        translatewmt\_xformer\_64 & 4D: Numeric & 1 \\
        \bottomrule
    \end{tabular}
\end{table}

Our experiments are scheduled, and the results are logged in a MySQL database using the PyExperimenter library \citep{tornede-jair23a}. 

\subsection{Cluster Setup}\label{app:detailed_experimental_setup:cluster_setup}
All experiments discussed in this paper were executed on HPC nodes equipped with 2 Intel(R) Xeon(R) Platinum 8470 @2.0GHz processors and 488GiB RAM, of which 2 CPU cores and 6GB RAM were allocated per run.

\subsection{Competitor Setup}\label{app:detailed_experimental_setup:competitors}
Our comparisons focus on vanilla-BO, \pibo, and PCs. We chose to disregard BoPro \citep{souza-ecml21a} as a baseline, since it is dominated substantially by \pibo.

\subsubsection{\tool, \pibo and Vanilla-BO}
Our experiments are built on top of SMAC3 \citep{lindauer-jmlr22a}, for vanilla BO, \pibo, and \tool. We use the Hyperparameter Optimization Facade but deactivate its default $\log$ transformations. We also refit the surrogate after every evaluated configuration.

\subsubsection{Probabilistic Circuits}
We base our setup of \citet{seng-automlconf25a}'s probabilistic circuits on their source code and the provided paper. All adaptations made were discussed with \citet{seng-automlconf25a} and are discussed below.

\paragraph{Initial Design} To facilitate a fair comparison, we adapted \citet{seng-automlconf25a}'s source code to use an initial design of the same size as for vanilla-BO, \pibo, and \tool. As the authors actively chose against a Sobol initial design, we do not adapt it.

\paragraph{Prior design}
For distribution priors, we utilize priors as described in \cref{app:different_prior_kinds}. For experiments with pointwise priors, which provide PCs with an unfair advantage for informative priors, we utilize the sampled prior center as a pointwise prior.

\paragraph{Applying Conditions on a Randomly Sampled Subset of Hyperparameters} The initial and adapted methodology can be seen in \cref{pc:original}, and \cref{pc:adapted} respectively. Comments as well as changes added by us are marked blue. Our modifications are motivated by the observation that if provided with priors for all hyperparameters, the PC is not used to create configurations (refer to line 11 of \cref{pc:original}). Here, configurations are sampled for all hyperparameters without a prior. However, since our prior setup sets priors for all hyperparameters, we randomly mask the prior for some hyperparameters (refer to lines 10 and 12 of \cref{pc:adapted}), allowing us to utilize the PC and the provided information. We also experimented with sampling k for each preliminary condition individually, which performed slightly worse.

\begin{algorithm}[H]
    \begin{algorithmic}[1]
    \STATE \textbf{Input:} Search space $\mathbf{\Theta}$ over $\bm{\mathcal{H}} = \{H_1, \dots, H_n\}$, problem instance $\mathbf{x} \in \mathcal{X}$, initial prior distribution $u(\bm{\mathcal{H}})$, objective $f: \mathbf{\Theta} \times \bm{\mathcal{X}} \rightarrow \mathbb{R}$, user prior $q(\hat{\bm{\mathcal{H}}})$ (optional and can be provided at any time), decay $\gamma$
    \STATE Sample $J$ configurations $\bm{\theta} \sim u(\bm{\mathcal{H}})$
    \STATE $\bm{\mathcal{D}} \gets \{(\bm{\theta}_i, f(\bm{\theta}_i; \mathbf{x}) \}$ for $i \in \{1, ..., J\}$
    \WHILE{not converged}
        \STATE Fit HPC $s$ on $\bm{\mathcal{D}}$ every $L$-th iteration
        \STATE Set $f^* \gets \max_f \bm{\mathcal{D}}$ and $b \sim \text{Ber}(\rho)$
        \IF{prior $q(\hat{\bm{\mathcal{H}}})$ is given and $b = 1$}
            \STATE Sample $N$ conditions $\bm{\theta} \sim q(\hat{\bm{\mathcal{H}}})$  \change{\COMMENT{One prior for point priors}}
            \STATE $\mathbf{C} \gets \emptyset$
            \FOR{condition $\bm{\theta}_i$ in $\bm{\theta}$ \change{\COMMENT{Contains only one element for point priors.}} }
                \STATE Sample $\bm{\theta}'_{1, \dots, B} \sim s(\bm{\mathcal{H}} \setminus \hat{\bm{\mathcal{H}}} | \hat{\bm{\mathcal{H}}}, f^*)$ 
                \change{\COMMENT{Sample values for hyperparameters where no prior is provided.}}
                \STATE $\bm{\theta}^*_i \gets \arg \max_{\bm{\theta}' \in \bm{\theta}'_{1, \dots, B}} s(\bm{\theta}' | f^*)$
                \STATE $\mathbf{C} \gets \mathbf{C} \cup \bm{\theta}^*_i$
            \ENDFOR
         \STATE $\bm{\theta}^* \sim \mathcal{U}(\mathbf{C})$
        \ELSE
        \STATE $\bm{\theta}^* \sim s(\bm{\mathcal{H}} | f^*)$
        \ENDIF
        \STATE set $\bm{\mathcal{D}} \gets \bm{\mathcal{D}} \cup \{(\bm{\theta}', f(\bm{\theta}', \mathbf{x}))\}$ and $\rho \gets \gamma \cdot \rho$
    \ENDWHILE
    \end{algorithmic}
    \caption{Optimize with PCs}
    \label{pc:original}
\end{algorithm}

\begin{algorithm}[H]
    \begin{algorithmic}[1]
    \STATE \textbf{Input:} Search space $\mathbf{\Theta}$ over $\bm{\mathcal{H}} = \{H_1, \dots, H_n\}$, problem instance $\mathbf{x} \in \mathcal{X}$, initial prior distribution $u(\bm{\mathcal{H}})$, objective $f: \mathbf{\Theta} \times \bm{\mathcal{X}} \rightarrow \mathbb{R}$, user prior $q(\hat{\bm{\mathcal{H}}})$ (optional and can be provided at any time), decay $\gamma$
    \STATE Sample $J$ configurations $\bm{\theta} \sim u(\bm{\mathcal{H}})$
    \STATE $\bm{\mathcal{D}} \gets \{(\bm{\theta}_i, f(\bm{\theta}_i; \mathbf{x}) \}$ for $i \in \{1, ..., J\}$
    \WHILE{not converged}
        \STATE Fit HPC $s$ on $\bm{\mathcal{D}}$ every $L$-th iteration
        \STATE Set $f^* \gets \max_f \bm{\mathcal{D}}$ and $b \sim \text{Ber}(\rho)$
        \IF{prior $q(\hat{\bm{\mathcal{H}}})$ is given and $b = 1$}
            \STATE Sample $N$ conditions $\bm{\theta} \sim q(\hat{\bm{\mathcal{H}}})$  \change{\COMMENT{One prior for point priors}}
            \STATE $\mathbf{C} \gets \emptyset$
            \STATE \change{$k \sim \mathcal{U}(1,...,n-1)$ \COMMENT{Sample how many dimensions are used for the prior} }
            \FOR{\change{preliminary condition $\bm{\theta}_i^p$} in $\bm{\theta}$} 
                \STATE \change{For k hyperparameters set $\bm{\theta}_i= \bm{\theta}_i^p$ \COMMENT{Set prior on subset of hyperparameters}  }
                \STATE Sample $\bm{\theta}'_{1, \dots, B} \sim s(\bm{\mathcal{H}} \setminus \hat{\bm{\mathcal{H}}} | \hat{\bm{\mathcal{H}}}, f^*)$ 
                \STATE $\bm{\theta}^*_i \gets \arg \max_{\bm{\theta}' \in \bm{\theta}'_{1, \dots, B}} s(\bm{\theta}' | f^*)$
                \STATE $\mathbf{C} \gets \mathbf{C} \cup \bm{\theta}^*_i$
            \ENDFOR
         \STATE $\bm{\theta}^* \sim \mathcal{U}(\mathbf{C})$
        \ELSE
        \STATE $\bm{\theta}^* \sim s(\bm{\mathcal{H}} | f^*)$
        \ENDIF
        \STATE set $\bm{\mathcal{D}} \gets \bm{\mathcal{D}} \cup \{(\bm{\theta}', f(\bm{\theta}', \mathbf{x}))\}$ and $\rho \gets \gamma \cdot \rho$
    \ENDWHILE
    \end{algorithmic}
    \caption{Optimize with PCs - Adapted}
    \label{pc:adapted}
\end{algorithm}

\newpage

\section{Additional Empirical Results} \label{app:further-results}
Our additional experimental results focus on further validating our approach. \cref{app:further-results:increased-budget} contains the detailed results for an increased budget on the PD1 benchmark \citep{wang-jmlr24} with random forests and Gaussian processes as a surrogate, respectively. \cref{app:further-results:lcb_results} contains the results obtained with LCB as the acquisition function. \cref{app:further-results:rejection_ablation} contains the scenario-wise results for the rejection sensitivity analysis. \cref{app:further-results:pc} contains the detailed results of a comparison with probabilistic circuits. \cref{app:further-results:gp-normal} contains the results of our comparison to \pibo with GPs as a surrogate model. For an easier comparison, the corresponding results with random forests are also supplied. \cref{app:further-results:dynamic} contains the results of experiments conducted with randomly sampled prior locations. \cref{pibo_extension} describes a na{\"i}ve dynamic extension for \pibo and reports its results. \cref{app:further-results:prior-decay} contains a small sensitivity analysis for the speed of prior decay. Finally, \cref{app:further-results:rior_rejection_sampling_budget_ablation} provides an ablation over the prior rejection sampling budget.

\subsection{Increased Budget Results}\label{app:further-results:increased-budget}
In our experiments with an increased budget, we introduce a prior every ten trials following the initial design. The results indicate that random forests perform poorly without prior rejection, whereas Gaussian processes remain robust. Nevertheless, \tool with prior rejection consistently outperforms \pibo on most scenarios.

\begin{figure}[H]
    \centering
    \setlength{\tabcolsep}{6pt}
    \renewcommand{\arraystretch}{1.05}
    \begin{tabular}{ccc}
     &  \textbf{Random Forest} &  \textbf{Gaussian Process} \\
    \rotatebox{90}{\hspace{.7cm} Overall} &
    \includegraphics[width=.43\linewidth, trim={.35cm .35cm .35cm .35cm}, clip]{figures/ICML_26_submission/final_results/increased_budget/rf/overall.pdf} &
    \includegraphics[width=.43\linewidth, trim={.35cm .35cm .35cm .35cm}, clip]{figures/ICML_26_submission/final_results/increased_budget/gp/overall.pdf} \\
    \rotatebox{90}{\hspace{.7cm}\texttt{widernet}} &
    \includegraphics[width=.43\linewidth, trim={.35cm .35cm .35cm .35cm}, clip]{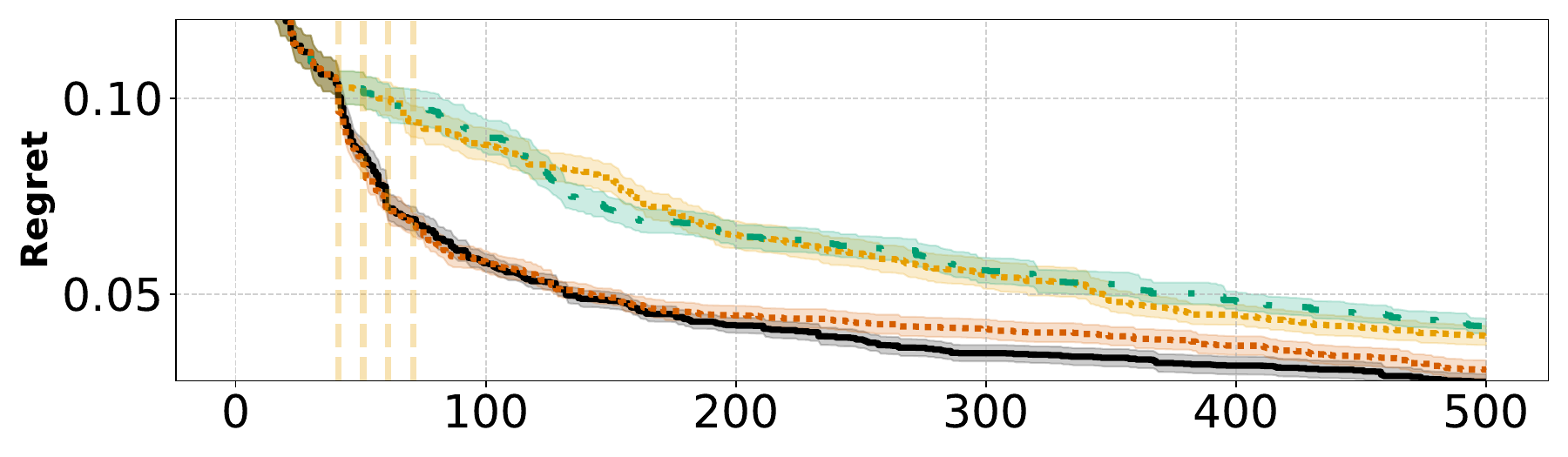} &
    \includegraphics[width=.43\linewidth, trim={.35cm .35cm .35cm .35cm}, clip]{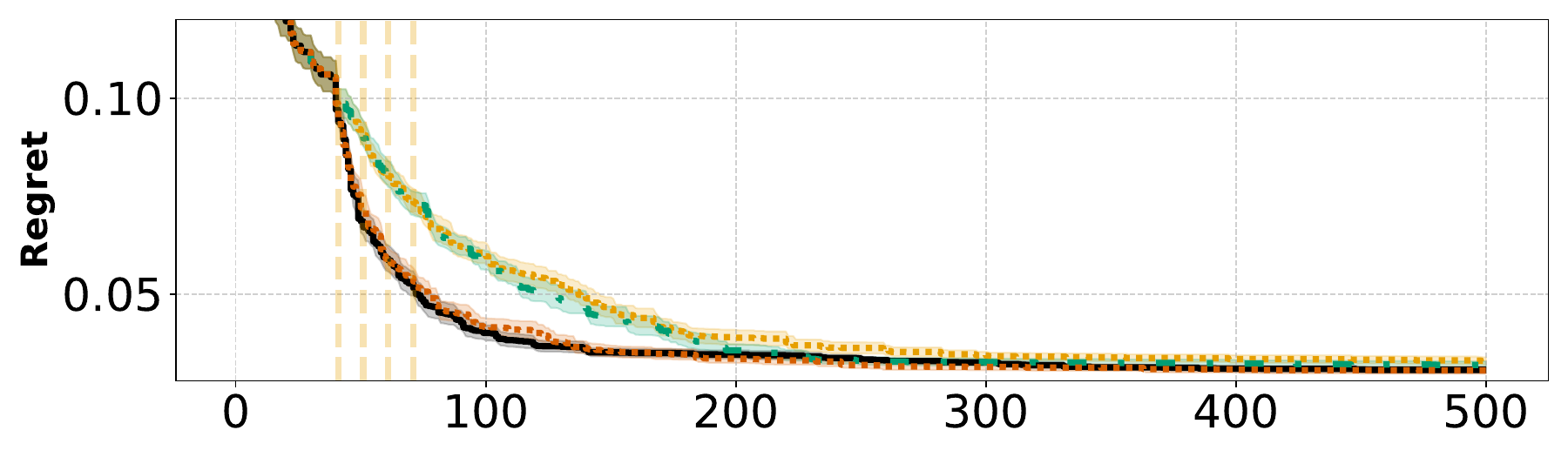} \\
    
    \rotatebox{90}{\hspace{.7cm}\texttt{resnet}} &
    \includegraphics[width=.43\linewidth, trim={.35cm .35cm .35cm .35cm}, clip]{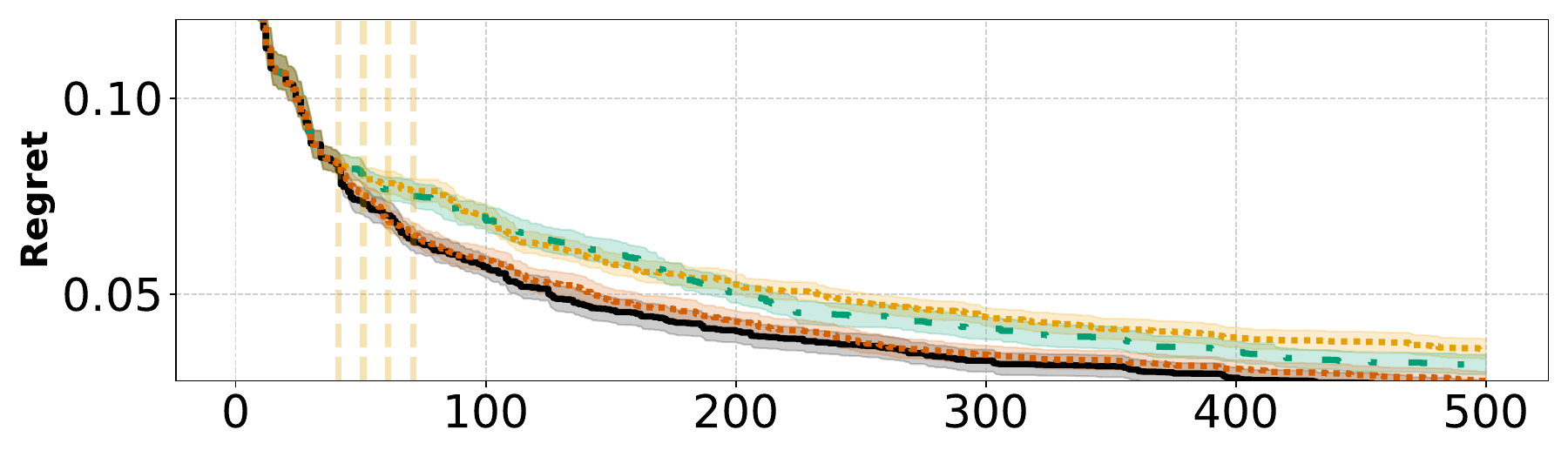} &
    \includegraphics[width=.43\linewidth, trim={.35cm .35cm .35cm .35cm}, clip]{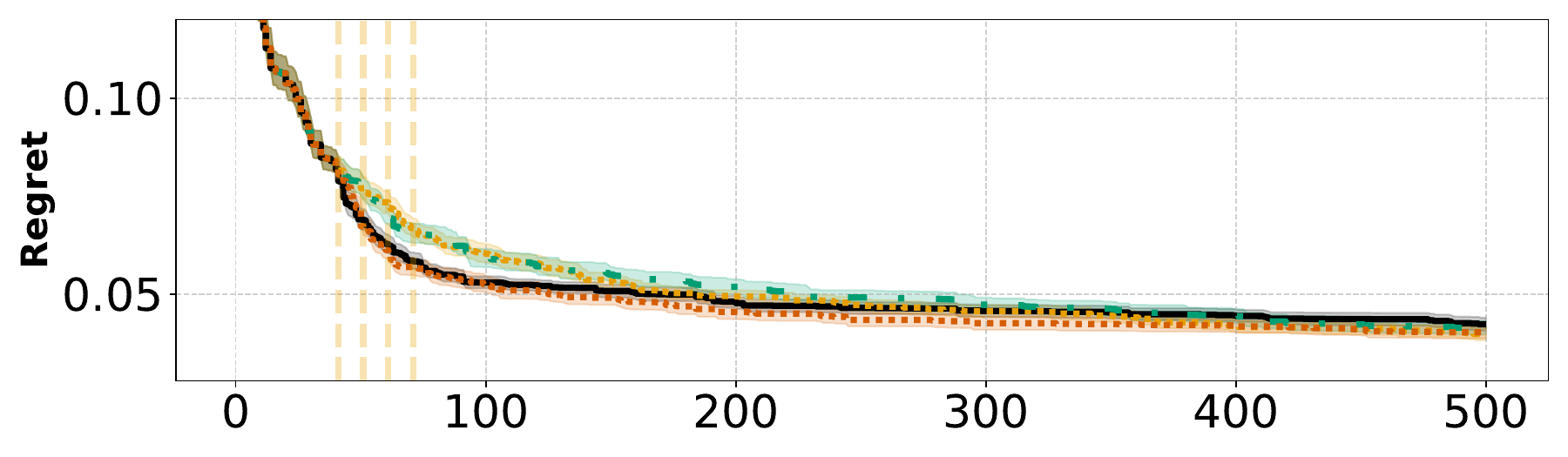} \\[6pt]
    \rotatebox{90}{\hspace{.7cm}\texttt{transf}} &
    \includegraphics[width=.43\linewidth, trim={.35cm .35cm .35cm .35cm}, clip]{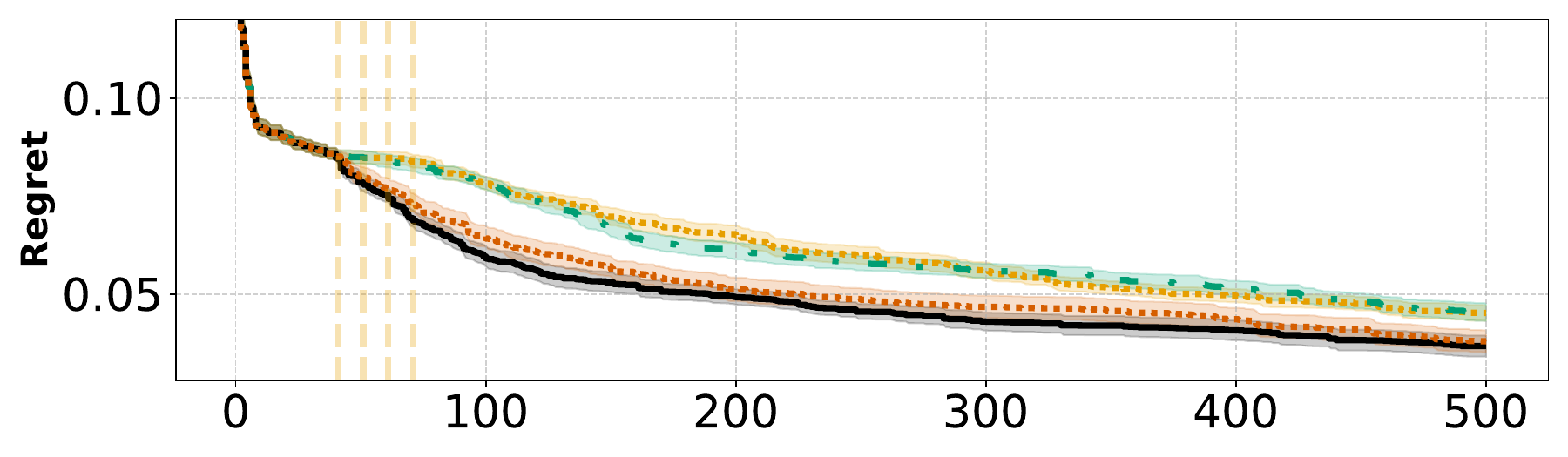} &
    \includegraphics[width=.43\linewidth, trim={.35cm .35cm .35cm .35cm}, clip]{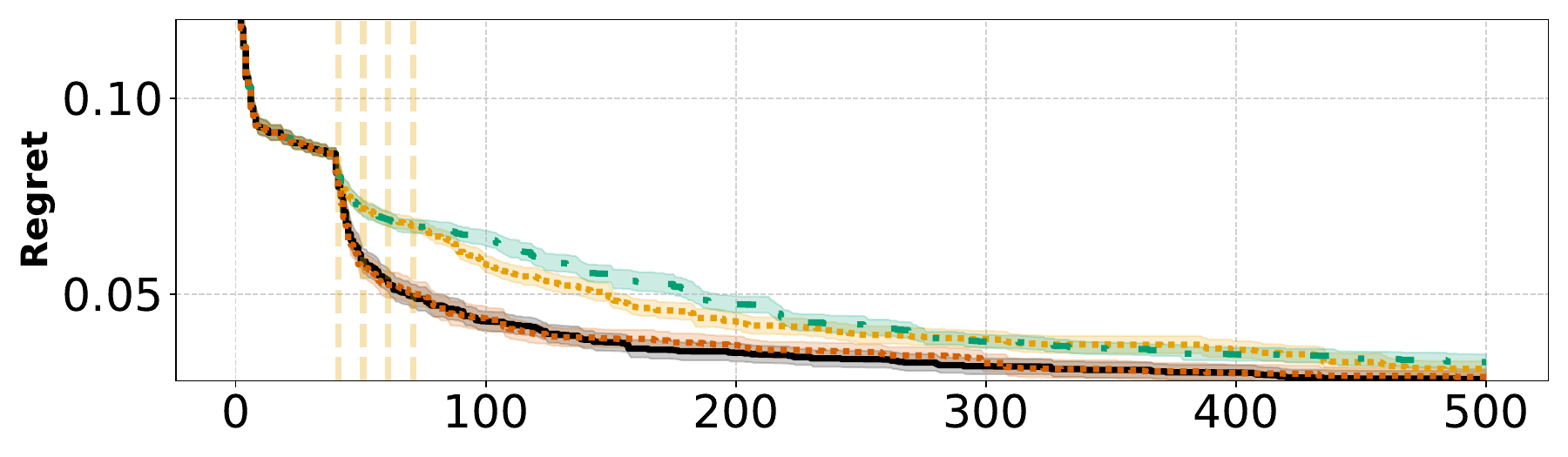} \\
    
    \rotatebox{90}{\hspace{.7cm}\texttt{xformer}} &
    \includegraphics[width=.43\linewidth, trim={.35cm .35cm .35cm .35cm}, clip]{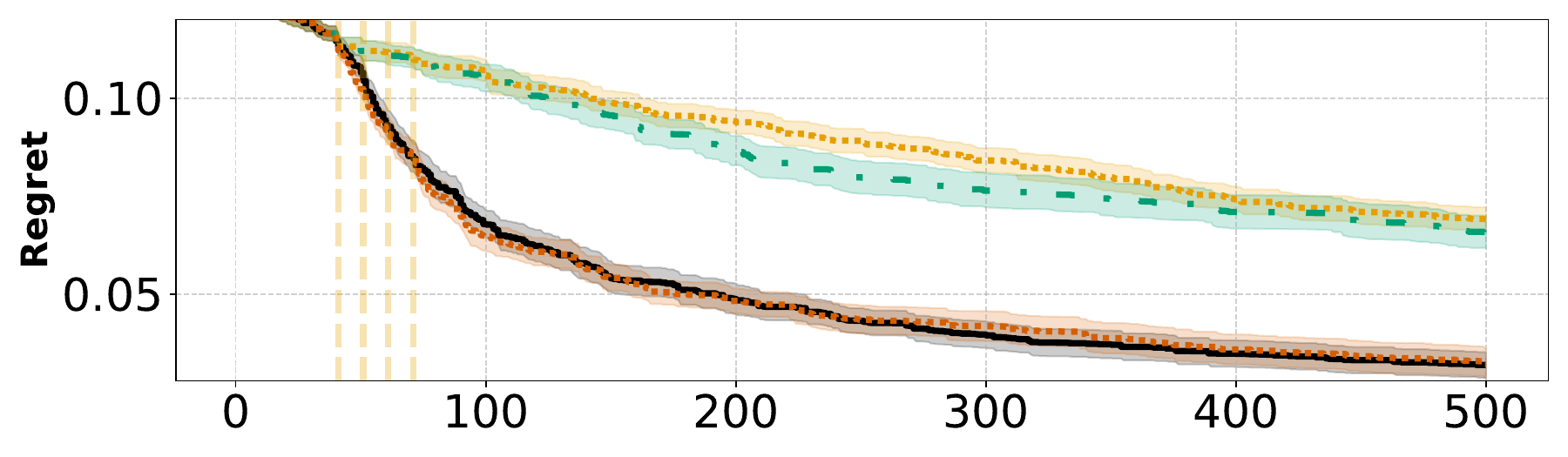} &
    \includegraphics[width=.43\linewidth, trim={.35cm .35cm .35cm .35cm}, clip]{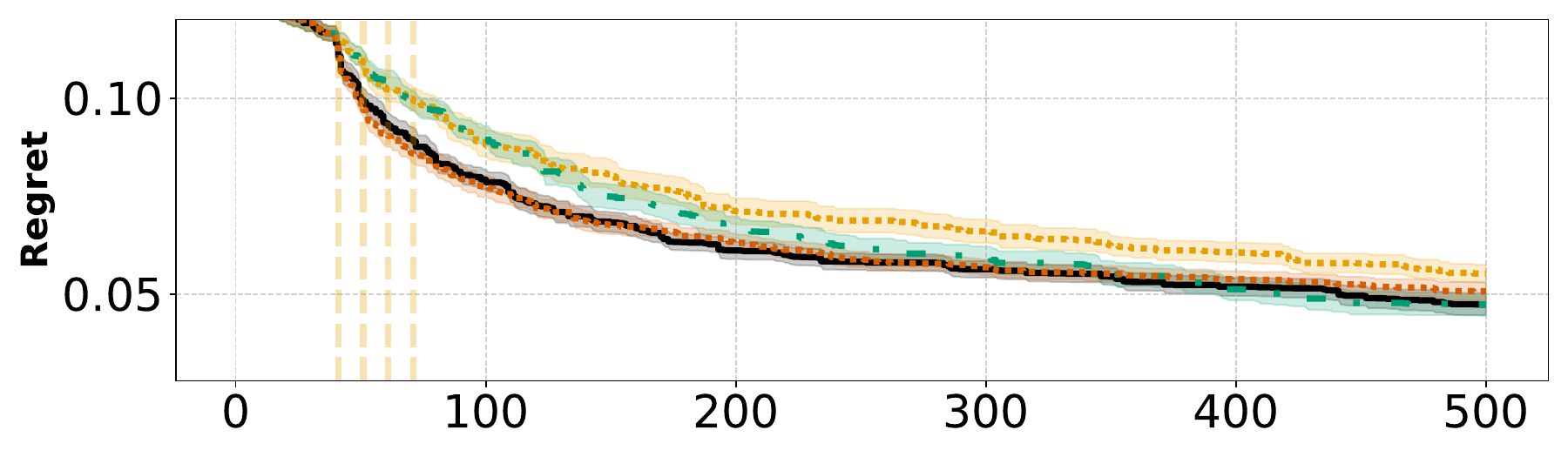} \\
    \end{tabular}
    \includegraphics[width=.7\textwidth,trim={.0cm 0cm 0cm 7.5cm},clip]{figures/iclr_submission/increased_budget/legend.pdf}
    \label{tab:surrogate_table_of_figures}
    \caption{Evaluation results across scenarios (rows) for two surrogate models (columns).}
\end{figure}
\newpage

\subsection{LCB Results}\label{app:further-results:lcb_results}

\begin{figure*}[h]
  \centering
  \setlength{\figheight}{0.132\textwidth}
  \begin{tabular}{@{}c@{}c@{}c@{}c@{}c@{}c@{}c@{}}
    & \hspace{.6cm}\texttt{widernet} & \texttt{resnet} & \texttt{transf} & \texttt{xformer} & \texttt{lcbench} & \texttt{xgboost}\\ 
    \rotatebox{90}{\hspace{.5cm}\texttt{Expert}} &
    \includegraphics[height=\figheight,trim={0.35cm 1.45cm 0cm 0.35cm},clip]{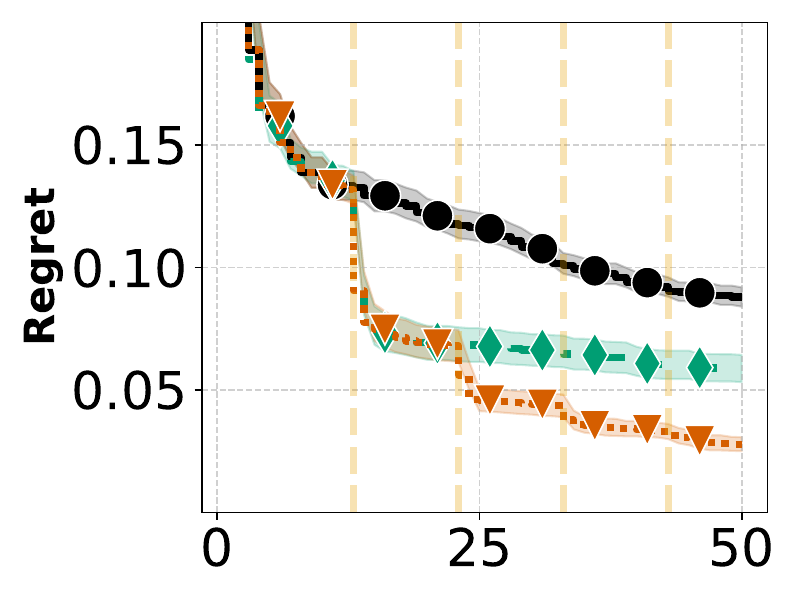} &
    \includegraphics[height=\figheight,trim={3.55cm 1.45cm 0cm 0.35cm},clip]{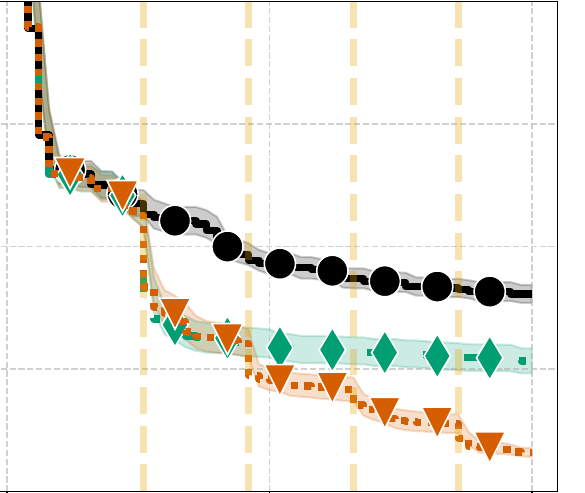} &
    \includegraphics[height=\figheight,trim={3.55cm 1.45cm 0cm 0.35cm},clip]{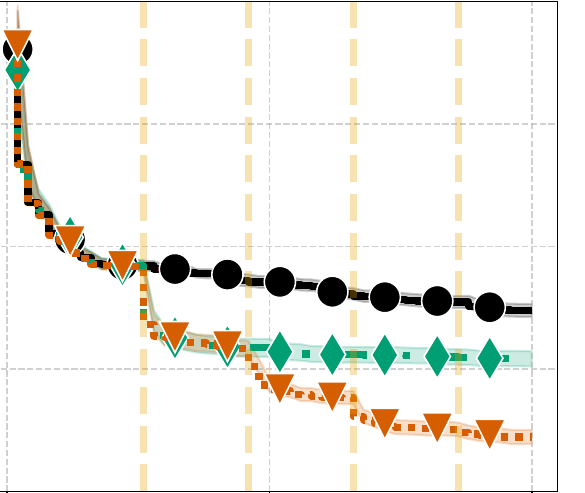} &
    \includegraphics[height=\figheight,trim={3.55cm 1.45cm 0cm 0.35cm},clip]{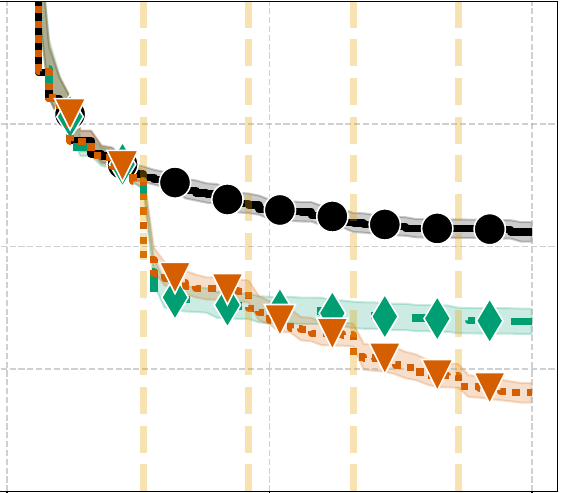} &
    \includegraphics[height=\figheight,trim={3.55cm 1.45cm 0cm 0.35cm},clip]{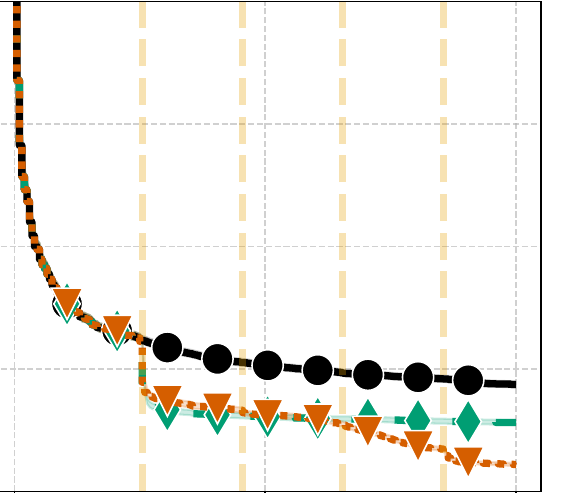} &
    \includegraphics[height=\figheight,trim={3.55cm 1.45cm 0cm 0.35cm},clip]{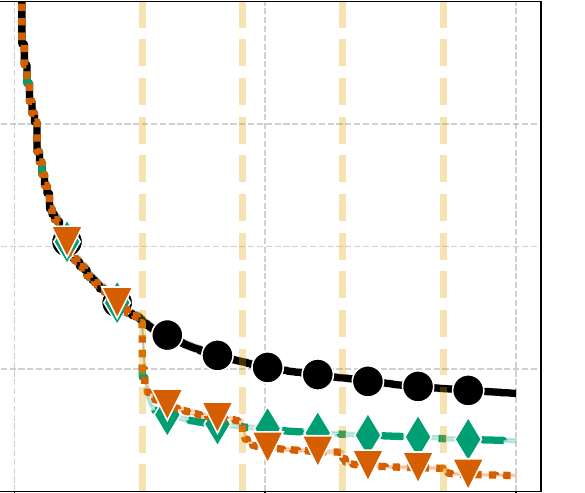}\\
 
    \rotatebox{90}{\hspace{.3cm}\texttt{Advanced}} &
    \includegraphics[height=\figheight,trim={0.35cm 1.45cm 0cm 0.35cm},clip]{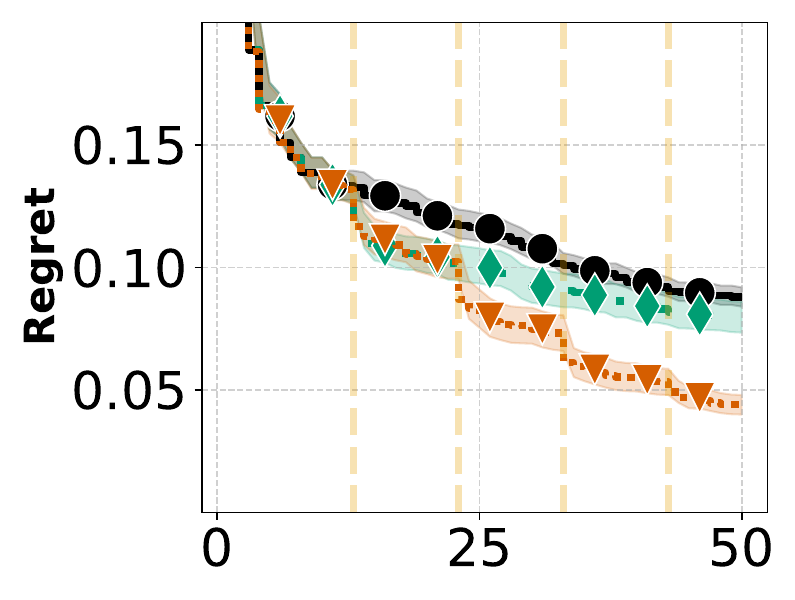} &
    \includegraphics[height=\figheight,trim={3.55cm 1.45cm 0cm 0.35cm},clip]{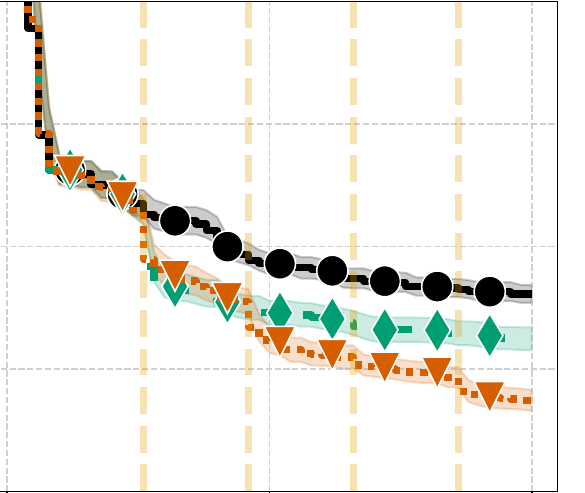} &
    \includegraphics[height=\figheight,trim={3.55cm 1.45cm 0cm 0.35cm},clip]{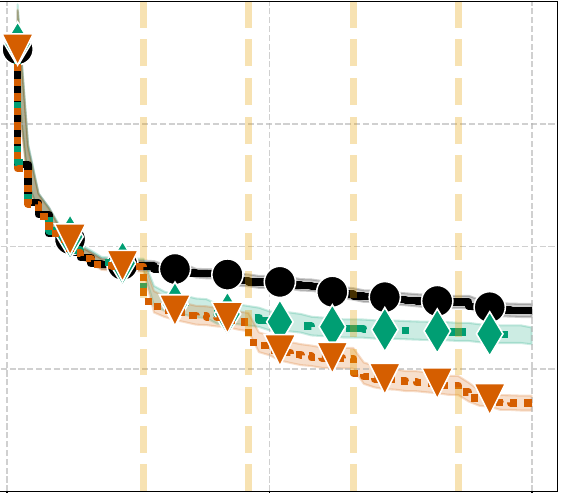} &
    \includegraphics[height=\figheight,trim={3.55cm 1.45cm 0cm 0.35cm},clip]{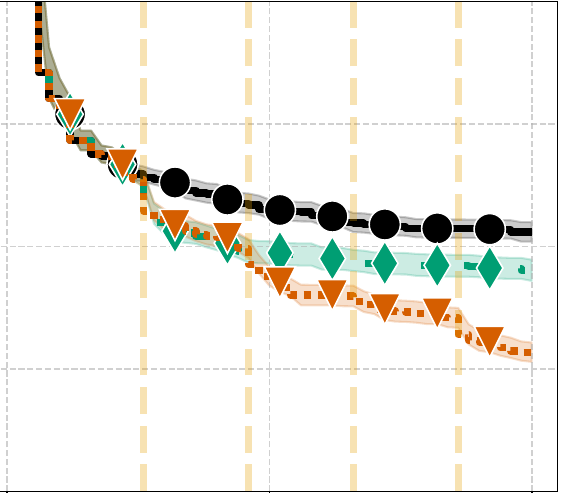} &
    \includegraphics[height=\figheight,trim={3.55cm 1.45cm 0cm 0.35cm},clip]{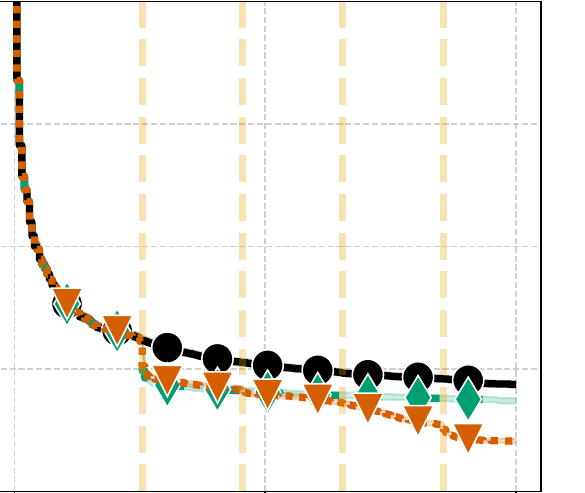} &
    \includegraphics[height=\figheight,trim={3.55cm 1.45cm 0cm 0.35cm},clip]{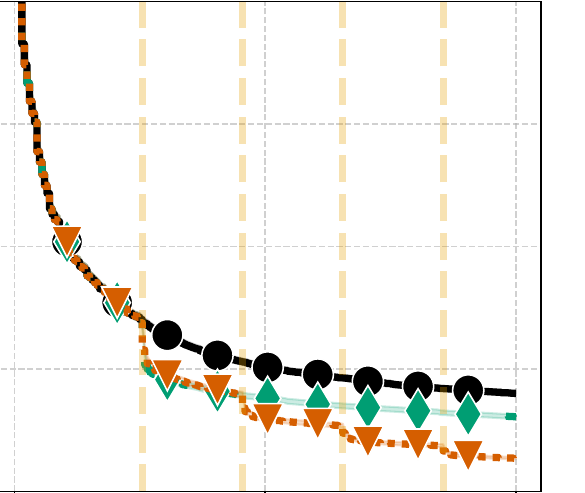}\\
 
    \rotatebox{90}{\hspace{.6cm}\texttt{Local}} &
    \includegraphics[height=\figheight,trim={0.35cm 1.45cm 0cm 0.35cm},clip]{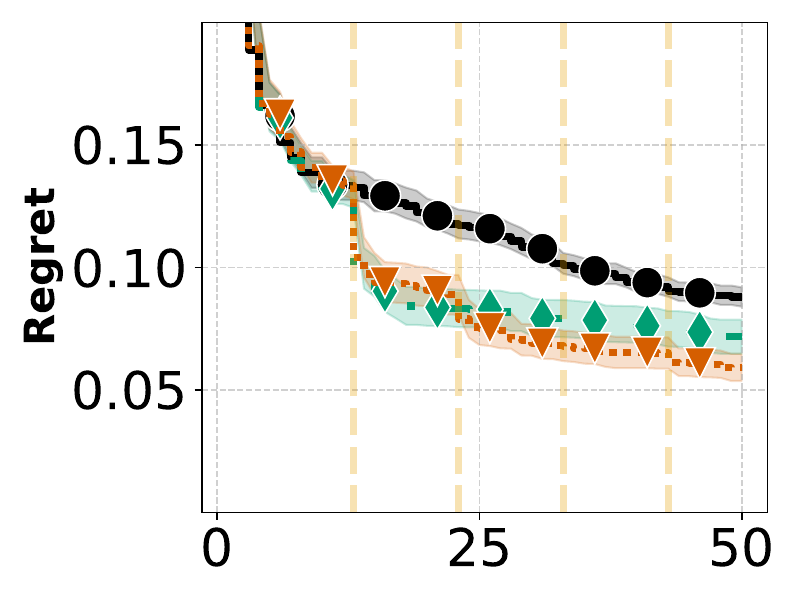} &
    \includegraphics[height=\figheight,trim={3.55cm 1.45cm 0cm 0.35cm},clip]{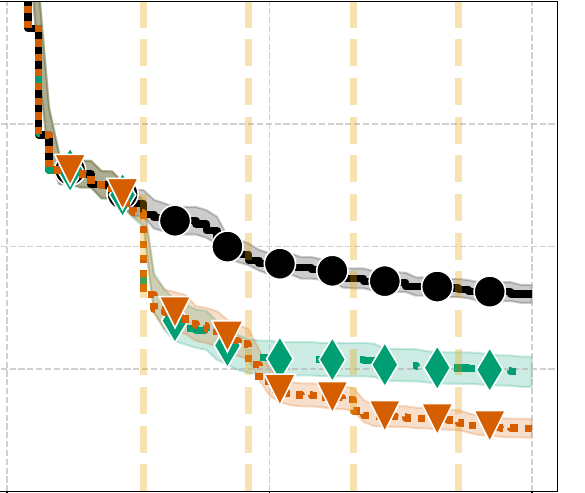} &
    \includegraphics[height=\figheight,trim={3.55cm 1.45cm 0cm 0.35cm},clip]{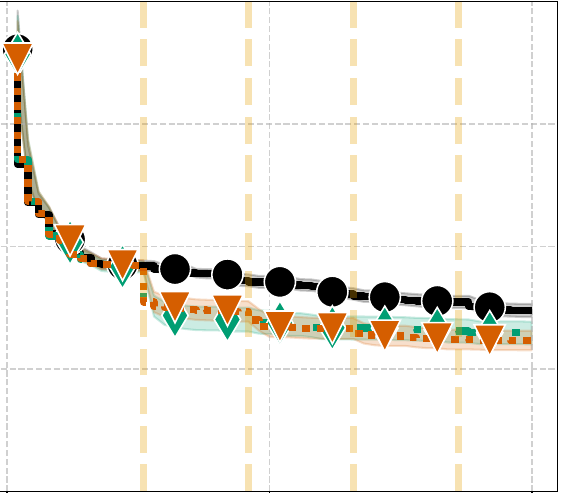} &
    \includegraphics[height=\figheight,trim={3.55cm 1.45cm 0cm 0.35cm},clip]{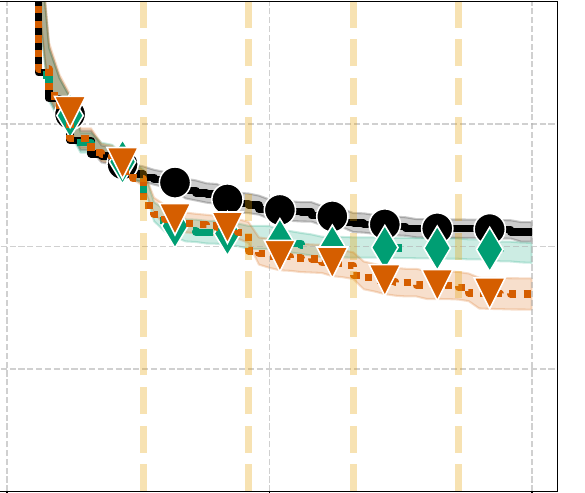} &
    \includegraphics[height=\figheight,trim={3.55cm 1.45cm 0cm 0.35cm},clip]{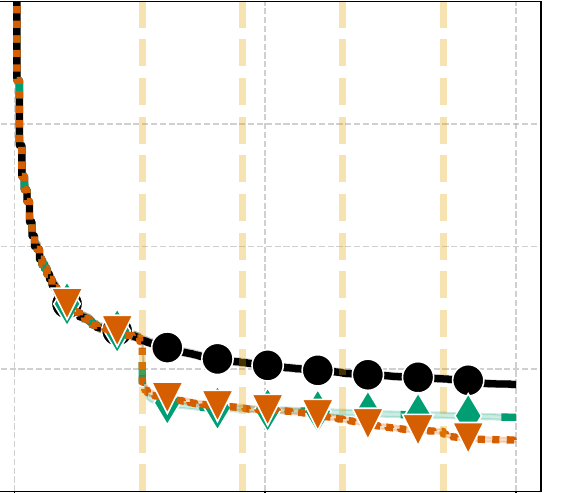} &    \includegraphics[height=\figheight,trim={3.55cm 1.45cm 0cm 0.35cm},clip]{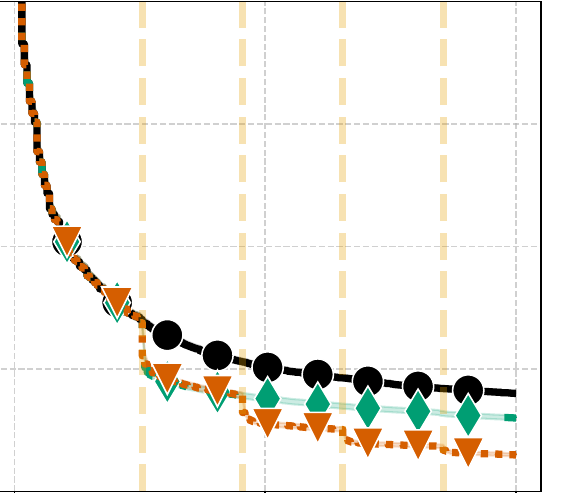}\\
 
    \rotatebox{90}{\hspace{.2cm}\texttt{Deceptive}} &
    \includegraphics[height=\figheight+0.13\figheight,trim={0.35cm 0.35cm 0cm 0.35cm},clip]{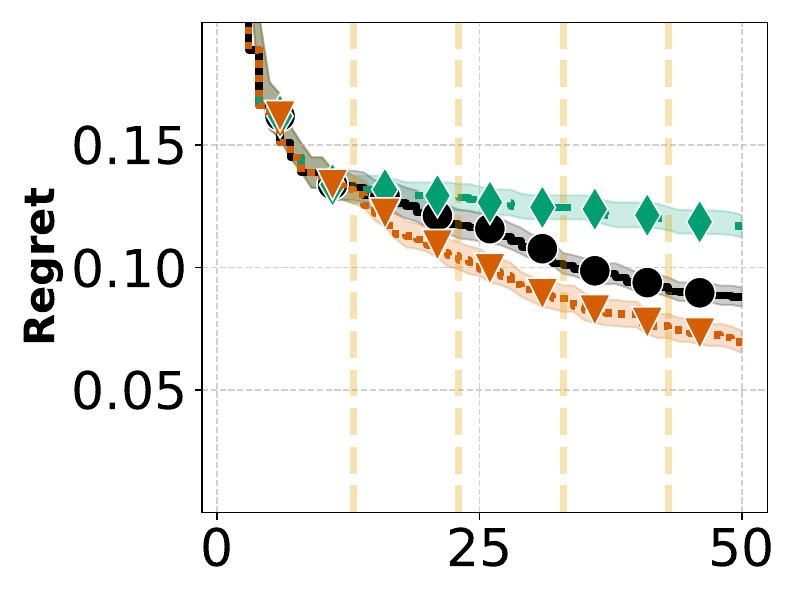} &
    \includegraphics[height=\figheight+0.13\figheight,trim={3.55cm 0.35cm 0cm 0.35cm},clip]{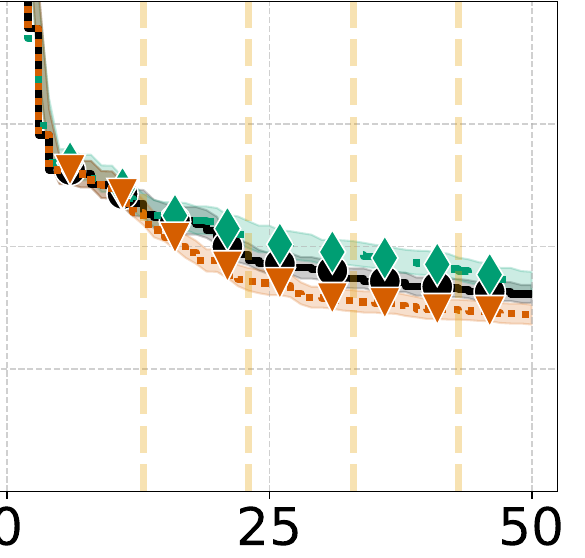} &
    \includegraphics[height=\figheight+0.13\figheight,trim={3.55cm 0.35cm 0cm 0.35cm},clip]{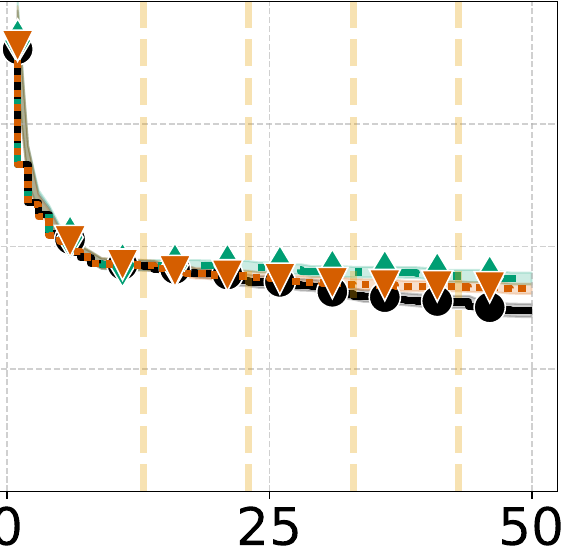} &
    \includegraphics[height=\figheight+0.13\figheight,trim={3.55cm 0.35cm 0cm 0.35cm},clip]{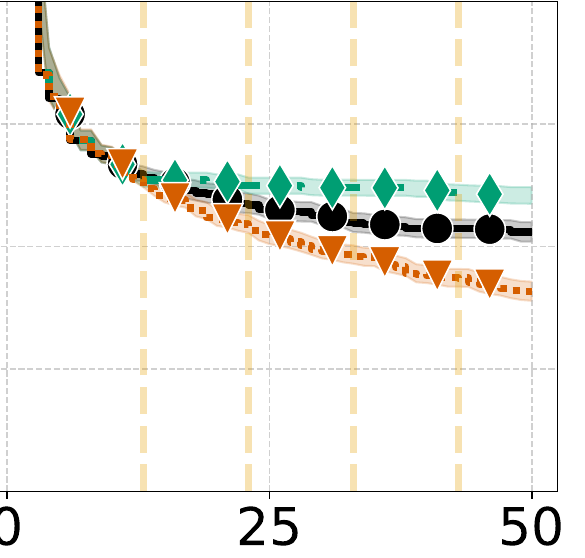} &
    \includegraphics[height=\figheight+0.13\figheight,trim={3.55cm 0.35cm 0cm 0.35cm},clip]{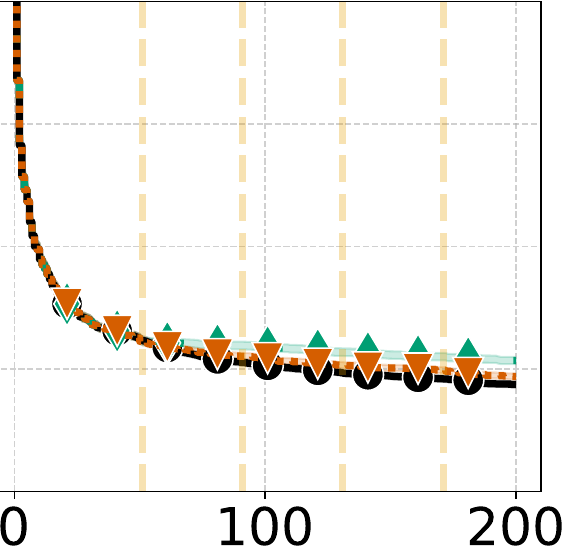} &
    \includegraphics[height=\figheight+0.13\figheight,trim={3.55cm 0.35cm 0cm 0.35cm},clip]{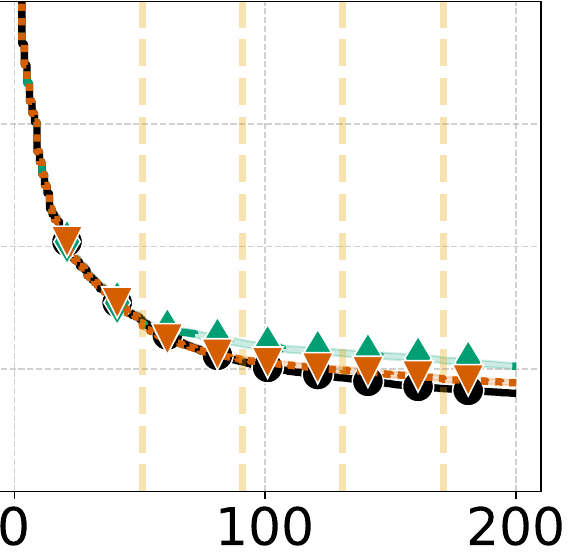}
  \end{tabular}
 
  \begin{minipage}{\textwidth}
    \centering
    \vspace{-.2cm}
    \includegraphics[width=.3\textwidth,trim={3cm 0cm .3cm 14cm},clip]{figures/iclr_submission/final_results/number_of_evaluations.pdf}
  \end{minipage}
 
  \begin{minipage}{\textwidth}
    \centering
    \includegraphics[width=.5\textwidth,trim={0cm 0cm 0cm 10.3cm},clip]{figures/ICML_26_submission/legend.pdf}
  \end{minipage}
 
  \caption{Mean regret for \texttt{lcbench}, \texttt{xgboost}, and \texttt{PD1} using \texttt{Expert}, \texttt{Advanced}, \texttt{Local}, and \texttt{Deceptive} priors. Priors are provided at vertical lines. The shaded areas visualize the standard error. For \texttt{lcbench} and \texttt{xgboost}, the plots average all datasets. The results indicate \tool outperforming \pibo and remaining competitive to vanilla BO for deceptive priors.}
  \label{fig:main-results:lcbrf}
\end{figure*}

As indicated in the main paper, we also conduct our main experiments using the LCB acquisition function both with random forests, and gaussian processes as surrogate models. The resulting plots can be found in \cref{fig:main-results:lcbrf}. The general trend aligns with the trend of the evaluations of the main paper.
\paragraph{Random Forest}

\begin{figure*}[h]
  \centering
  \setlength{\figheight}{0.132\textwidth}
  \begin{tabular}{@{}c@{}c@{}c@{}c@{}c@{}}
    & \hspace{.6cm}\texttt{widernet} & \texttt{resnet} & \texttt{transf} & \texttt{xformer} \\ 
    \rotatebox{90}{\hspace{.5cm}\texttt{Expert}} &
    \includegraphics[height=\figheight,trim={0.35cm 1.45cm 0cm 0.35cm},clip]{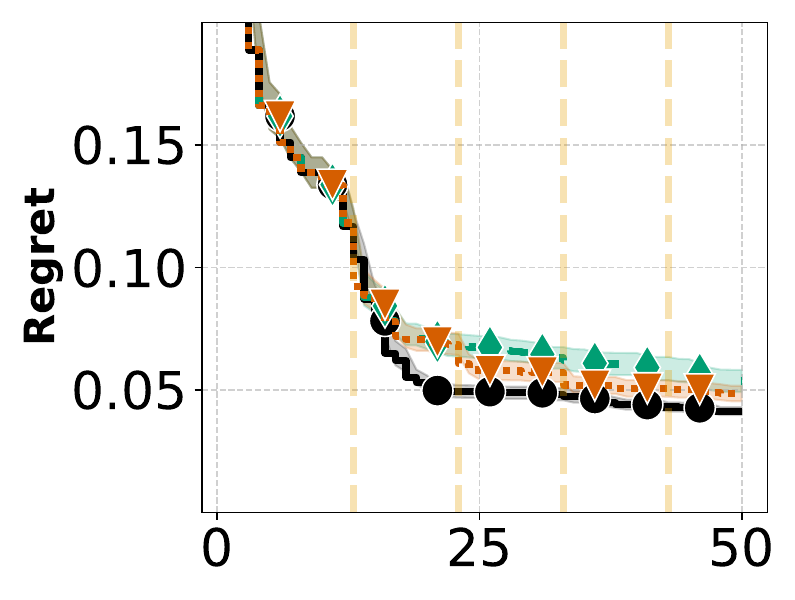} &
    \includegraphics[height=\figheight,trim={3.55cm 1.45cm 0cm 0.35cm},clip]{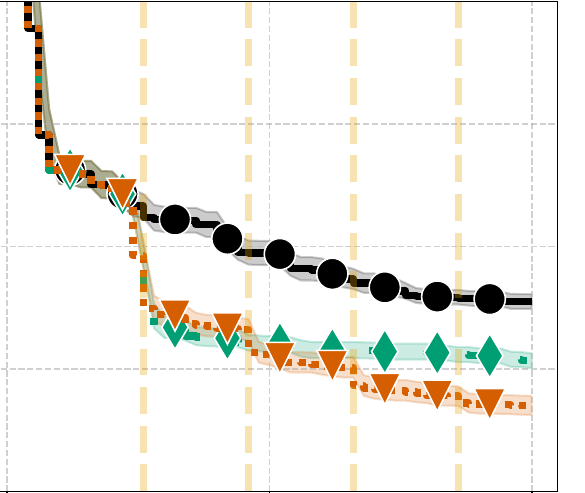} &
    \includegraphics[height=\figheight,trim={3.55cm 1.45cm 0cm 0.35cm},clip]{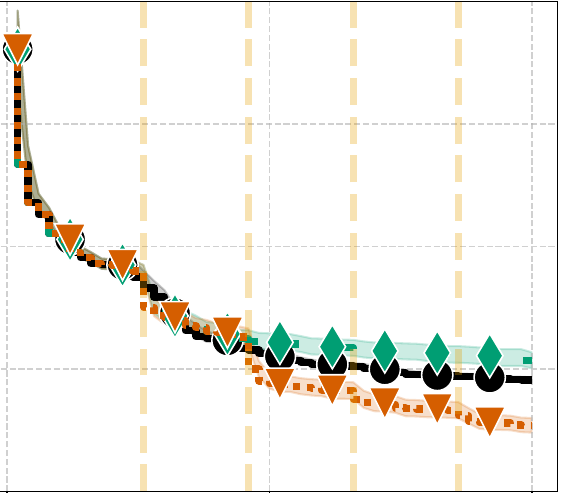} &
    \includegraphics[height=\figheight,trim={3.55cm 1.45cm 0cm 0.35cm},clip]{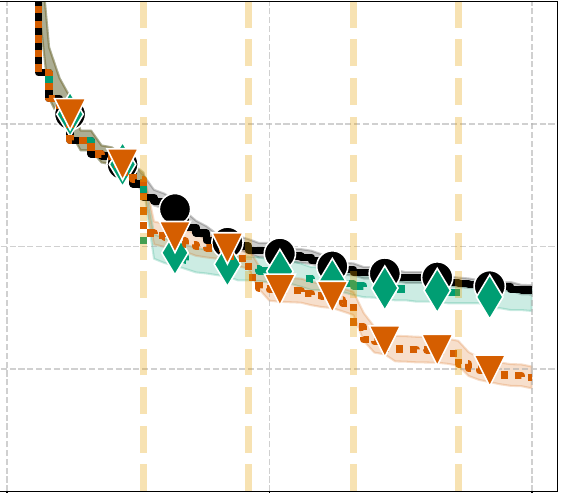} \\
    
    \rotatebox{90}{\hspace{.3cm}\texttt{Advanced}} &
    \includegraphics[height=\figheight,trim={0.35cm 1.45cm 0cm 0.35cm},clip]{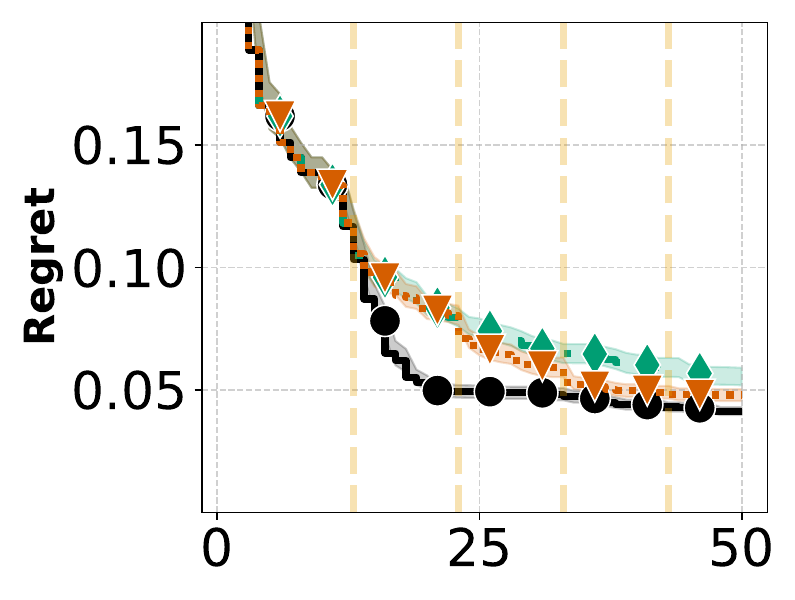} &
    \includegraphics[height=\figheight,trim={3.55cm 1.45cm 0cm 0.35cm},clip]{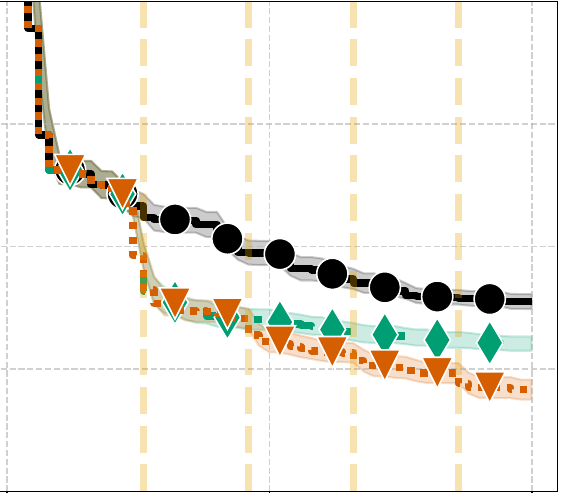} &
    \includegraphics[height=\figheight,trim={3.55cm 1.45cm 0cm 0.35cm},clip]{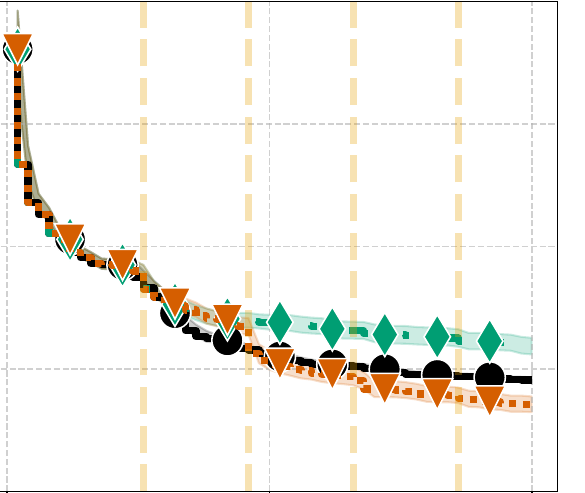} &
    \includegraphics[height=\figheight,trim={3.55cm 1.45cm 0cm 0.35cm},clip]{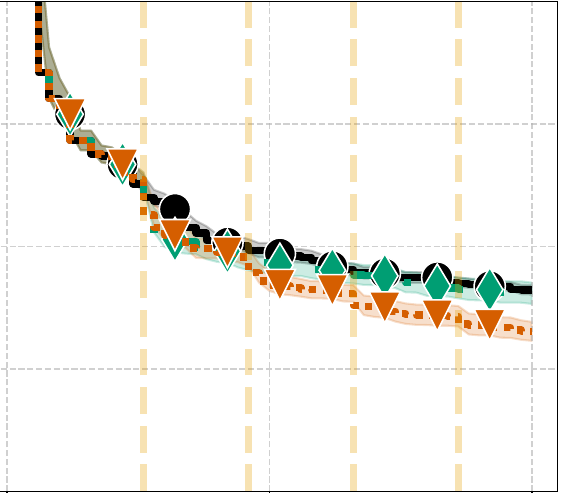} \\
    
    \rotatebox{90}{\hspace{.6cm}\texttt{Local}} &
    \includegraphics[height=\figheight,trim={0.35cm 1.45cm 0cm 0.35cm},clip]{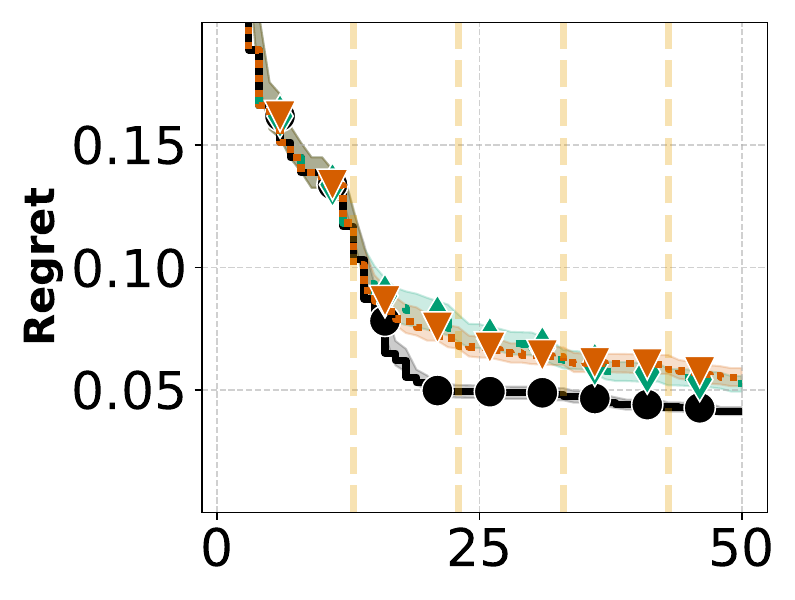} &
    \includegraphics[height=\figheight,trim={3.55cm 1.45cm 0cm 0.35cm},clip]{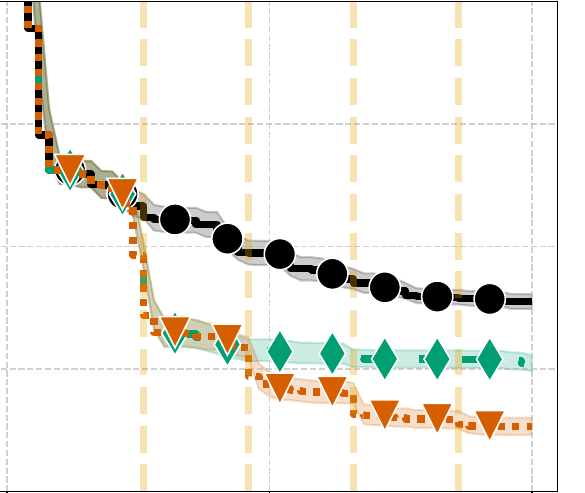} &
    \includegraphics[height=\figheight,trim={3.55cm 1.45cm 0cm 0.35cm},clip]{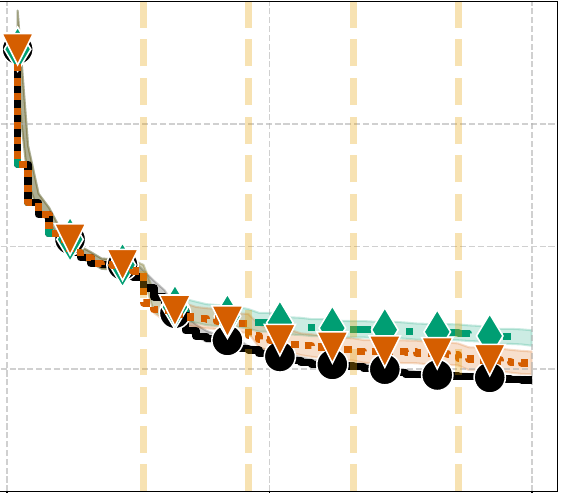} &
    \includegraphics[height=\figheight,trim={3.55cm 1.45cm 0cm 0.35cm},clip]{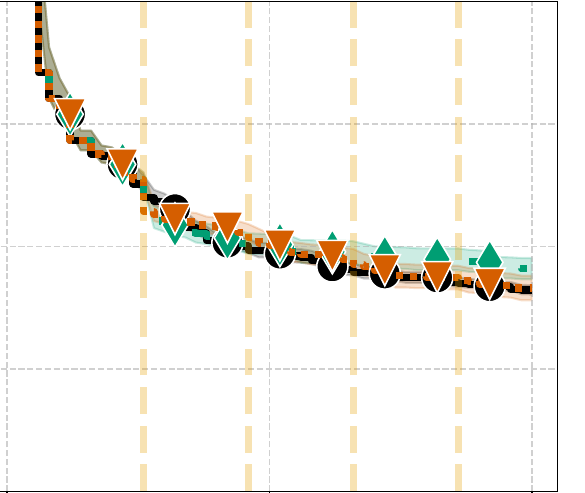} \\
    
    \rotatebox{90}{\hspace{.2cm}\texttt{Deceptive}} &
    \includegraphics[height=\figheight+0.13\figheight,trim={0.35cm 0.35cm 0cm 0.35cm},clip]{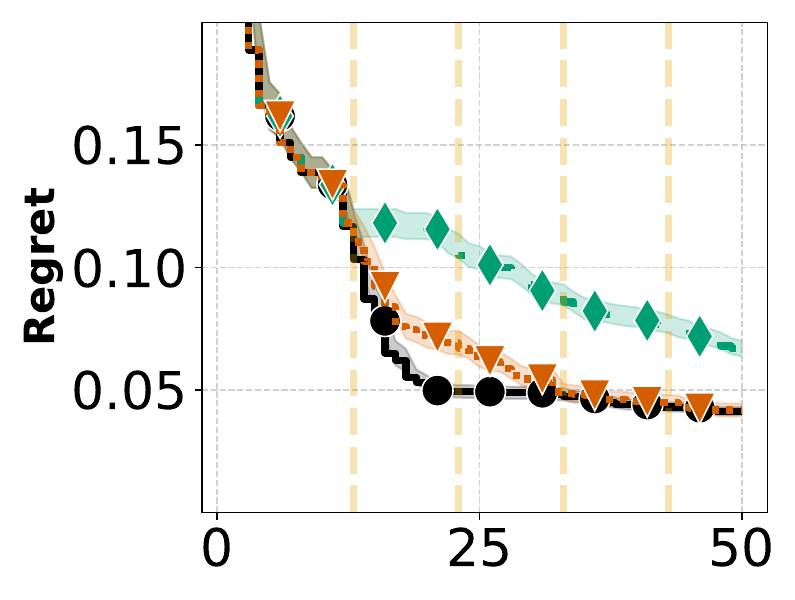} &
    \includegraphics[height=\figheight+0.13\figheight,trim={3.55cm 0.35cm 0cm 0.35cm},clip]{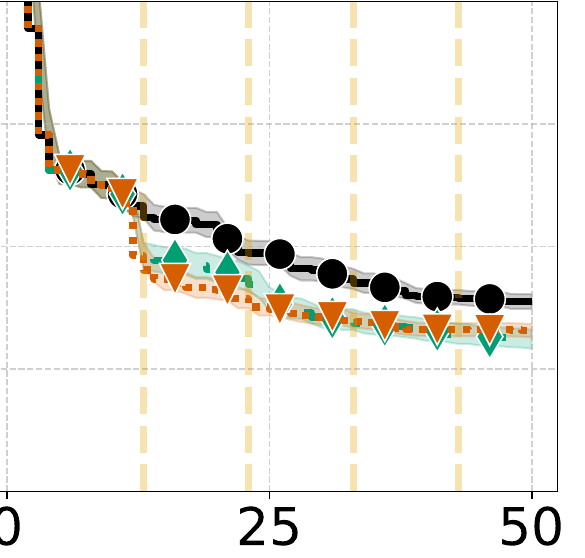} &
    \includegraphics[height=\figheight+0.13\figheight,trim={3.55cm 0.35cm 0cm 0.35cm},clip]{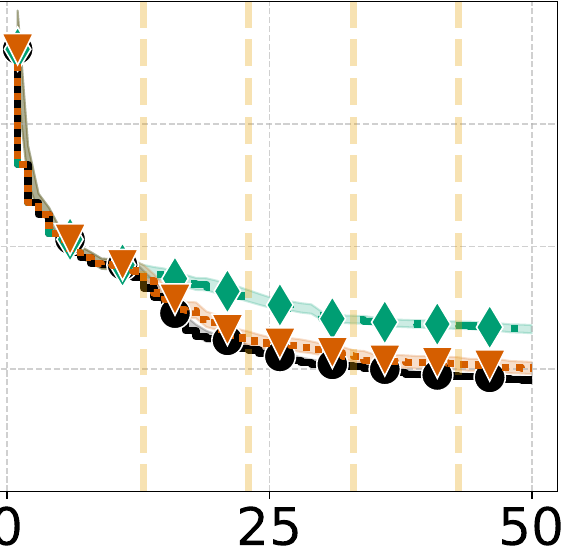} &
    \includegraphics[height=\figheight+0.13\figheight,trim={3.55cm 0.35cm 0cm 0.35cm},clip]{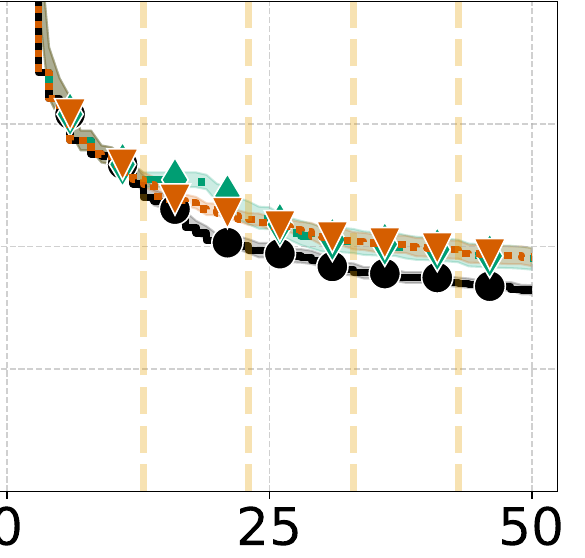} \\
  \end{tabular}
 
  \begin{minipage}{\textwidth}
    \centering
    \vspace{-.2cm}
    \includegraphics[width=.3\textwidth,trim={3cm 0cm .3cm 14cm},clip]{figures/iclr_submission/final_results/number_of_evaluations.pdf}
  \end{minipage}
 
  \begin{minipage}{\textwidth}
    \centering
    \includegraphics[width=.5\textwidth,trim={0cm 0cm 0cm 10.3cm},clip]{figures/ICML_26_submission/legend.pdf}
  \end{minipage}
 
  \caption{Mean regret for  \texttt{PD1} using \texttt{Expert}, \texttt{Advanced}, \texttt{Local}, and \texttt{Deceptive} priors. Priors are provided at vertical lines. The shaded areas visualize the standard error. For \texttt{lcbench} and \texttt{xgboost}, the plots average all datasets. The results indicate \tool outperforming \pibo and remaining competitive to vanilla BO for deceptive priors.}
  \label{fig:main-results:lcbgp}
\end{figure*}

\newpage

\subsection{Analysis of the Prior Rejection Criterion}\label{app:further-results:rejection_ablation}
\paragraph{Sensitivity Analysis of the Prior Rejection Criterion}

In this section, we first discuss the ablation study for the threshold $\tau$ followed by a scenario wide analysis of the rejected priors.

Our prior rejection scheme utilizes a threshold $\tau$, encoding the minimum estimated average improvement over the current incumbent needed to accept a prior. This improvement is quantified with LCB, one option for \cref{eq:threshold}. When $\tau < 0$, priors are accepted, even if they are predicted to be misleading; $\tau > 0$ ensures that only priors of ample potential are accepted.

\begin{figure}[H]
    \centering
        \texttt{overall}\\
        \includegraphics[width=.8\textwidth,trim={0cm 3.6cm 0cm .2cm},clip]{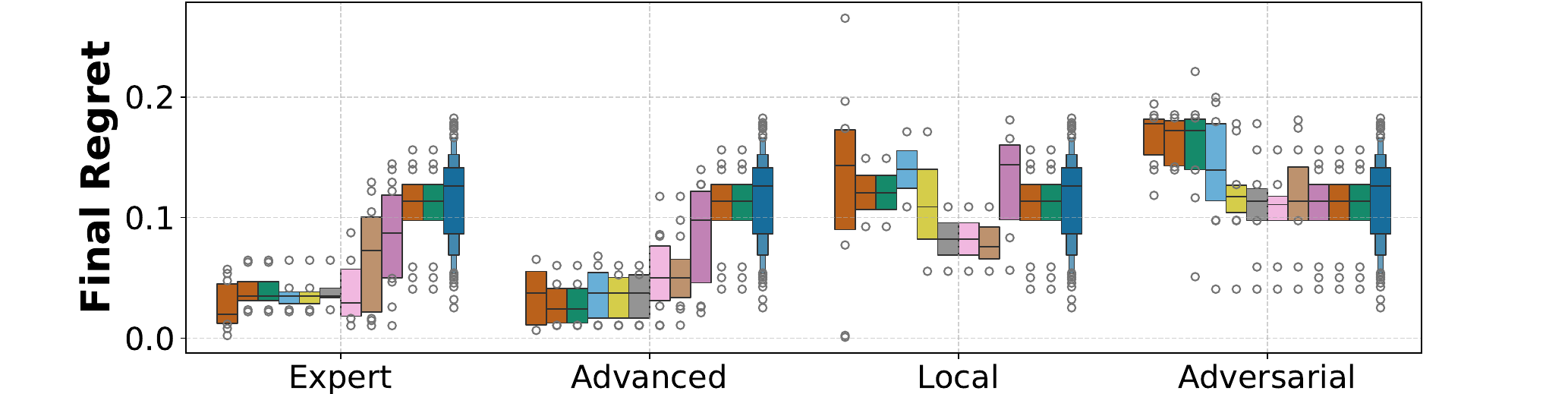}\\
        \texttt{widernet}\\
        \includegraphics[width=.8\textwidth,trim={0cm 3.6cm 0cm .2cm},clip]{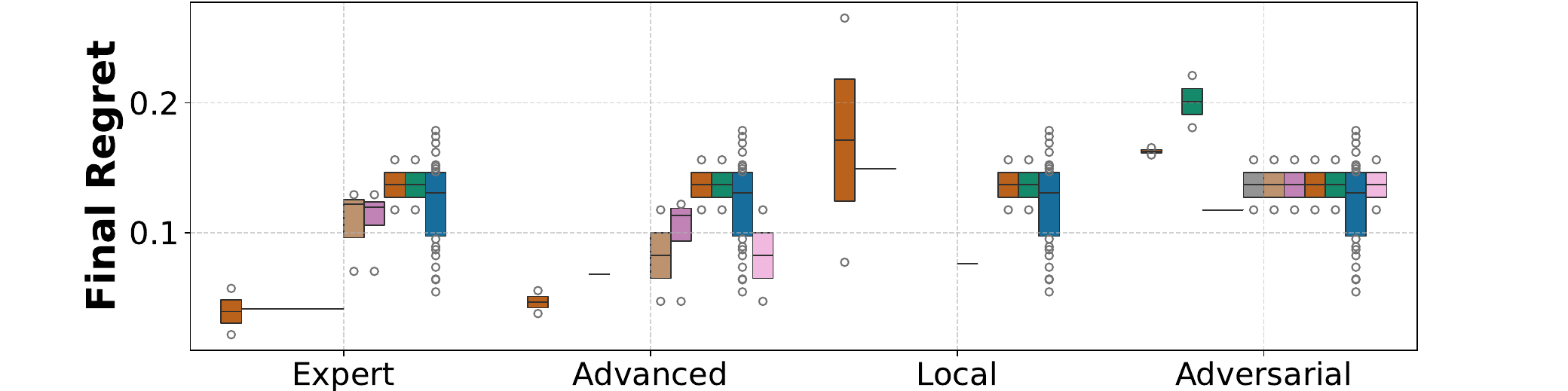}
        \includegraphics[width=.7\textwidth,trim={0cm 0cm 0cm 10cm},clip]{figures/iclr_submission/prior_rejection_ablation/final_cost_barplot_overall_final_cost_barplot.pdf}\\
        \texttt{resnet}\\
        \includegraphics[width=.8\textwidth,trim={0cm 3.6cm 0cm .2cm},clip]{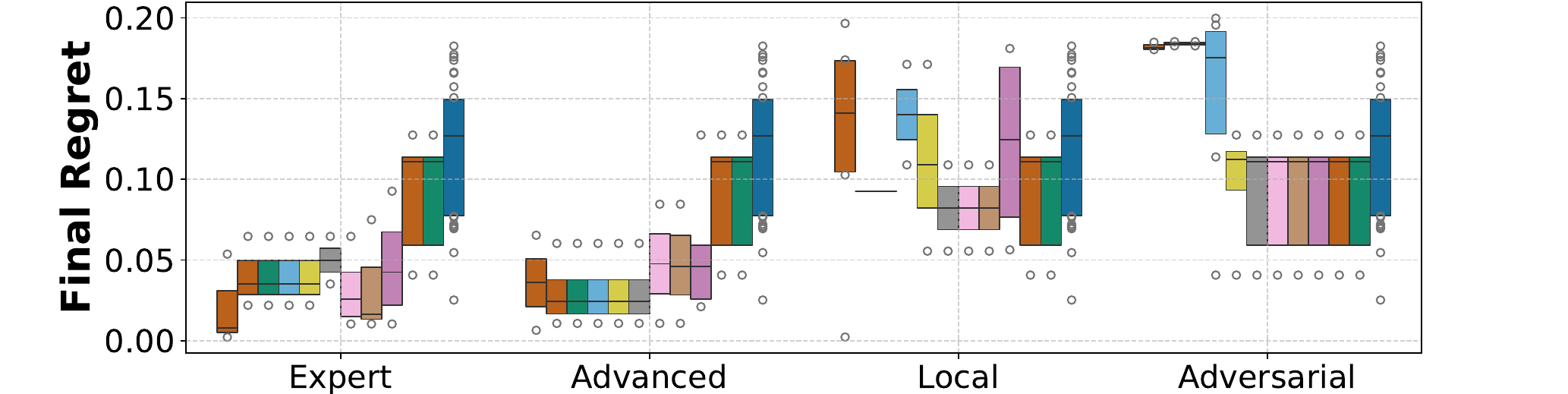}\\
        \texttt{transf}\\
        \includegraphics[width=.8\textwidth,trim={0cm 3.6cm 0cm .2cm},clip]{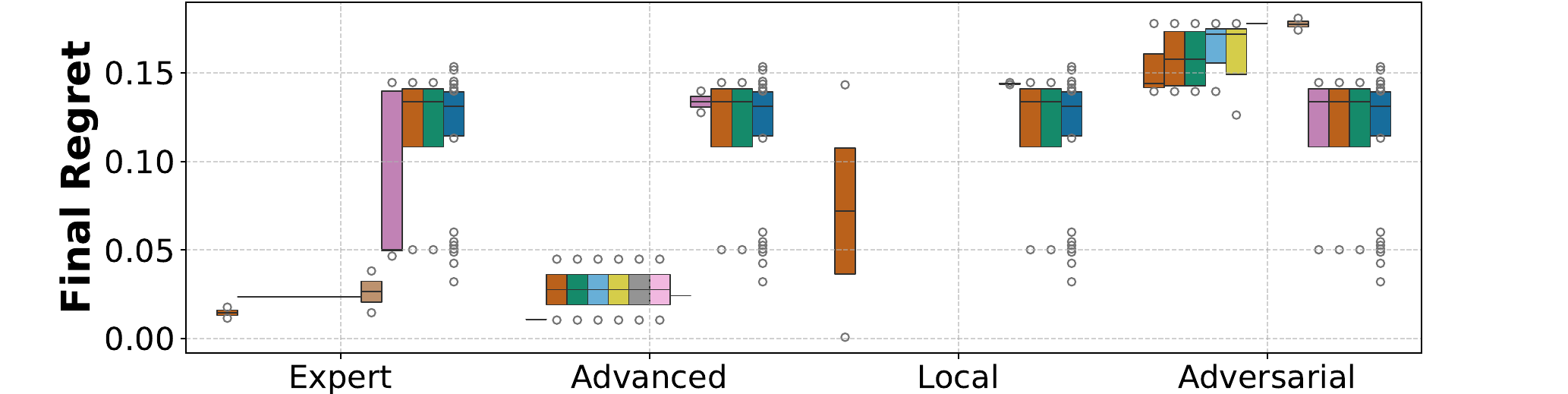}\\
    \caption{Sensitivity analysis of different thresholds $\tau$ (Part 1/2). $\tau = -\infty$ accepts all, and $\tau = \infty$ rejects all priors. \texttt{overall} contains the merged results from all scenarios. The following plots contain the results for one scenario respectively.}
    \label{fig:additional-ablation-results-1}
\end{figure}

\begin{figure}[H]
    \centering
        \texttt{xformer}\\
        \includegraphics[width=.8\textwidth,trim={0cm 3.6cm 0cm .2cm},clip]{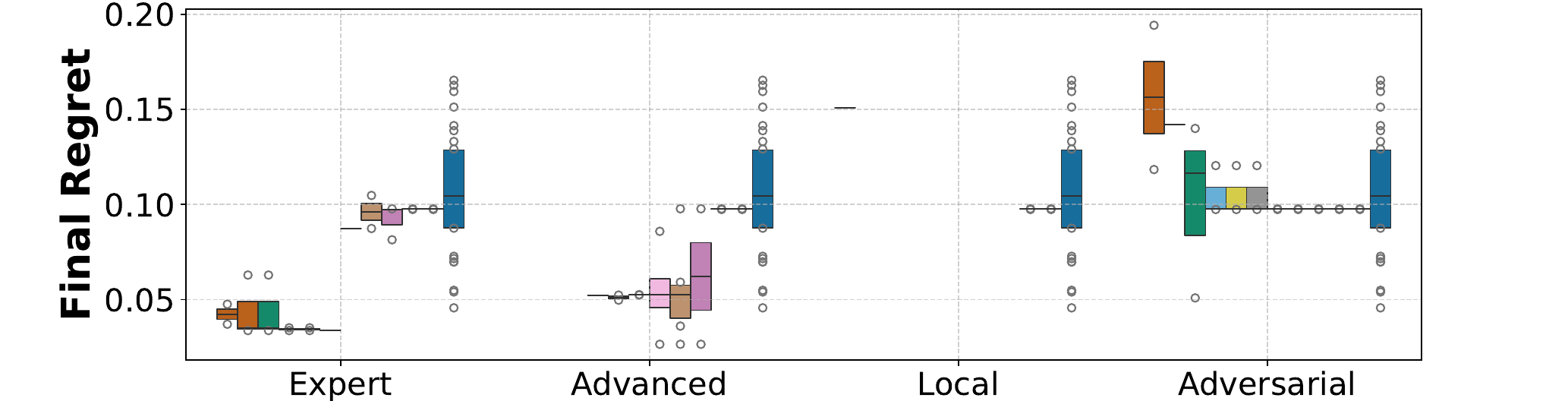}
        \includegraphics[width=.7\textwidth,trim={0cm 0cm 0cm 10cm},clip]{figures/iclr_submission/prior_rejection_ablation/final_cost_barplot_overall_final_cost_barplot.pdf}
    \caption{Sensitivity analysis of different thresholds $\tau$ (continued from Figure~\ref{fig:additional-ablation-results-1})}
    \label{fig:additional-ablation-results-2}
\end{figure}

To study the impact of $\tau$ on \tool, we conduct a sensitivity analysis on PD1's \citep{wang-jmlr24} optimization scenarios, shown in \cref{fig:additional-ablation-results-1,fig:additional-ablation-results-2}. As anticipated, $\tau$ enables a tradeoff between being permissive to potentially helpful and rejecting misleading priors. While $\tau$ could in principle be customized to reflect the user's confidence or expertise level, we find that setting $-0.25 \leq \tau \leq -0.05$ strikes a good balance. The preceding experiments utilize $\tau = -0.15$.

\paragraph{Analysis of Prior Rejection Criterion Behavior}

\begin{figure}[H]
    \centering
    \includegraphics[width=\textwidth,trim={0cm 0cm 0cm 0cm},clip]{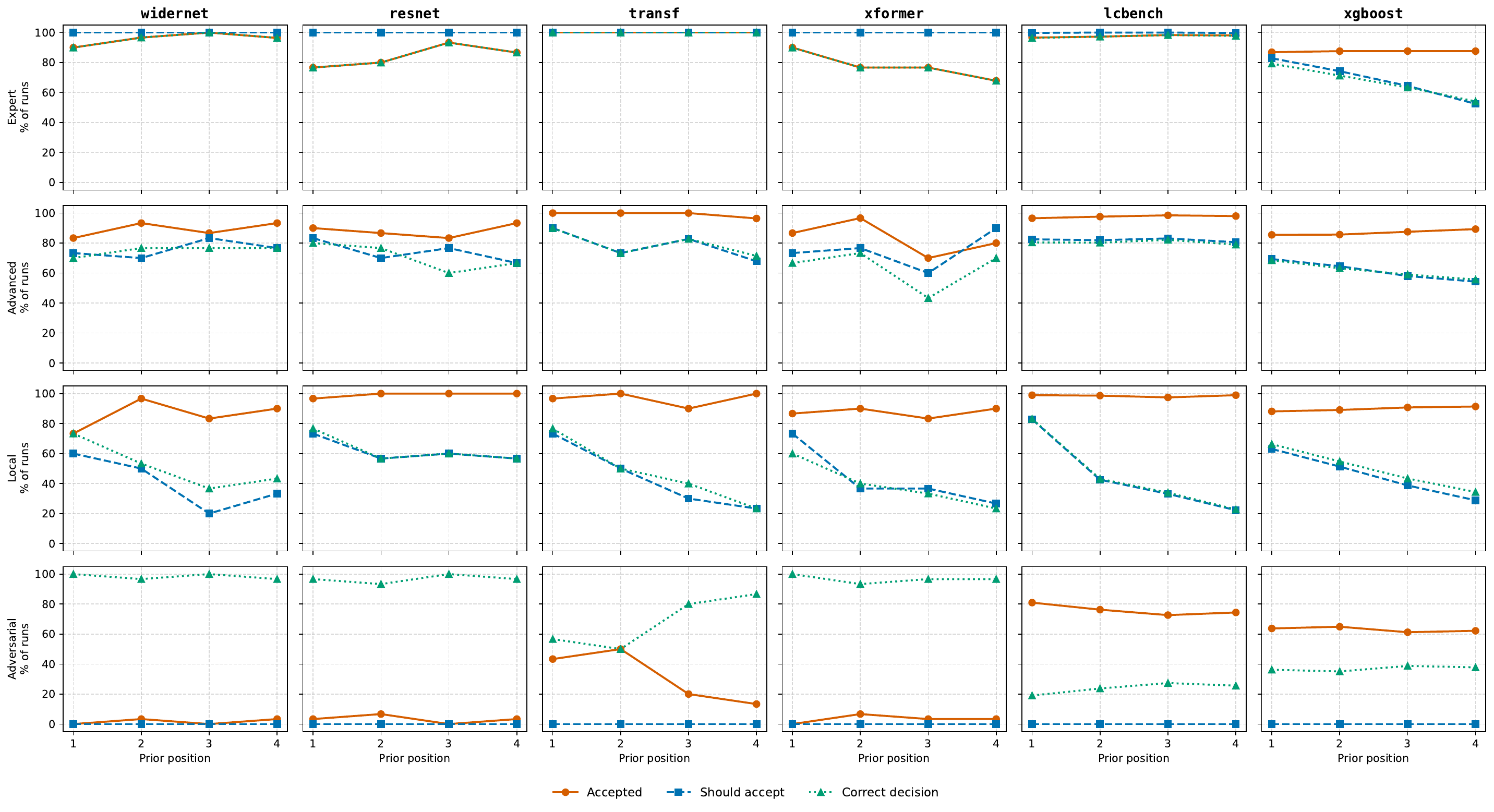}
    \caption{Behavior of the prior rejection criterion ($\tau = -0.15$) across all six scenarios. Rows are prior kinds, columns scenarios; $x$-axis is the four priors offered per run, $y$-axis the percentage of runs (over seeds, and over datasets for the \yahpo scenarios). \textbf{Accepted}: priors \tool{} accepted; \textbf{Should accept}: priors actually superior to the incumbent (oracle); \textbf{Correct decision}: where \tool{} agreed with the oracle.}
    \label{fig:rejection-criterion-behavior}
\end{figure}

\cref{fig:rejection-criterion-behavior} reports, for each prior offered during optimization, three quantities: the frequency with which \tool{} accepted the prior (Accepted), the frequency with which the prior was genuinely superior to the current incumbent (i.e., the quality of the configuration at the prior center exceeded that of the incumbent), and the frequency with which the accept/reject decision agreed with this oracle (Correct decision). 

Two trends hold consistently across the five well-scaled scenarios (the four PD1 scenarios and rbv2 XGBoost). First, the criterion tracks prior quality: for Expert priors, nearly all priors are both superior and accepted, and as quality degrades (Advanced to Local), the fraction of genuinely superior priors declines and \tool{} correspondingly, and correctly, declines to adopt them more often. Second, for deceptive priors, which are almost never superior, \tool{} rejects nearly all of them, yielding a high correct-decision rate despite the priors being uniformly misleading. The should-accept rate likewise decays with prior position: later in the run, the incumbent is stronger, so a prior must be of higher quality to improve upon it. Taken together, these results indicate that the criterion adapts to the actual usefulness of incoming priors rather than to their nominal category.

\begin{figure}[H]
    \centering
    \includegraphics[width=\linewidth]{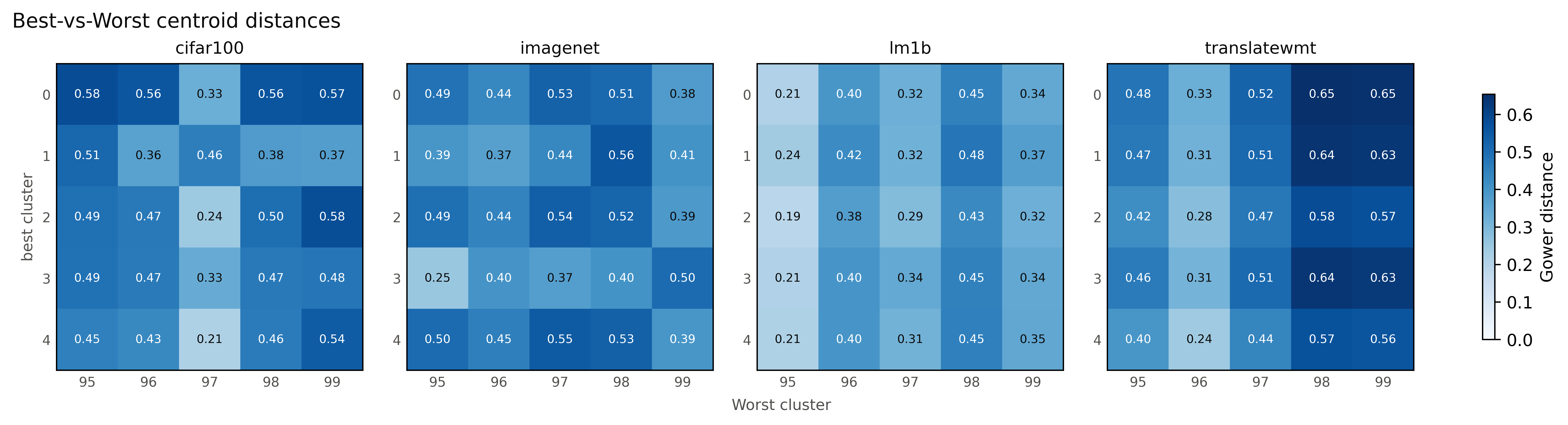}
    \caption{Medoid Distances of the top-5 clusters to the worst-5 clusters.}
    \label{fig:search_space_analysis}
\end{figure}

However, for the scenario \texttt{lm1b}, deceptive priors are less likely to be rejected. We hypothesize that this is caused by the loss landscape. As shown in \cref{fig:search_space_analysis} and \cref{tab:cluster-distances}, the best and worst clusters are located closer together in terms of Gower Distance \citep{gower-biometrics1971a}. 

\begin{table}[htbp]
  \centering
  \begin{tabular}{lrrr}
    \toprule
    Scenario & Best$\leftrightarrow$Worst & Mean dist. (all pairs)  & Nearest best \\
    \midrule
    cifar100\_wideresnet\_2048   & 0.452 & 0.320  & 0.355 \\
    imagenet\_resnet\_512        & 0.449 & 0.301  & 0.355 \\
    lm1b\_transformer\_2048      & 0.344 & 0.300  & 0.320 \\
    translatewmt\_xformer\_64    & 0.491 & 0.306  & 0.440 \\
    \bottomrule
  \end{tabular}
  \caption{Cluster distance metrics by scenario.}
  \label{tab:cluster-distances}
\end{table}

Similarly, the quality of the prior rejection criterion degrades for the \yahpo scenarios. We hypothesize that this is a result of the differing search space. Nevertheless, \tool achieves impressive performance over all scenarios. 

\newpage

\subsection{Detailed Comparison with Probabilistic Circuits}\label{app:further-results:pc}

\begin{figure}[H]
  \centering
  \setlength{\figheight}{0.2\textwidth}
  
  \resizebox{\textwidth}{!}{
  \begin{tabular}{@{}c@{}c@{}c@{}c@{}c@{}c}
    & \hspace{.6cm}\texttt{widernet} & \texttt{resnet} & \texttt{transf} & \texttt{xformer}\\ 
    \rotatebox{90}{\texttt{\hspace{.4cm} Expert}} &
    \includegraphics[height=\figheight,trim={0.35cm 1.45cm 0cm 0.35cm},clip]{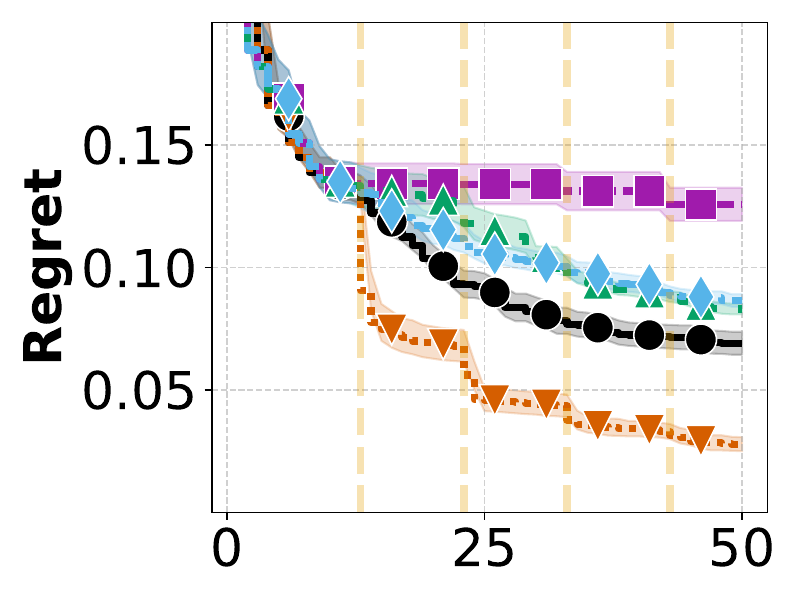} &
    \includegraphics[height=\figheight,trim={3.55cm 1.45cm 0cm 0.35cm},clip]{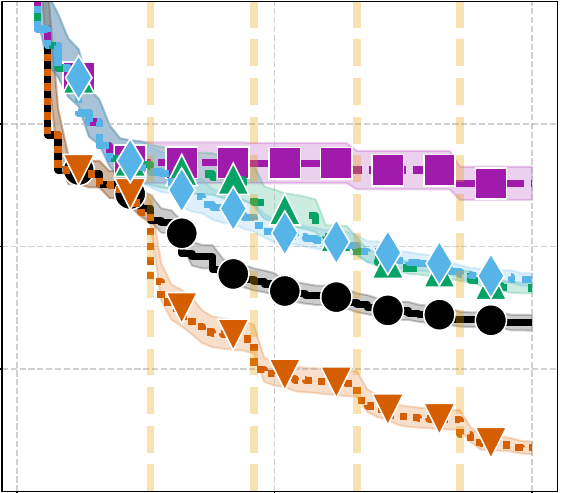} &
    \includegraphics[height=\figheight,trim={3.55cm 1.45cm 0cm 0.35cm},clip]{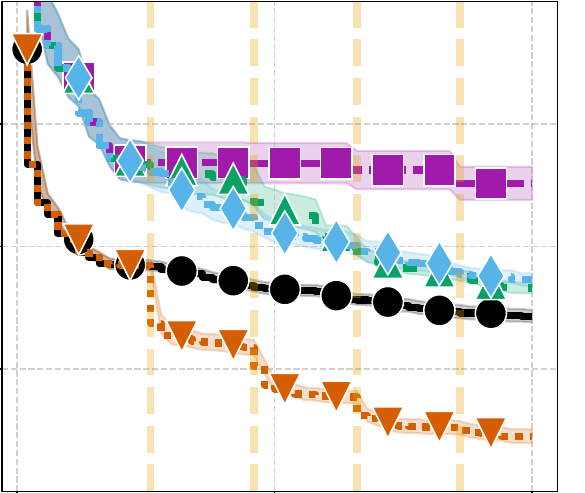} &
    \includegraphics[height=\figheight,trim={3.55cm 1.45cm 0cm 0.35cm},clip]{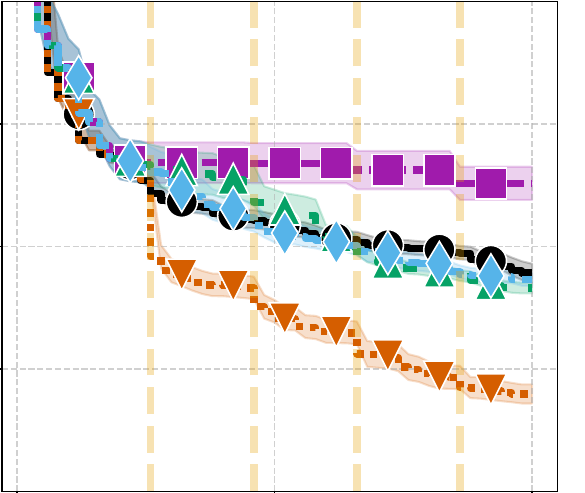} &
    \\

    \rotatebox{90}{\hspace{.2cm}\texttt{Advanced}} &
    \includegraphics[height=\figheight,trim={0.35cm 1.45cm 0cm 0.35cm},clip]{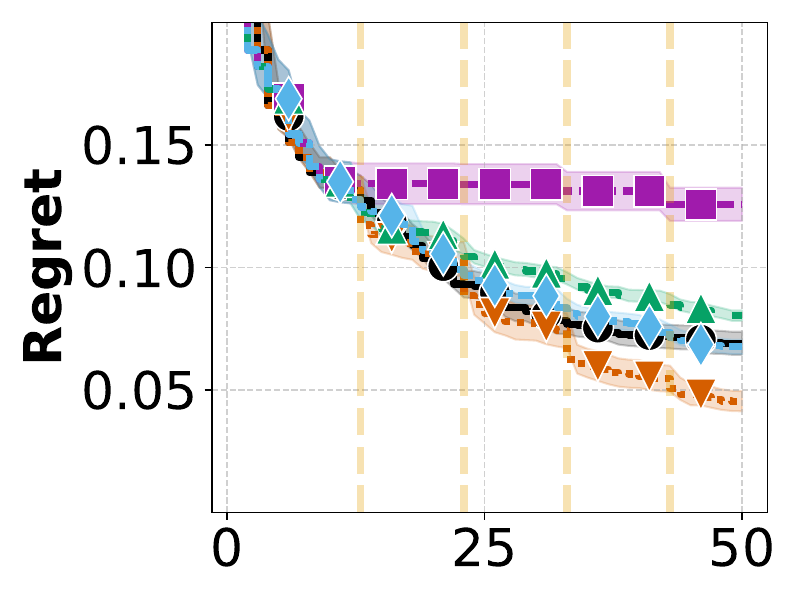} &
    \includegraphics[height=\figheight,trim={3.55cm 1.45cm 0cm 0.35cm},clip]{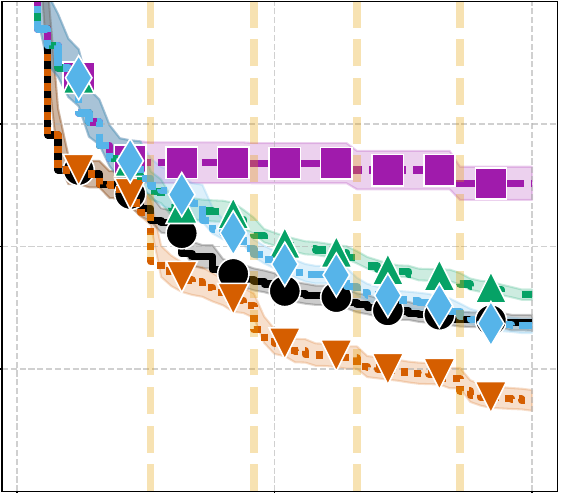} &
    \includegraphics[height=\figheight,trim={3.55cm 1.45cm 0cm 0.35cm},clip]{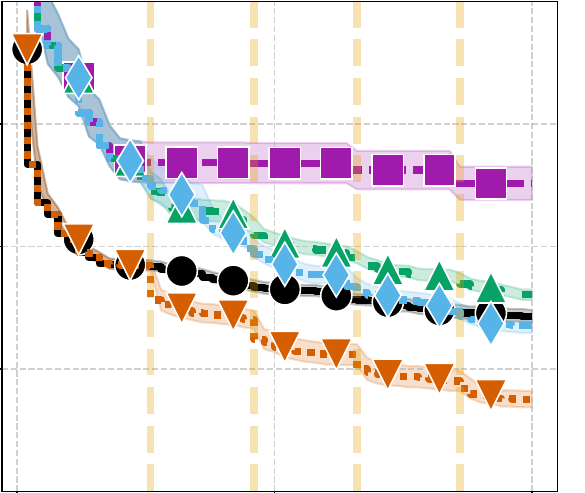} &
    \includegraphics[height=\figheight,trim={3.55cm 1.45cm 0cm 0.35cm},clip]{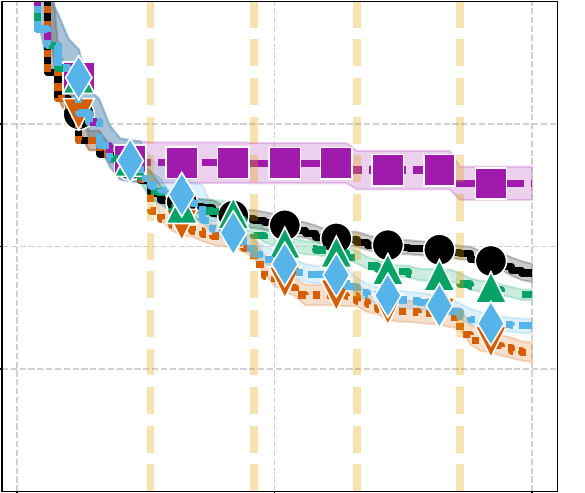} &
    \\

    \rotatebox{90}{\hspace{.75cm}\texttt{Local}} &
    \includegraphics[height=\figheight,trim={0.35cm 1.45cm 0cm 0.35cm},clip]{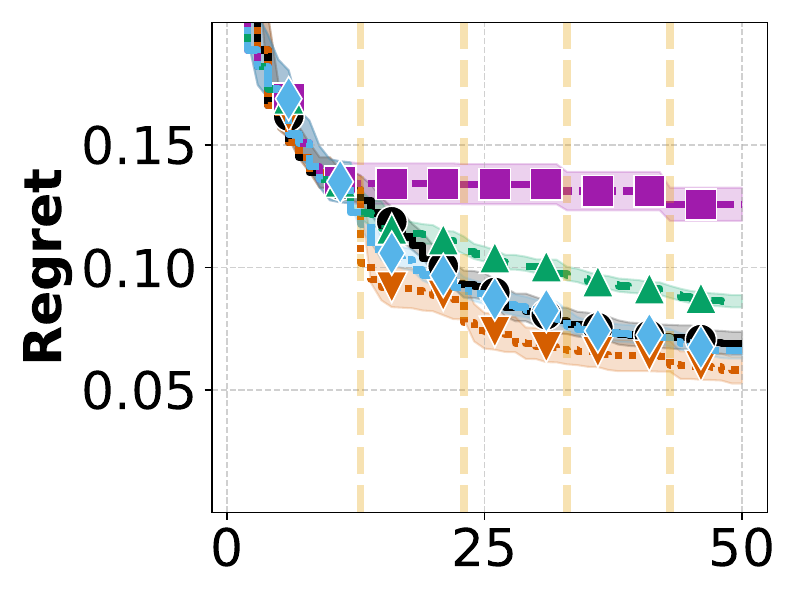} &
    \includegraphics[height=\figheight,trim={3.55cm 1.45cm 0cm 0.35cm},clip]{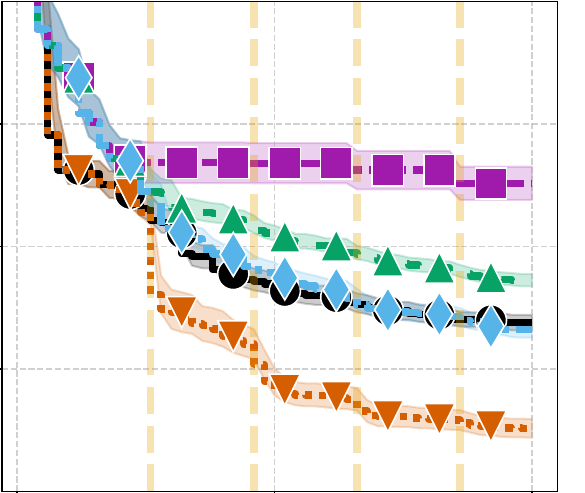} &
    \includegraphics[height=\figheight,trim={3.55cm 1.45cm 0cm 0.35cm},clip]{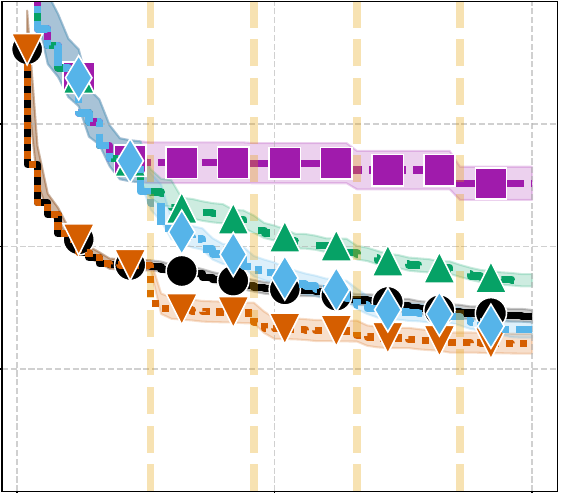} &
    \includegraphics[height=\figheight,trim={3.55cm 1.45cm 0cm 0.35cm},clip]{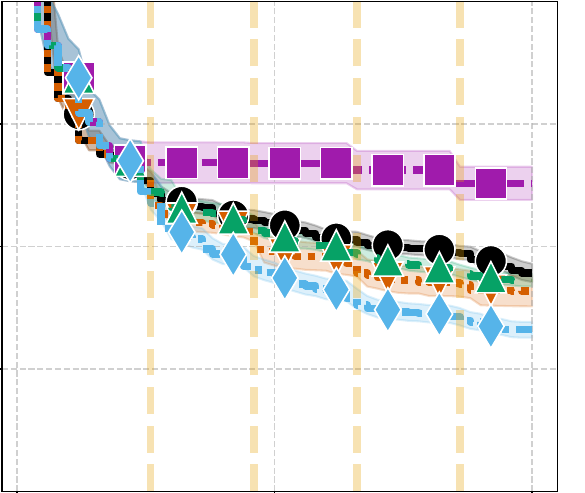} &
    \\

    \rotatebox{90}{\hspace{.3cm}\texttt{Deceptive}} &
    \includegraphics[height=\figheight+0.13\figheight,trim={0.35cm 0.35cm 0cm 0.35cm},clip]{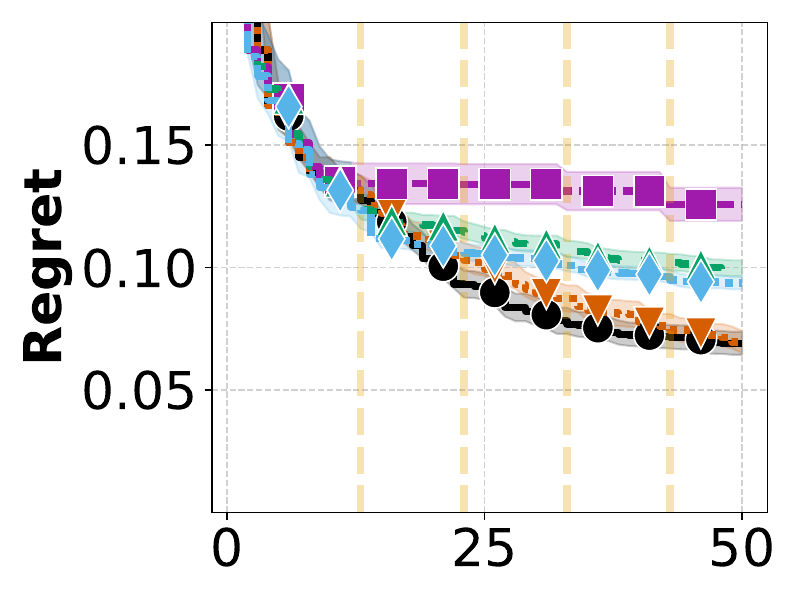} &
    \includegraphics[height=\figheight+0.13\figheight,trim={3.55cm 0.35cm 0cm 0.35cm},clip]{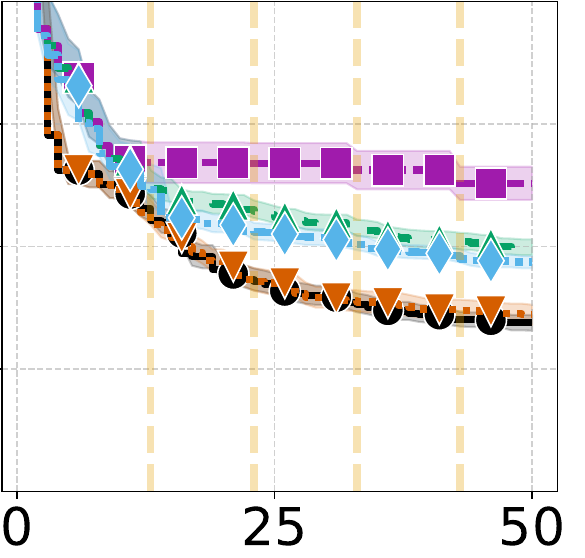} &
    \includegraphics[height=\figheight+0.13\figheight,trim={3.55cm 0.35cm 0cm 0.35cm},clip]{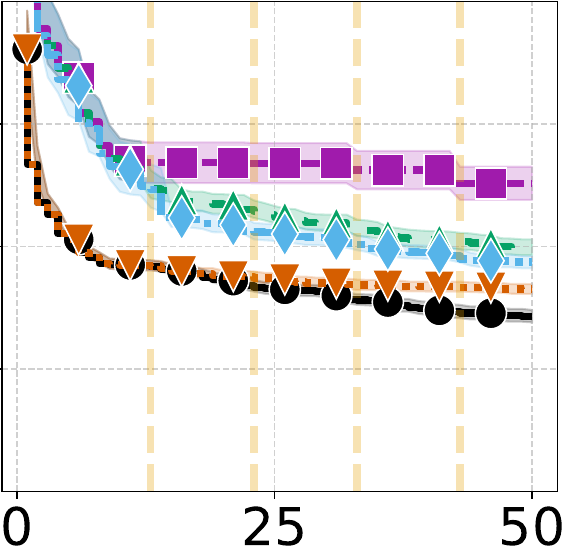} &
    \includegraphics[height=\figheight+0.13\figheight,trim={3.55cm 0.35cm 0cm 0.35cm},clip]{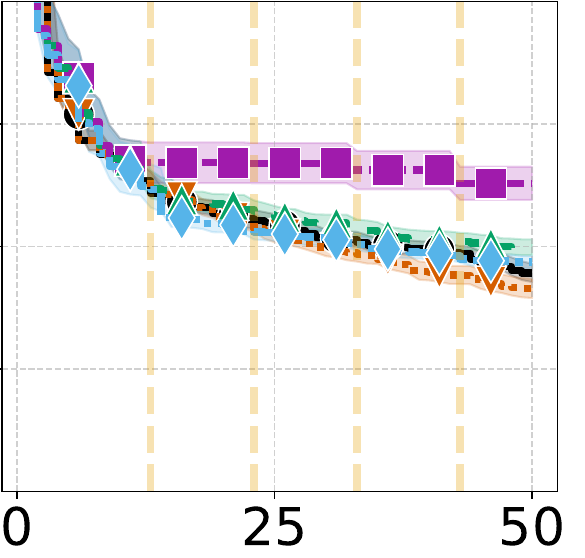} &
  \end{tabular}}

    \begin{minipage}{\textwidth}
    \centering
    \vspace{-.2cm}
    \includegraphics[width=.3\textwidth,trim={3cm 0cm .3cm 14cm},clip]{figures/iclr_submission/final_results/number_of_evaluations.pdf}
  \end{minipage}

    \begin{minipage}{\textwidth}
    \centering
    \includegraphics[width=.7\textwidth,trim={0cm 0cm 0cm 10cm},clip]{figures/ICML_26_submission/final_results/pc_comparison/overall/legend.pdf}
  \end{minipage}
  \label{fig:detailed_pc_comparison}
  \caption{Comparison of vanilla-BO, \tool, and different variants of probabilistic circuits on \texttt{PD1} using Expert, Advanced, Local, and Deceptive priors. The plots show the mean regret over time with standard errors visualized as shaded areas. Priors are provided at the dashed vertical lines.}
\end{figure}

\paragraph{Discussion of the Overall Results} Our results show that both \tool and vanilla-BO outperform PCs. The difference becomes even more pronounced when GPs are considered. This contrasts with the results reported by \citep{seng-automlconf25a}. We believe that this is due to several reasons. Firstly, \citep{seng-automlconf25a} evaluates vanilla-BO and \pibo using an older version of SMAC3 with bugs and inferior defaults. Secondly, our prior setup differs in two ways: (a) we do not provide priors directly on the optimum, but as described in \cref{sec:experiment-setup}, (b) their distribution prior uses only uniform priors, effectively shrinking the search space to the region around the found optimum. Lastly, we focus on different optimization problems. It is important to note that \citep{seng-automlconf25a} already indicated an exploration issue of PCs, and we believe that this was partially mitigated by their prior setup.

A detailed discussion of the reported results is provided below: We first focus on the different prior kinds, and then explicitly discuss pointwise priors. 

\paragraph{Comparison of Vanilla-BO and PCs Without Priors}
A comparison of vanilla-BO and PCs without priors shows that in our experimental evaluation, vanilla-BO outperforms plain probabilistic circuits for all scenarios. Importantly, even the initial Sobol design is competitive with PCs after 50 trials.

\paragraph{Impact of Priors on PCs Optimization Behavior} 
It appears PCs benefit from the addition of priors, regardless of the prior type. Interestingly, the effect of priors does not align with the effect observed for \tool, i.e., local priors lead to a larger performance boost, and even deceptive priors improve performance. This is further discussed below.

\paragraph{Discussion of Expert and Advanced Priors}
Even though advanced priors are of lower quality, they yield slightly larger performance improvements than expert priors. We hypothesize that lower-quality priors lead to more exploration than locally optimal expert priors.

\paragraph{Discussion of Local Priors} We believe the superior quality of local priors is caused by the poor performance of the initial design. Remember that expert and advanced priors use clusters with medoids that perform better than the current incumbent. As a result, the cluster is chosen relative to the current incumbent's performance, meaning that it is chosen when the initial optimization failed to find a strong incumbent; in this case, the priors are of decreased quality. Since local priors consider clusters in the local neighborhood and greedily select the best performing one, they may, in some cases, be of higher quality. On \texttt{xformer}, they even outperform vanilla-BO and \tool (if equipped with random forests). 

\paragraph{Discussion of Deceptive Priors} Since all priors are constructed based on vanilla-BO runs, and the initial design of vanilla-BO outperforms PCs without priors, even deceptive priors might be locally helpful. Additionally, they might help with the PC exploration issue.

\paragraph{Pointwise Priors} Equipping PCs with point priors allows them the unfair advantage of using the provided prior center directly. However, while this improves PC performance, PCs still perform inferior to \tool on all but one benchmark.

\newpage

\subsection{Gaussian Process Main Results}\label{app:further-results:gp-normal}
Generally, utilizing GPs as a surrogate produces the same results, as reported in \cref{sec:results:pibo} and \cref{fig:main-results}. The only outlier is the scenario \texttt{cifar100\_wideresnet\_2048}. Here, the baseline, equipped with GP as a surrogate, performs very well, resulting in no gain due to the addition of priors. However, on the other three scenarios, the \tool outperforms the competitors by a large margin.

\begin{figure}[h]
  \centering
  \setlength{\figheight}{0.18\textwidth}
  
  \resizebox{\textwidth}{!}{
  \begin{tabular}{@{}c@{}c@{}c@{}c@{}c@{}c}
    & \hspace{.6cm}\texttt{widernet} & \texttt{resnet} & \texttt{transf} & \texttt{xformer}\\ 
    \rotatebox{90}{\texttt{\hspace{.4cm} Expert}} &
    \includegraphics[height=\figheight,trim={0.35cm 1.45cm 0cm 0.35cm},clip]{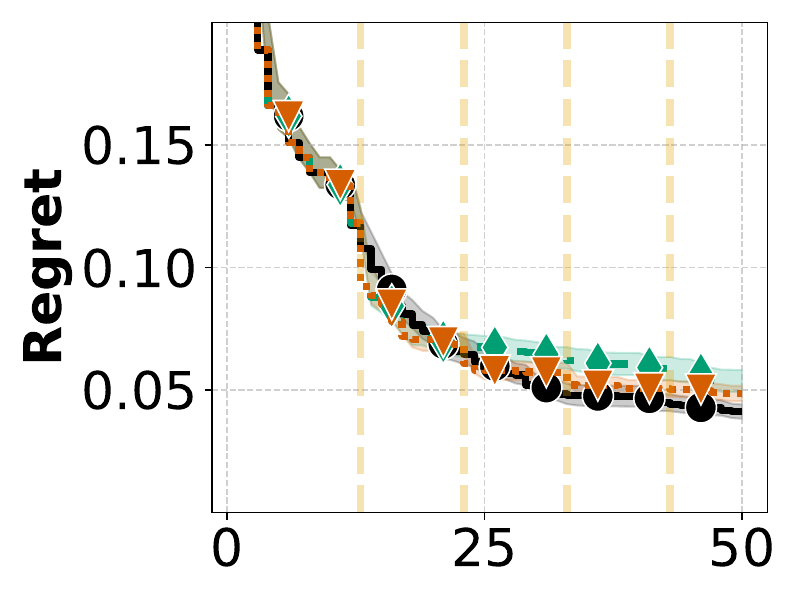} &
    \includegraphics[height=\figheight,trim={3.55cm 1.45cm 0cm 0.35cm},clip]{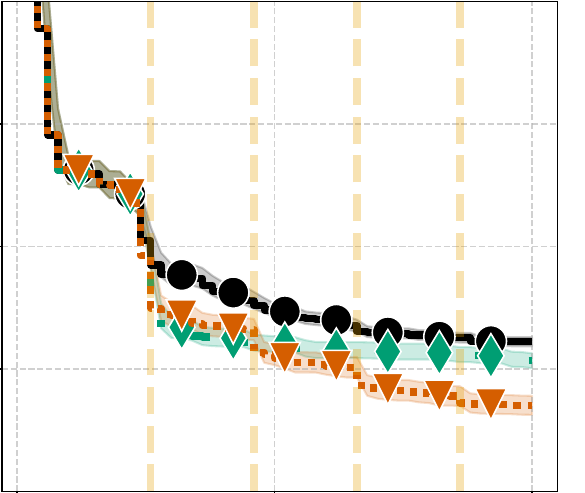} &
    \includegraphics[height=\figheight,trim={3.55cm 1.45cm 0cm 0.35cm},clip]{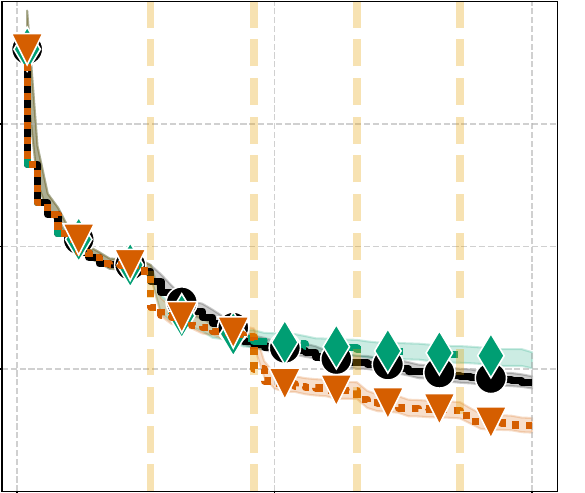} &
    \includegraphics[height=\figheight,trim={3.55cm 1.45cm 0cm 0.35cm},clip]{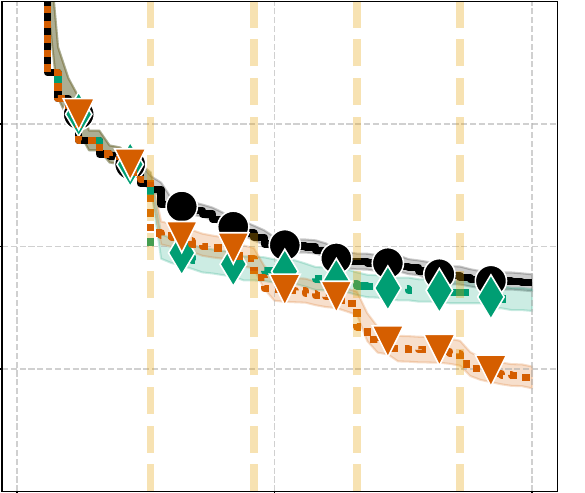} &
    \\

    \rotatebox{90}{\hspace{.2cm}\texttt{Advanced}} &
    \includegraphics[height=\figheight,trim={0.35cm 1.45cm 0cm 0.35cm},clip]{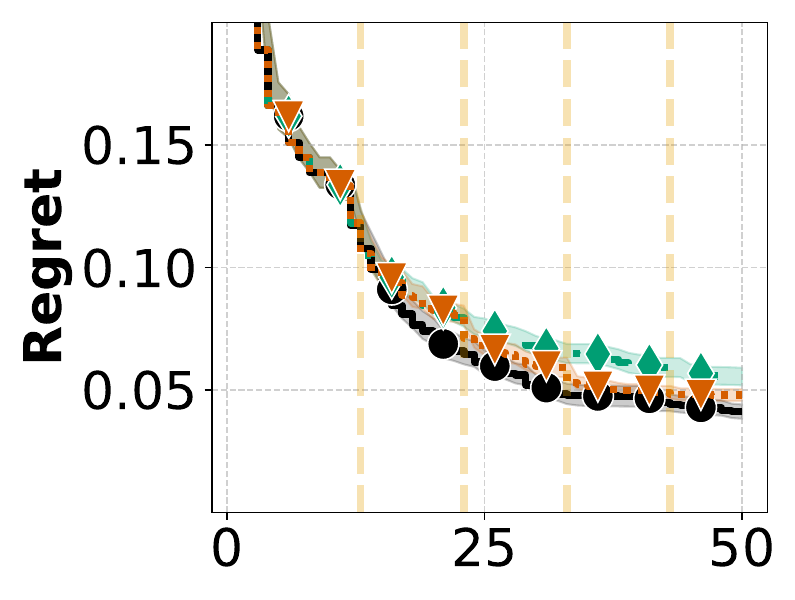} &
    \includegraphics[height=\figheight,trim={3.55cm 1.45cm 0cm 0.35cm},clip]{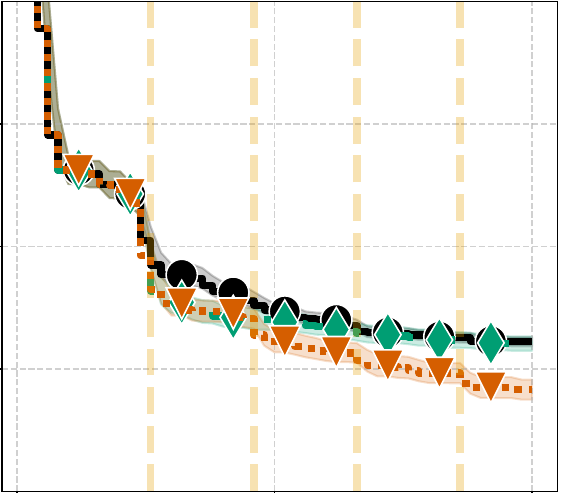} &
    \includegraphics[height=\figheight,trim={3.55cm 1.45cm 0cm 0.35cm},clip]{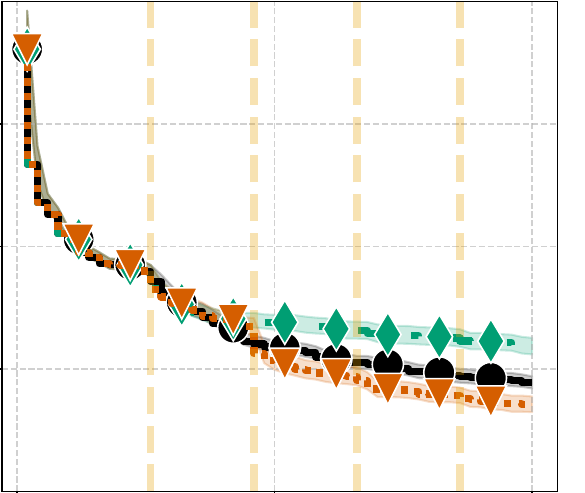} &
    \includegraphics[height=\figheight,trim={3.55cm 1.45cm 0cm 0.35cm},clip]{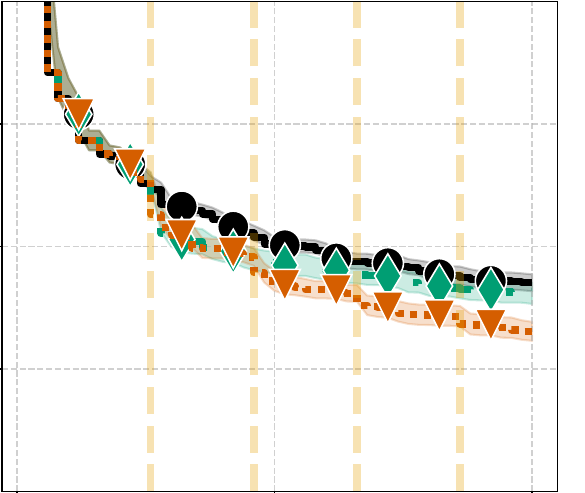} &
    \\

    \rotatebox{90}{\hspace{.75cm}\texttt{Local}} &
    \includegraphics[height=\figheight,trim={0.35cm 1.45cm 0cm 0.35cm},clip]{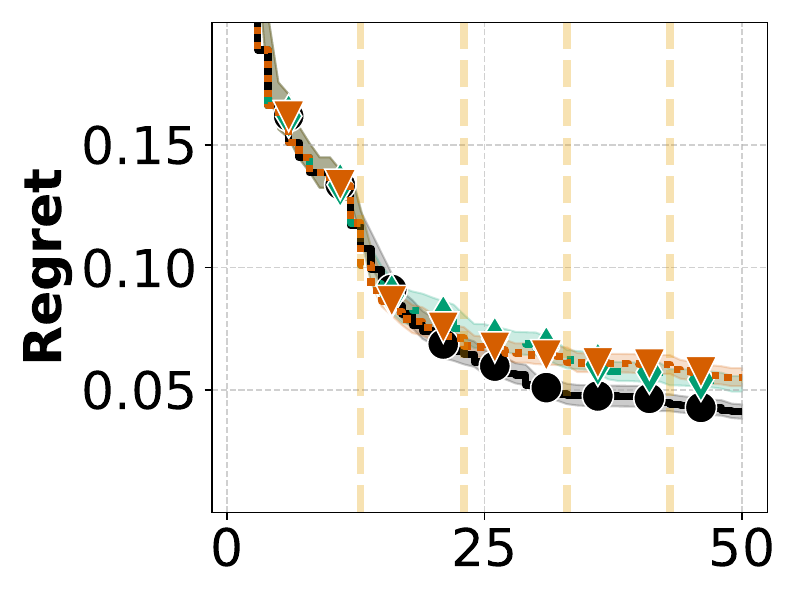} &
    \includegraphics[height=\figheight,trim={3.55cm 1.45cm 0cm 0.35cm},clip]{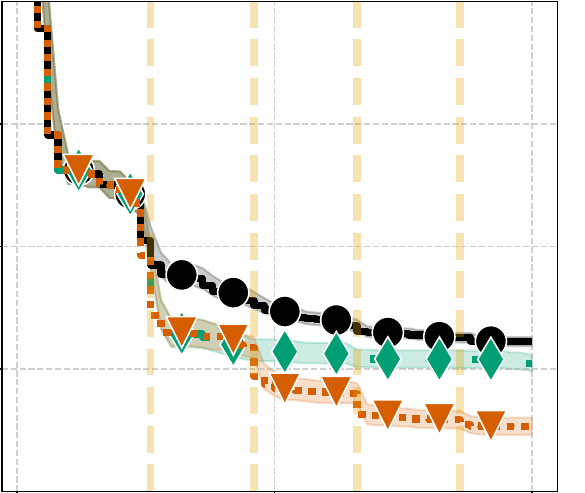} &
    \includegraphics[height=\figheight,trim={3.55cm 1.45cm 0cm 0.35cm},clip]{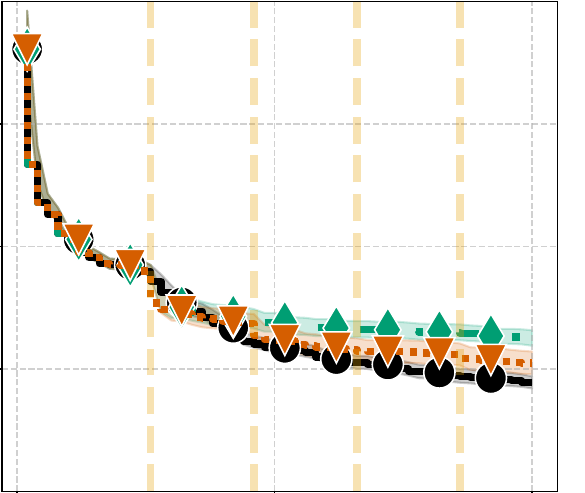} &
    \includegraphics[height=\figheight,trim={3.55cm 1.45cm 0cm 0.35cm},clip]{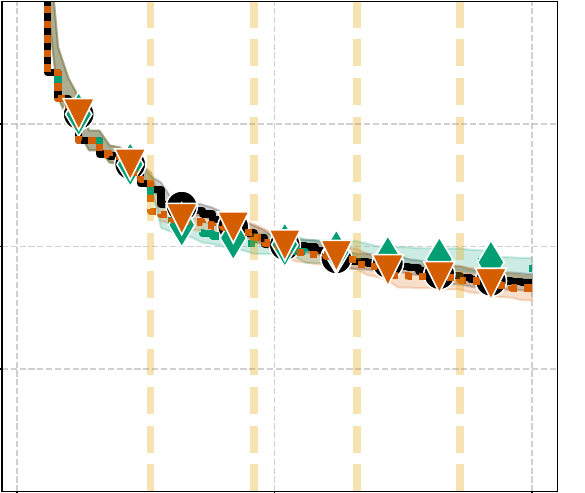} &
    \\

    \rotatebox{90}{\hspace{.3cm}\texttt{Deceptive}} &
    \includegraphics[height=\figheight+0.13\figheight,trim={0.35cm 0.35cm 0cm 0.35cm},clip]{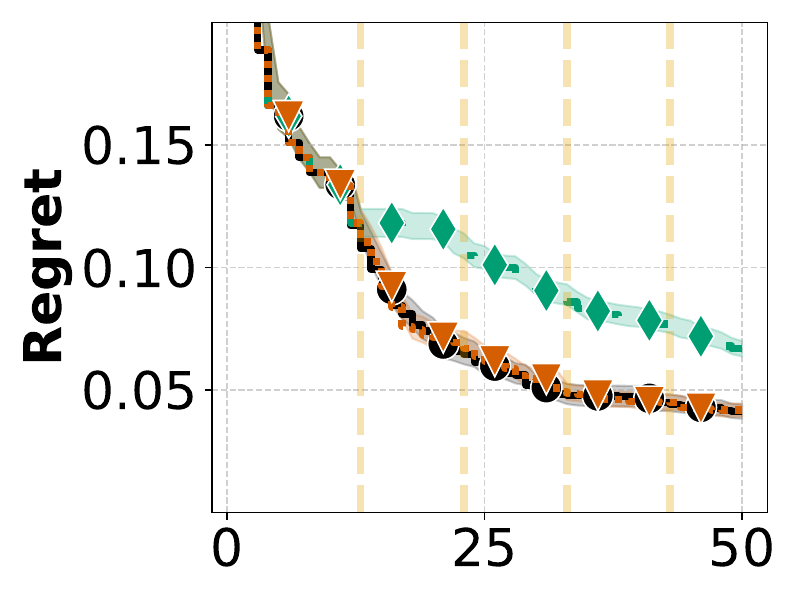} &
    \includegraphics[height=\figheight+0.13\figheight,trim={3.55cm 0.35cm 0cm 0.35cm},clip]{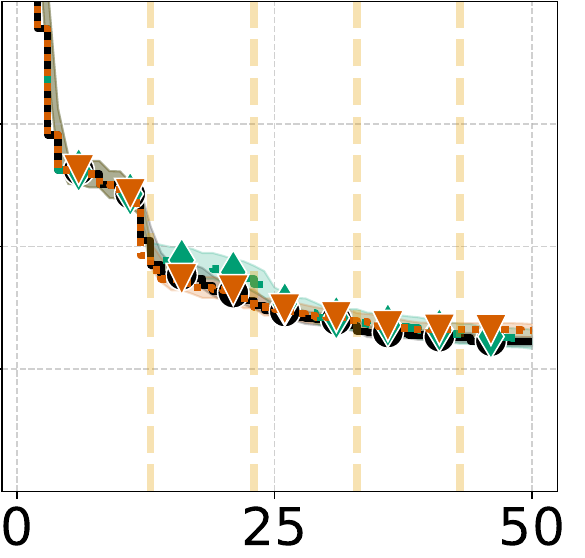} &
    \includegraphics[height=\figheight+0.13\figheight,trim={3.55cm 0.35cm 0cm 0.35cm},clip]{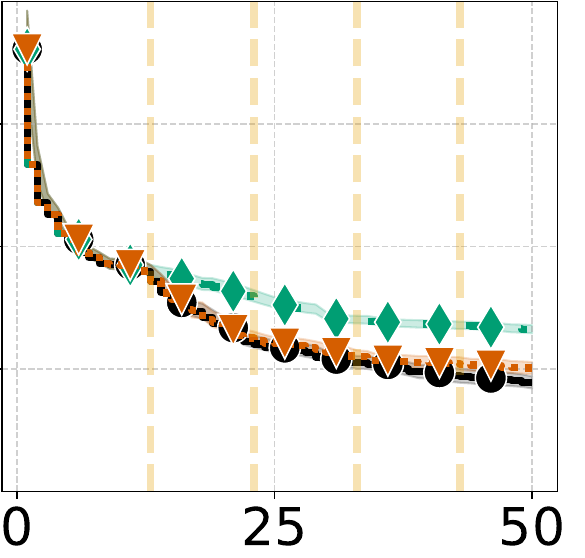} &
    \includegraphics[height=\figheight+0.13\figheight,trim={3.55cm 0.35cm 0cm 0.35cm},clip]{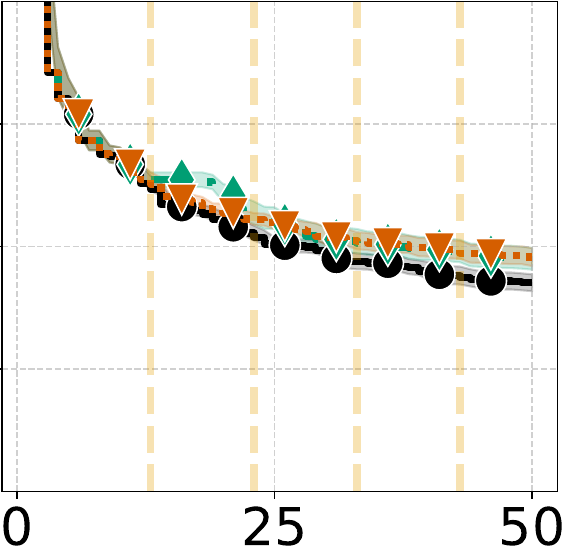} &
  \end{tabular}}

  \begin{minipage}{\textwidth}
    \centering
    \vspace{-.2cm}
    \includegraphics[width=.3\textwidth,trim={3cm 0cm .3cm 14cm},clip]{figures/iclr_submission/final_results/number_of_evaluations.pdf}
  \end{minipage}

    \begin{minipage}{\textwidth}
    \centering
    \includegraphics[width=.5\textwidth,trim={0cm 0cm 0cm 10.3cm},clip]{figures/ICML_26_submission/legend.pdf}
    \end{minipage}

  \caption{Mean regret for \texttt{PD1} using Expert, Advanced, Local, and Deceptive priors, with Gaussian processes as surrogate models. Priors are provided at vertical lines. The shaded areas visualize the standard error. The results indicate \tool outperforming \pibo and remaining competitive with vanilla BO for deceptive priors.}
\end{figure}

\begin{figure}[H]
  \centering
  \setlength{\figheight}{0.2\textwidth}
  
  \resizebox{\textwidth}{!}{
  \begin{tabular}{@{}c@{}c@{}c@{}c@{}c@{}c}
    & \hspace{.6cm}\texttt{widernet} & \texttt{resnet} & \texttt{transf} & \texttt{xformer}\\ 
    \rotatebox{90}{\texttt{\hspace{.4cm} Expert}} &
    \includegraphics[height=\figheight,trim={0.35cm 1.45cm 0cm 0.35cm},clip]{figures/ICML_26_submission/final_results/pd1/rf/cifar100_wideresnet_2048/good.pdf} &
    \includegraphics[height=\figheight,trim={3.55cm 1.45cm 0cm 0.35cm},clip]{figures/ICML_26_submission/final_results/pd1/rf/imagenet_resnet_512/good.pdf} &
    \includegraphics[height=\figheight,trim={3.55cm 1.45cm 0cm 0.35cm},clip]{figures/ICML_26_submission/final_results/pd1/rf/lm1b_transformer_2048/good.pdf} &
    \includegraphics[height=\figheight,trim={3.55cm 1.45cm 0cm 0.35cm},clip]{figures/ICML_26_submission/final_results/pd1/rf/translatewmt_xformer_64/good.pdf} &
    \\

    \rotatebox{90}{\hspace{.2cm}\texttt{Advanced}} &
    \includegraphics[height=\figheight,trim={0.35cm 1.45cm 0cm 0.35cm},clip]{figures/ICML_26_submission/final_results/pd1/rf/cifar100_wideresnet_2048/medium.pdf} &
    \includegraphics[height=\figheight,trim={3.55cm 1.45cm 0cm 0.35cm},clip]{figures/ICML_26_submission/final_results/pd1/rf/imagenet_resnet_512/medium.pdf} &
    \includegraphics[height=\figheight,trim={3.55cm 1.45cm 0cm 0.35cm},clip]{figures/ICML_26_submission/final_results/pd1/rf/lm1b_transformer_2048/medium.pdf} &
    \includegraphics[height=\figheight,trim={3.55cm 1.45cm 0cm 0.35cm},clip]{figures/ICML_26_submission/final_results/pd1/rf/translatewmt_xformer_64/medium.pdf} &
    \\

    \rotatebox{90}{\hspace{.75cm}\texttt{Local}} &
    \includegraphics[height=\figheight,trim={0.35cm 1.45cm 0cm 0.35cm},clip]{figures/ICML_26_submission/final_results/pd1/rf/cifar100_wideresnet_2048/misleading.pdf} &
    \includegraphics[height=\figheight,trim={3.55cm 1.45cm 0cm 0.35cm},clip]{figures/ICML_26_submission/final_results/pd1/rf/imagenet_resnet_512/misleading.pdf} &
    \includegraphics[height=\figheight,trim={3.55cm 1.45cm 0cm 0.35cm},clip]{figures/ICML_26_submission/final_results/pd1/rf/lm1b_transformer_2048/misleading.pdf} &
    \includegraphics[height=\figheight,trim={3.55cm 1.45cm 0cm 0.35cm},clip]{figures/ICML_26_submission/final_results/pd1/rf/translatewmt_xformer_64/misleading.pdf} &
    \\

    \rotatebox{90}{\hspace{.3cm}\texttt{Deceptive}} &
    \includegraphics[height=\figheight+0.13\figheight,trim={0.35cm 0.35cm 0cm 0.35cm},clip]{figures/ICML_26_submission/final_results/pd1/rf/cifar100_wideresnet_2048/deceiving.pdf} &
    \includegraphics[height=\figheight+0.13\figheight,trim={3.55cm 0.35cm 0cm 0.35cm},clip]{figures/ICML_26_submission/final_results/pd1/rf/imagenet_resnet_512/deceiving.pdf} &
    \includegraphics[height=\figheight+0.13\figheight,trim={3.55cm 0.35cm 0cm 0.35cm},clip]{figures/ICML_26_submission/final_results/pd1/rf/lm1b_transformer_2048/deceiving.pdf} &
    \includegraphics[height=\figheight+0.13\figheight,trim={3.55cm 0.35cm 0cm 0.35cm},clip]{figures/ICML_26_submission/final_results/pd1/rf/translatewmt_xformer_64/deceiving.pdf} &
  \end{tabular}}

    \begin{minipage}{\textwidth}
    \centering
    \vspace{-.2cm}
    \includegraphics[width=.3\textwidth,trim={3cm 0cm .3cm 14cm},clip]{figures/iclr_submission/final_results/number_of_evaluations.pdf}
  \end{minipage}

    \begin{minipage}{\textwidth}
    \centering
    \includegraphics[width=.5\textwidth,trim={0cm 0cm 0cm 10.3cm},clip]{figures/ICML_26_submission/legend.pdf}
  \end{minipage}

  \caption{Mean regret for \texttt{PD1} using Expert, Advanced, Local, and Deceptive priors, with random forests as surrogate models. Priors are provided at vertical lines. The shaded areas visualize the standard error. The results indicate \tool outperforming \pibo and remaining competitive with vanilla BO for deceptive priors.}
\end{figure}

\newpage

\subsection{Random Prior Location}\label{app:further-results:dynamic}
For experiments with randomly chosen prior locations, we model user behavior as follows. Each user provides an initial prior at the start of the optimization. If the last prior was given at time $t_i$, a new prior is provided at time $m$ with probability:  
\[
\mathbb{P}_{t_i}(\pi^{(m)}) = 1 - e^{-0.15 \cdot (m - t_i)}.
\]  
The resulting outcomes exhibit trends consistent with those observed in the main experiments.

\begin{figure}[h]
  \centering
  \resizebox{\textwidth}{!}{
  \setlength{\figheight}{0.2\textwidth}
  \begin{tabular}{@{}c@{}c@{}c@{}c@{}c@{}c}
    & \hspace{.6cm}\texttt{widernet} & \texttt{resnet} & \texttt{transf} & \texttt{xformer}\\ 
    \rotatebox{90}{\hspace{.4cm} Expert} &
    \includegraphics[height=\figheight,trim={0.35cm 1.45cm 0cm 0.35cm},clip]{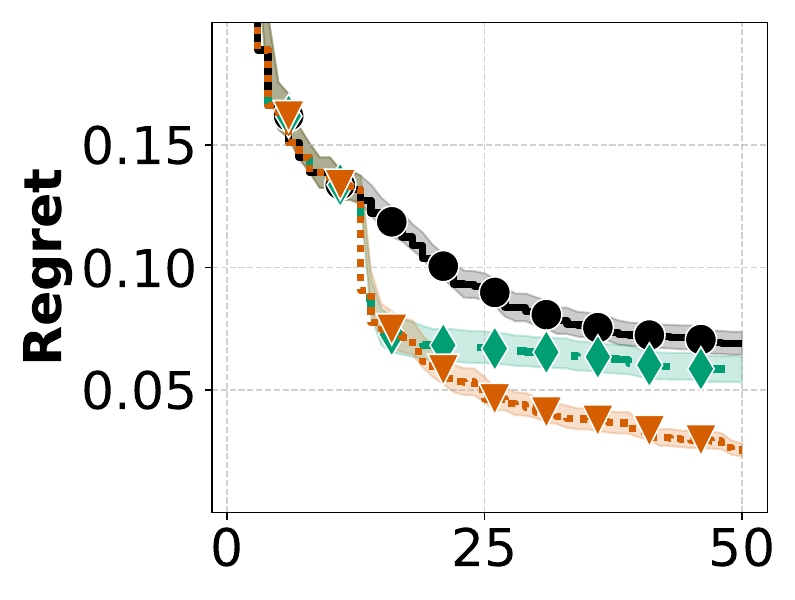} &
    \includegraphics[height=\figheight,trim={3.55cm 1.45cm 0cm 0.35cm},clip]{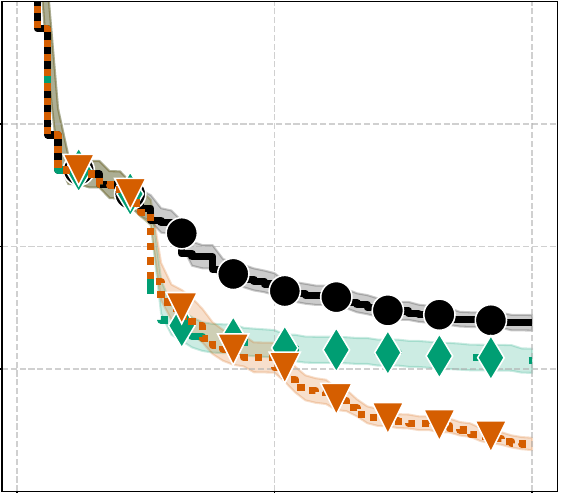} &
    \includegraphics[height=\figheight,trim={3.55cm 1.45cm 0cm 0.35cm},clip]{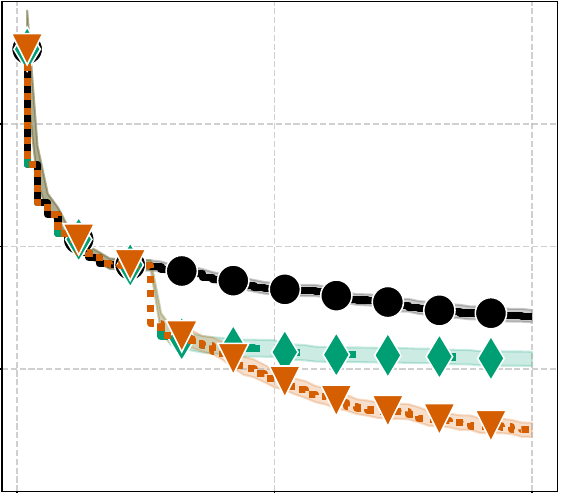} &
    \includegraphics[height=\figheight,trim={3.55cm 1.45cm 0cm 0.35cm},clip]{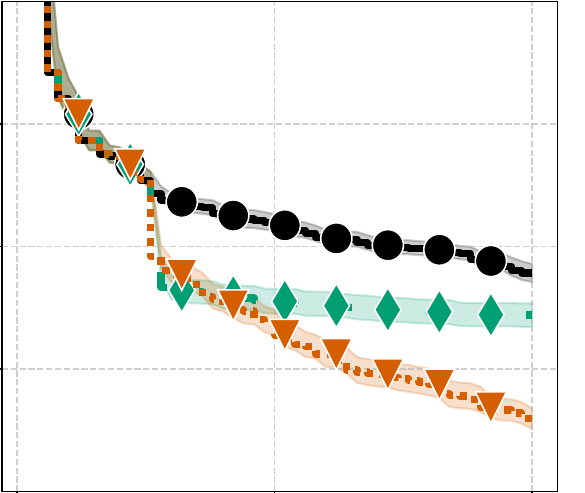} &
    \\

    \rotatebox{90}{\hspace{.2cm}\texttt{Advanced}} &
    \includegraphics[height=\figheight,trim={0.35cm 1.45cm 0cm 0.35cm},clip]{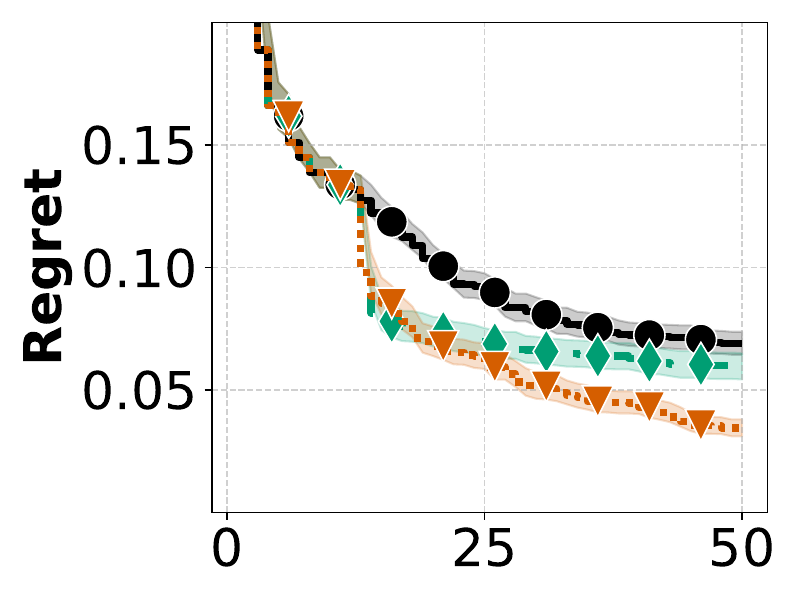} &
    \includegraphics[height=\figheight,trim={3.55cm 1.45cm 0cm 0.35cm},clip]{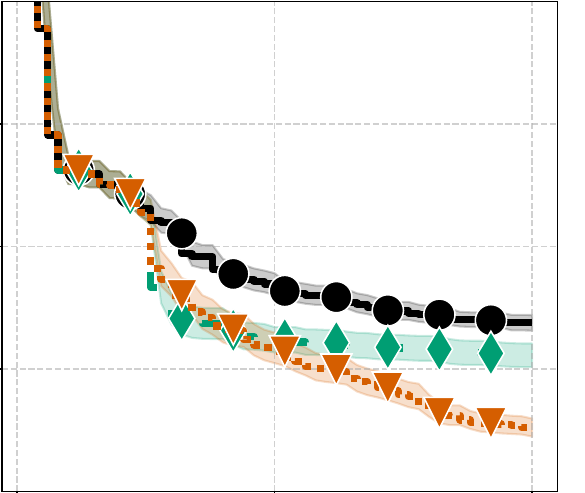} &
    \includegraphics[height=\figheight,trim={3.55cm 1.45cm 0cm 0.35cm},clip]{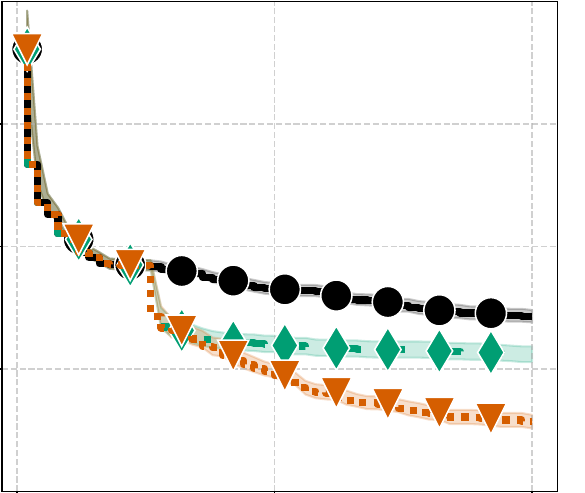} &
    \includegraphics[height=\figheight,trim={3.55cm 1.45cm 0cm 0.35cm},clip]{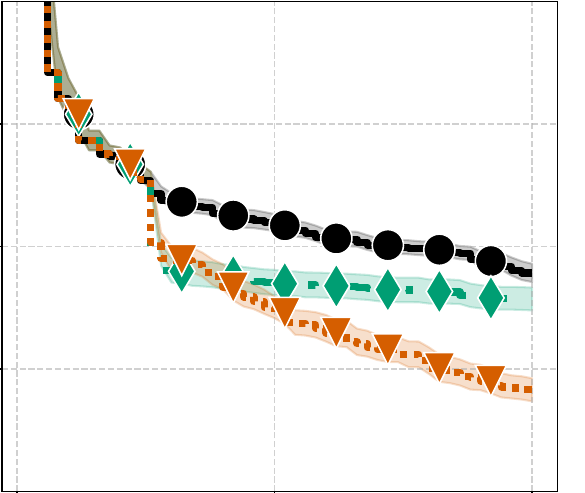} &
    \\

    \rotatebox{90}{\hspace{.75cm}\texttt{Local}} &
    \includegraphics[height=\figheight,trim={0.35cm 1.45cm 0cm 0.35cm},clip]{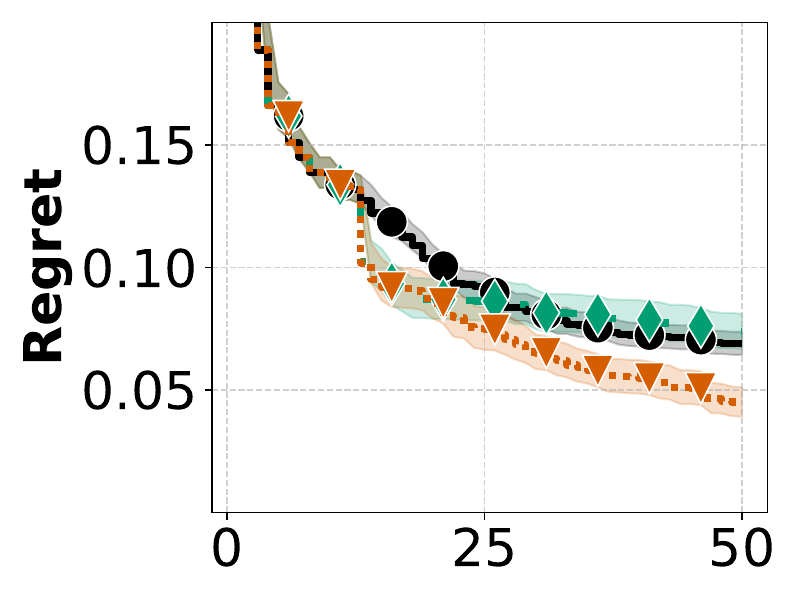} &
    \includegraphics[height=\figheight,trim={3.55cm 1.45cm 0cm 0.35cm},clip]{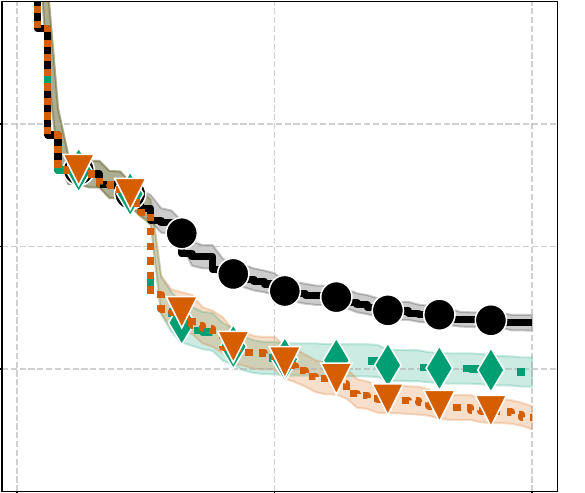} &
    \includegraphics[height=\figheight,trim={3.55cm 1.45cm 0cm 0.35cm},clip]{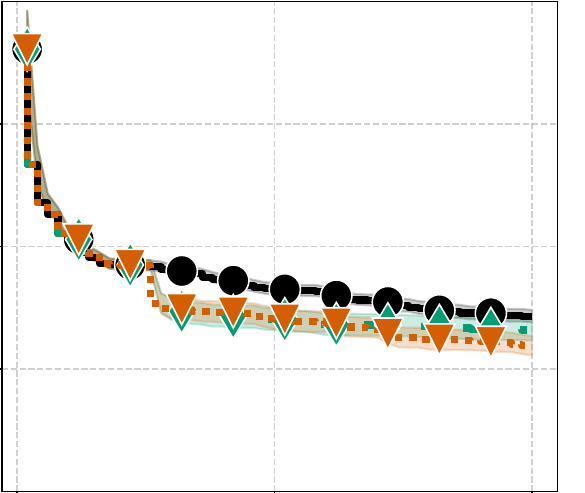} &
    \includegraphics[height=\figheight,trim={3.55cm 1.45cm 0cm 0.35cm},clip]{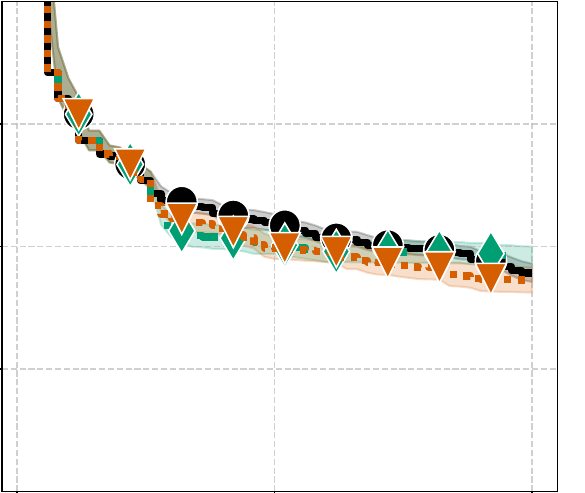} &
    \\

    \rotatebox{90}{\hspace{.3cm}\texttt{Deceptive}} &
    \includegraphics[height=\figheight+0.13\figheight,trim={0.35cm 0.35cm 0cm 0.35cm},clip]{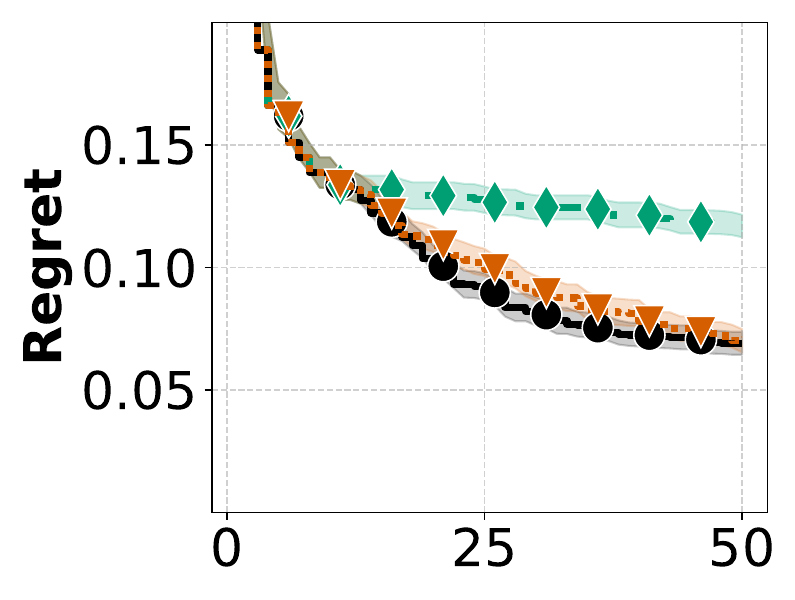} &
    \includegraphics[height=\figheight+0.13\figheight,trim={3.55cm 0.35cm 0cm 0.35cm},clip]{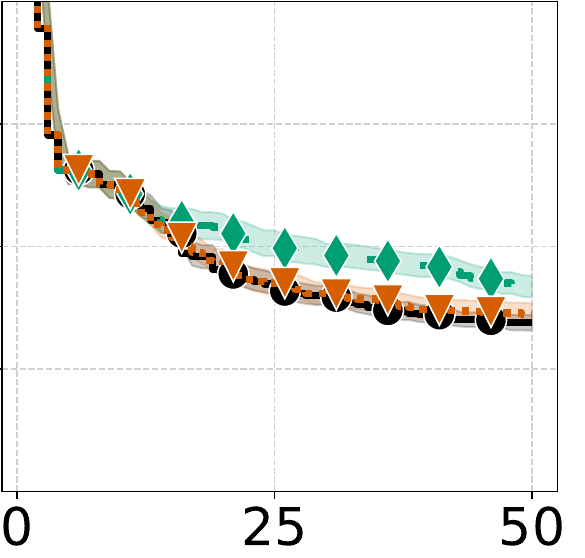} &
    \includegraphics[height=\figheight+0.13\figheight,trim={3.55cm 0.35cm 0cm 0.35cm},clip]{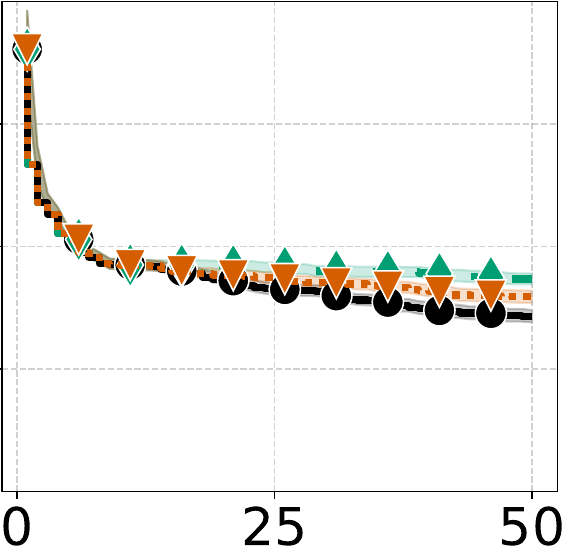} &
    \includegraphics[height=\figheight+0.13\figheight,trim={3.55cm 0.35cm 0cm 0.35cm},clip]{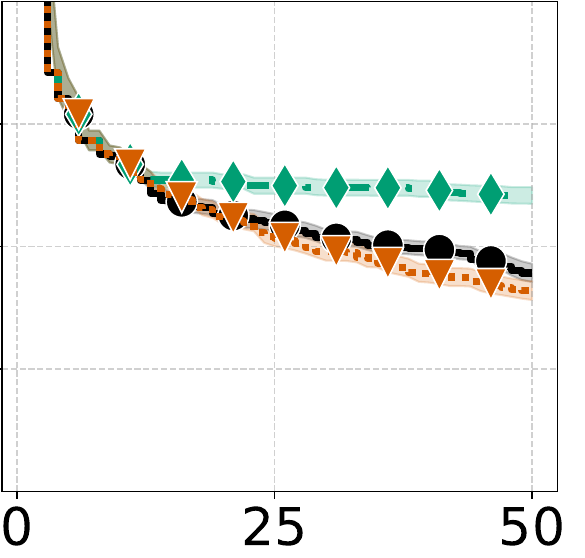} &
  \end{tabular}}

    \begin{minipage}{\textwidth}
    \centering
    \vspace{-.2cm}
    \includegraphics[width=.3\textwidth,trim={3cm 0cm .3cm 14cm},clip]{figures/iclr_submission/final_results/number_of_evaluations.pdf}
  \end{minipage}
    \begin{minipage}{\textwidth}
    \centering
    \includegraphics[width=.5\textwidth,trim={0cm 0cm 0cm 10.3cm},clip]{figures/ICML_26_submission/legend.pdf}
  \end{minipage}

  \caption{{Mean regret for \texttt{PD1} using Expert, Advanced, Local, and Deceptive priors, with random forests as surrogate models. The shaded areas visualize the standard error. The results indicate \tool outperforming \pibo and remaining competitive with vanilla BO for deceptive priors.}}
\end{figure}
\FloatBarrier

\newpage

\subsection{Na{\"i}ve Dynamic Extension for $\pi$BO}\label{pibo_extension}

\begin{figure}[h]
  \centering
  \resizebox{\textwidth}{!}{
  \setlength{\figheight}{0.2\textwidth}
  \begin{tabular}{@{}c@{}c@{}c@{}c@{}c@{}c}
    & \hspace{.6cm}\texttt{widernet} & \texttt{resnet} & \texttt{transf} & \texttt{xformer}\\ 
    \rotatebox{90}{\hspace{.4cm} Expert} &
    \includegraphics[height=\figheight,trim={0.35cm 1.45cm 0cm 0.35cm},clip]{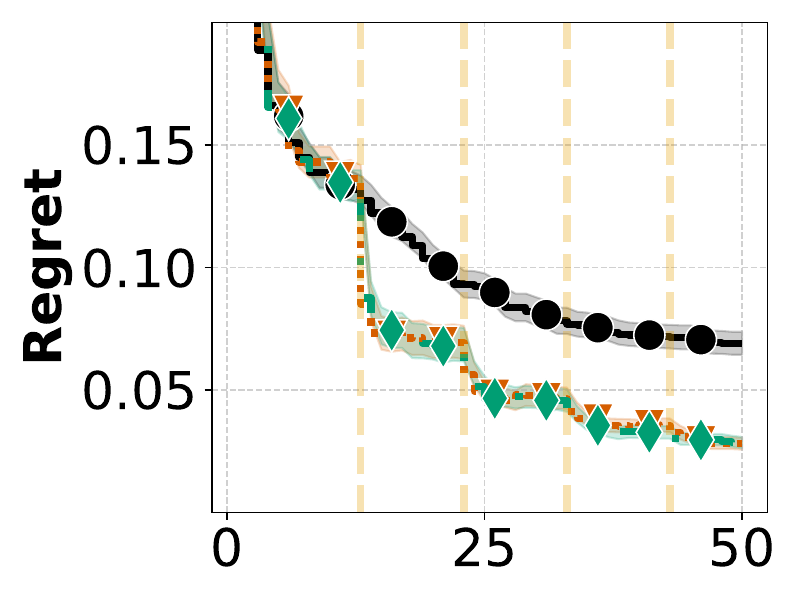} &
    \includegraphics[height=\figheight,trim={3.55cm 1.45cm 0cm 0.35cm},clip]{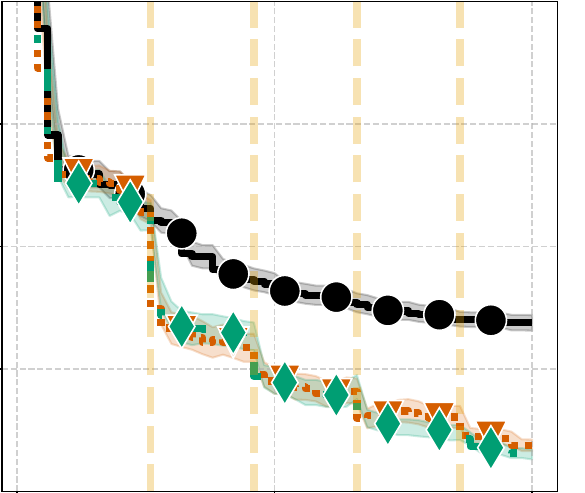} &
    \includegraphics[height=\figheight,trim={3.55cm 1.45cm 0cm 0.35cm},clip]{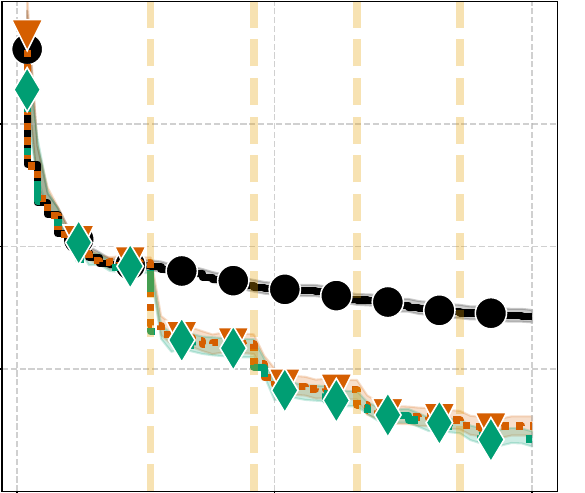} &
    \includegraphics[height=\figheight,trim={3.55cm 1.45cm 0cm 0.35cm},clip]{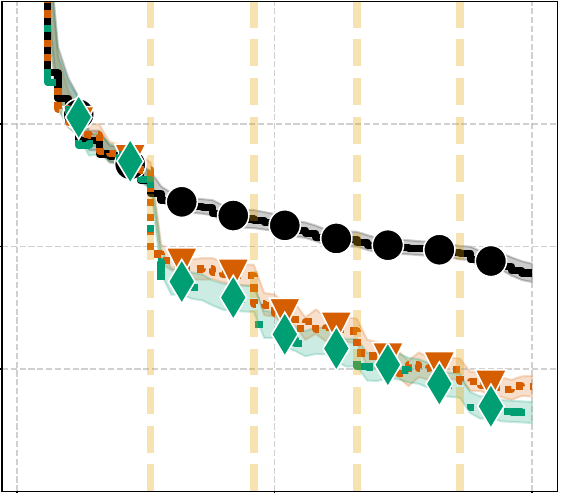} &
    \\

    \rotatebox{90}{\hspace{.2cm}\texttt{Advanced}} &
    \includegraphics[height=\figheight,trim={0.35cm 1.45cm 0cm 0.35cm},clip]{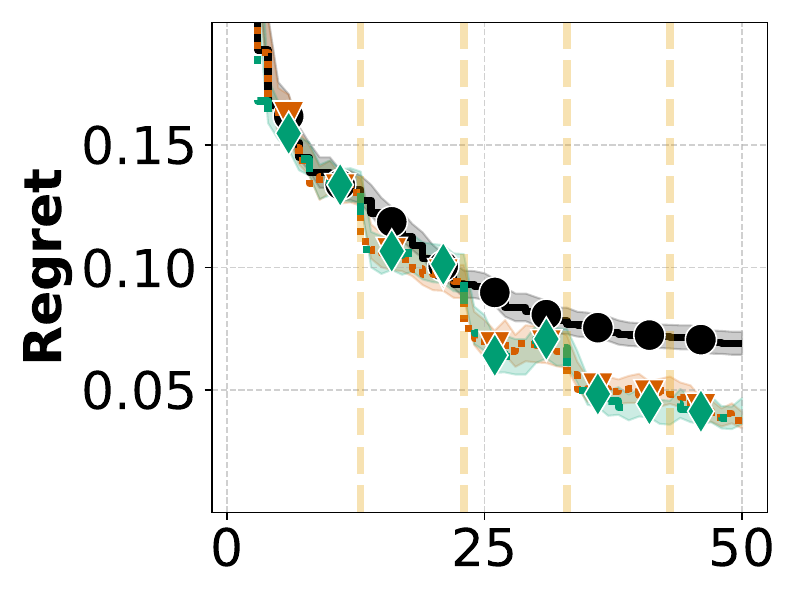} &
    \includegraphics[height=\figheight,trim={3.55cm 1.45cm 0cm 0.35cm},clip]{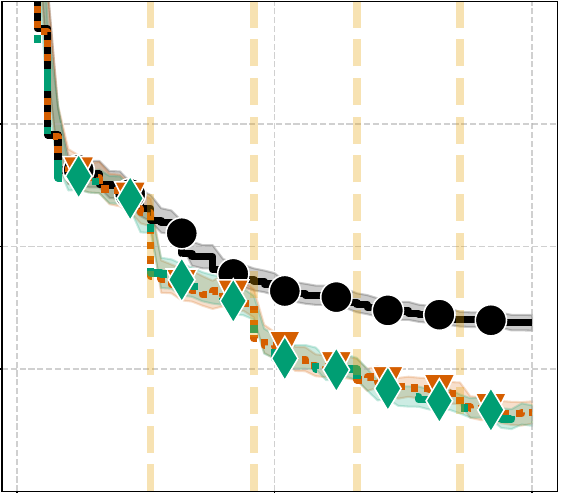} &
    \includegraphics[height=\figheight,trim={3.55cm 1.45cm 0cm 0.35cm},clip]{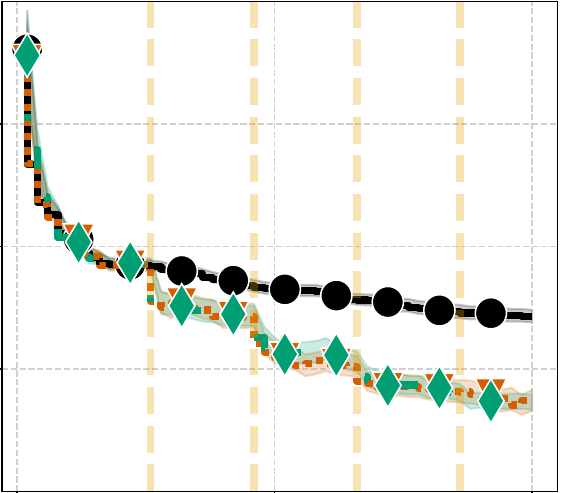} &
    \includegraphics[height=\figheight,trim={3.55cm 1.45cm 0cm 0.35cm},clip]{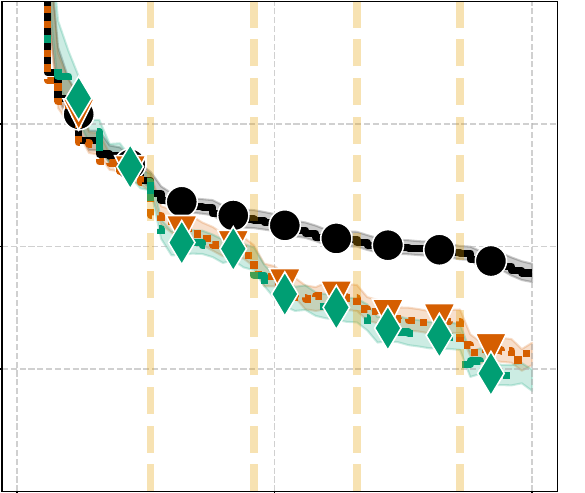} &
    \\

    \rotatebox{90}{\hspace{.75cm}\texttt{Local}} &
    \includegraphics[height=\figheight,trim={0.35cm 1.45cm 0cm 0.35cm},clip]{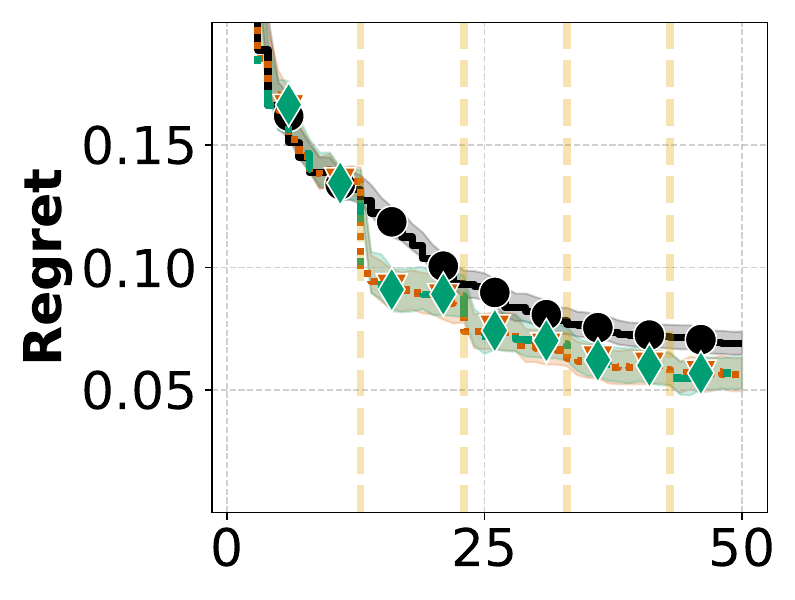} &
    \includegraphics[height=\figheight,trim={3.55cm 1.45cm 0cm 0.35cm},clip]{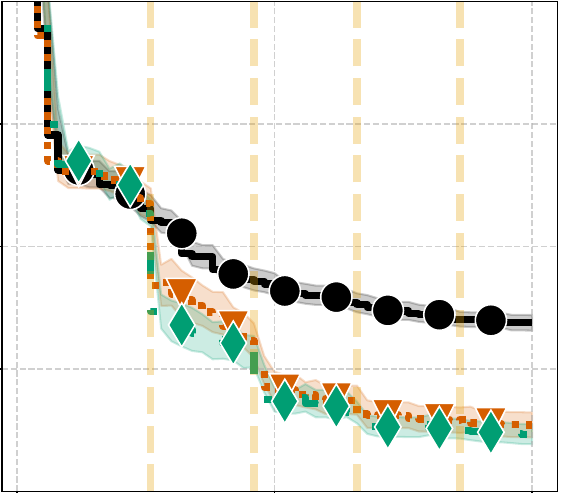} &
    \includegraphics[height=\figheight,trim={3.55cm 1.45cm 0cm 0.35cm},clip]{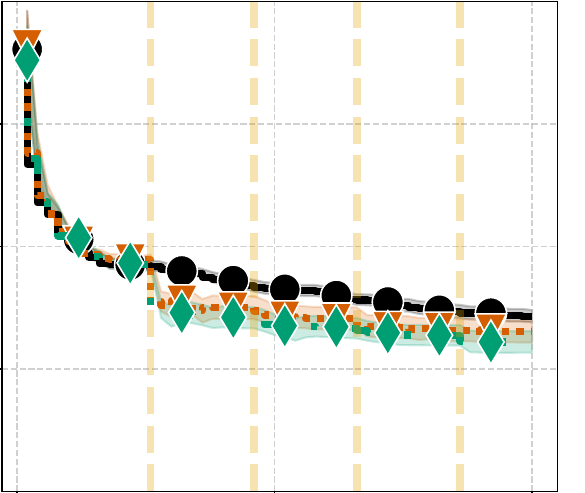} &
    \includegraphics[height=\figheight,trim={3.55cm 1.45cm 0cm 0.35cm},clip]{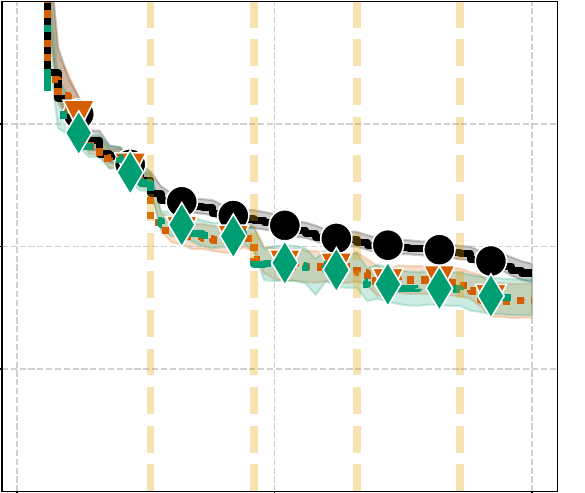} &
    \\

    \rotatebox{90}{\hspace{.3cm}\texttt{Deceptive}} &
    \includegraphics[height=\figheight+0.13\figheight,trim={0.35cm 0.35cm 0cm 0.35cm},clip]{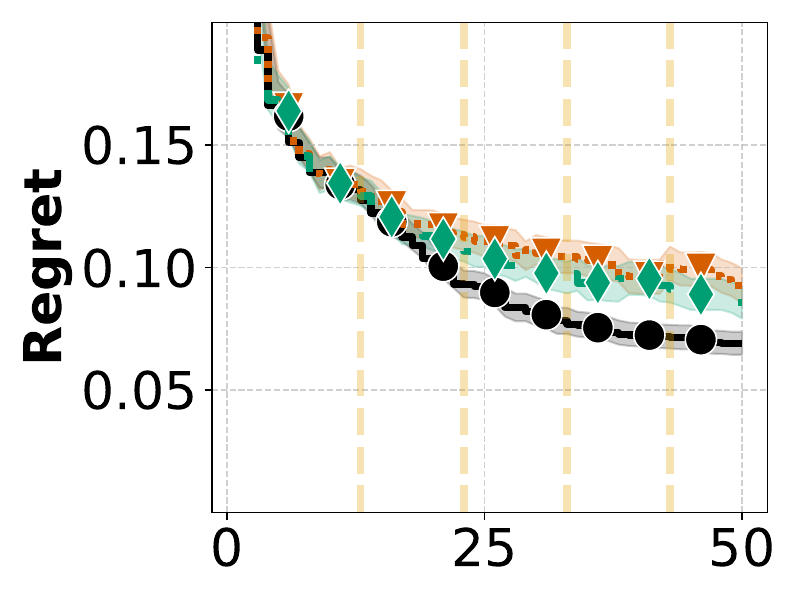} &
    \includegraphics[height=\figheight+0.13\figheight,trim={3.55cm 0.35cm 0cm 0.35cm},clip]{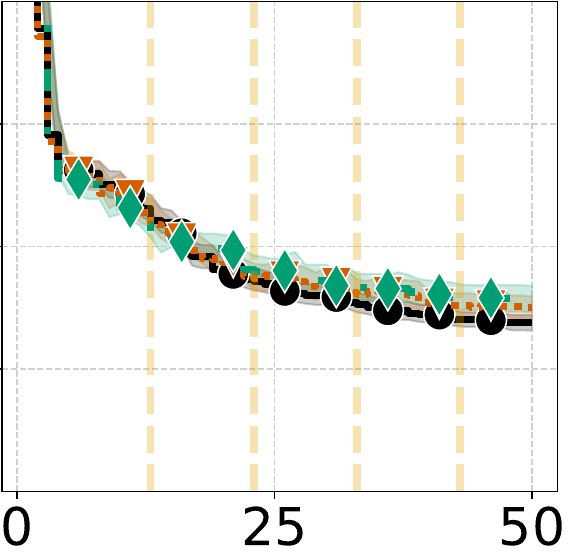} &
    \includegraphics[height=\figheight+0.13\figheight,trim={3.55cm 0.35cm 0cm 0.35cm},clip]{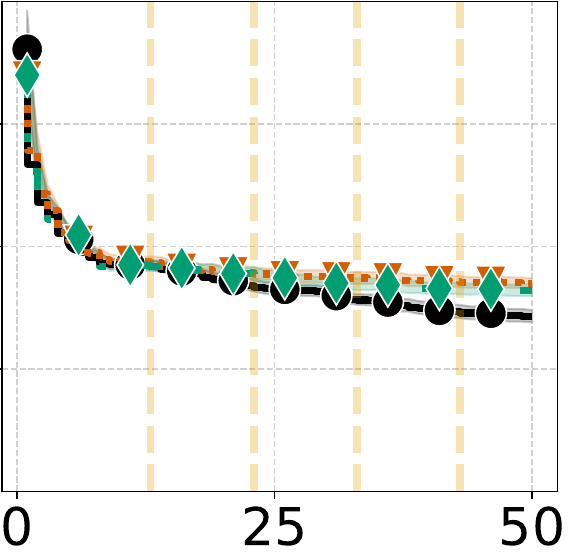} &
    \includegraphics[height=\figheight+0.13\figheight,trim={3.55cm 0.35cm 0cm 0.35cm},clip]{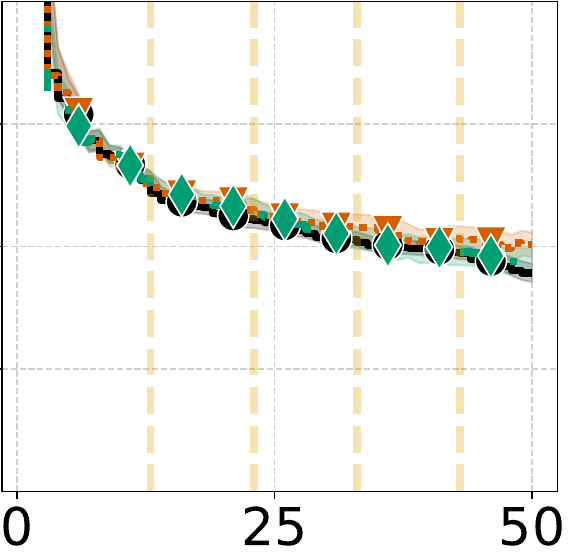} &
  \end{tabular}}

    \begin{minipage}{\textwidth}
    \centering
    \vspace{-.2cm}
    \includegraphics[width=.3\textwidth,trim={3cm 0cm .3cm 14cm},clip]{figures/iclr_submission/final_results/number_of_evaluations.pdf}
  \end{minipage}
    \begin{minipage}{\textwidth}
    \centering
    \includegraphics[width=.5\textwidth,trim={.2cm 0cm .2cm 10cm},clip]{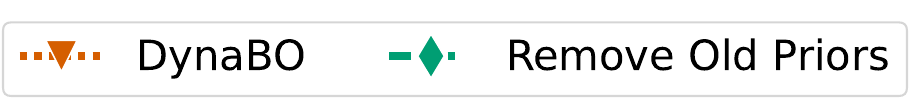}
  \end{minipage}

  \caption{{Mean regret for \texttt{PD1} using Expert, Advanced, Local, and Deceptive priors, with random forests as surrogate models. The shaded areas visualize the standard error. The results indicate \tool outperforming \pibo and remaining competitive with vanilla BO for deceptive priors.}}
  \label{fig:comparison-to-naive-dynamic-pibo}
\end{figure}

In this ablation, we compare a less sophisticated baseline of na{\"i}vely removing old priors to the proposed mechanism of \tool for summing priors. The results are visualized in Figure~\ref{fig:comparison-to-naive-dynamic-pibo}. We find that no substantial performance difference can be observed between the two methods in our standard evaluation setup. This means that there is no harm in continuing to use the old priors.

\begin{figure}[H]
    \centering

    \begin{minipage}{\textwidth}
        \centering
        \texttt{cifar100\_wideresnet\_2048}\\
        \includegraphics[width=.6\textwidth,trim={0cm 0.35cm 0cm 0cm},clip]{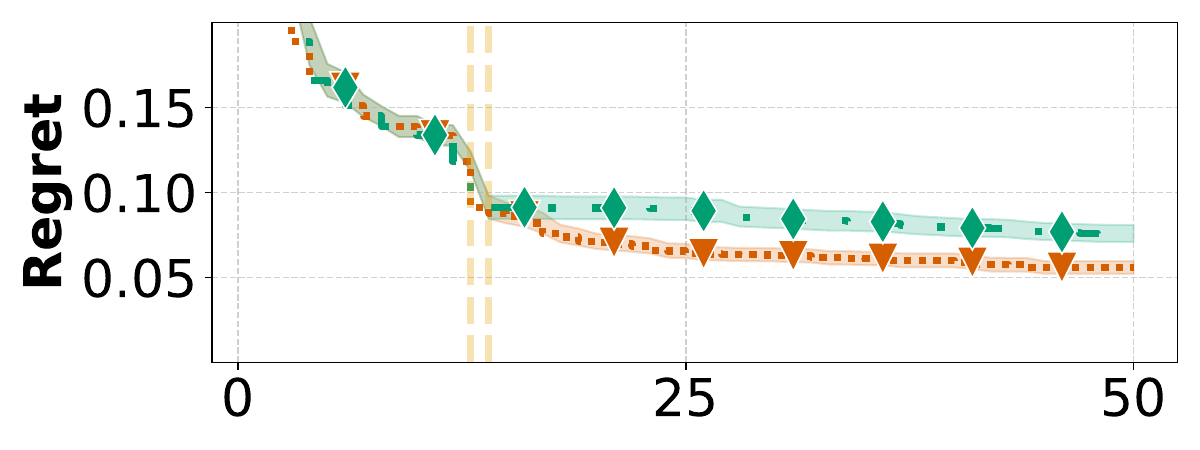}
    \end{minipage}

    \medskip

    \begin{minipage}{\textwidth}
        \centering
        \texttt{imagene\_resnet\_512}\\
        \includegraphics[width=.6\textwidth,trim={0cm 0.35cm 0cm 0cm},clip]{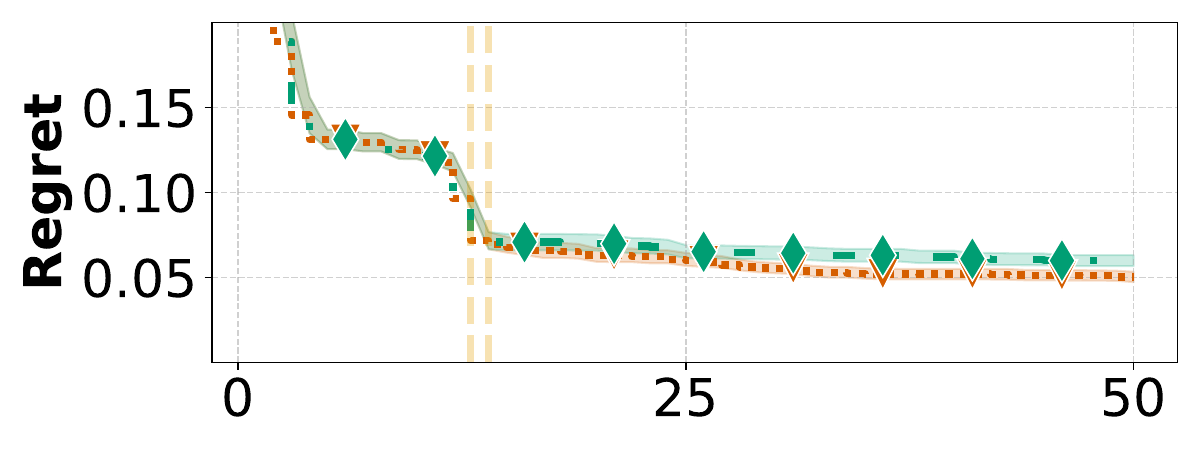}
    \end{minipage}

    \medskip

    \begin{minipage}{\textwidth}
        \centering
        \texttt{lm1b\_transformer\_2048}\\
        \includegraphics[width=.6\textwidth,trim={0cm 0.35cm 0cm 0cm},clip]{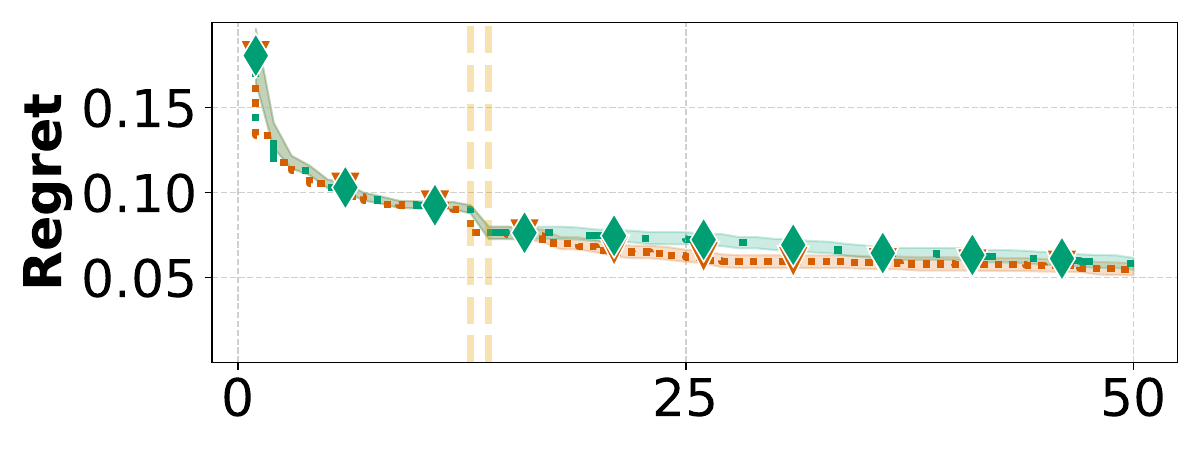}
    \end{minipage}

    \medskip

    \begin{minipage}{\textwidth}
        \centering
        \texttt{translatewmt\_xformer\_64}\\
        \includegraphics[width=.6\textwidth,trim={0cm 0.35cm 0cm 0cm},clip]{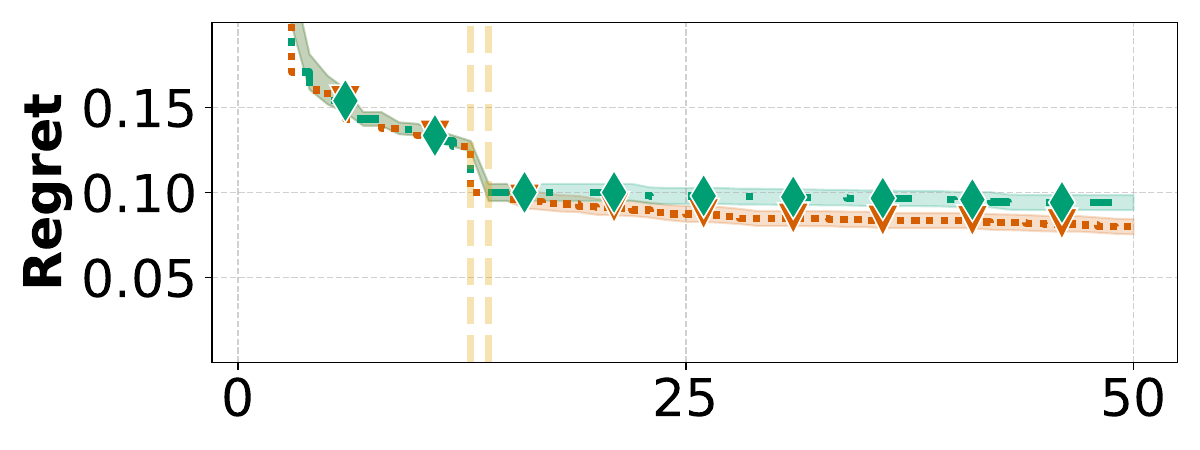}
    \end{minipage}

    \medskip

    \begin{minipage}{\textwidth}
        \centering
        \vspace{-.2cm}
        \includegraphics[width=.3\textwidth,trim={3cm 0cm .3cm 14cm},clip]{figures/iclr_submission/final_results/number_of_evaluations.pdf}
    \end{minipage}

    \begin{minipage}{\textwidth}
        \centering
        \includegraphics[width=.5\textwidth,trim={.2cm 0cm .2cm 10cm},clip]{figures/iclr_submission/mixed/legend.pdf}
    \end{minipage}

    \caption{Investigation of helpful priors followed by deceptive priors.}
    \label{fig:comparison-to-naive-dynamic-pibo-with-switching-prior-type}
\end{figure}

However, to evaluate whether \tool's intuition of old information being useful holds in practice, one has to evaluate the impact of comparing positive and negative priors. To that end, we evaluate what happens if an expert, advanced, or local prior, each provided with a chance of $\nicefrac{1}{3}$ is followed by a deceptive prior. In this setup, the quality of the two methods differs significantly. For example, when positive priors are followed by negative priors, the results show a degraded performance, as can be seen in Figure~\ref{fig:comparison-to-naive-dynamic-pibo-with-switching-prior-type}. This result holds, even though the positive prior results in an immediate performance boost for both approaches.

\FloatBarrier

\newpage

\subsection{Prior Decay Ablation} \label{app:further-results:prior-decay}

Due to providing multiple priors during the experiment, we reinvestigated the speed at which priors decay. We therefore ablated the function $\phi$ in \cref{tab:prior_decay_regret} for PD1 and \cref{tab:prior_decay_regret_yahpo} for \yahpo, respectively. For \yahpo, we only utilized 3 seeds.

\[
\af^\text{dyna}_{\surrogate}(\conf) := \af_{\surrogate}(\conf)\, \cdot \sum_{m=1}^M \prior^{(m)}(\conf)^{\beta/\phi{(t-t^{(m)})}}\, 
\]

Our investigation reveals that the optimal decay rate depends on the prior quality and the optimization scenario. In our paper, we use linear decay because it offers a good trade-off between informative and deceptive priors, but other decay strategies can be justified as well, especially for differing execution budgets.

\begin{table}[ht]
    \centering
    \caption{Mean regret ($\mu$) and standard error (SE) for each prior type across decay configurations. All values are rounded to three decimal places.}
    \begin{tabular}{lcccccccc}
    \toprule
    \textbf{Config} 
        & \multicolumn{2}{c}{\textbf{Expert}} 
        & \multicolumn{2}{c}{\textbf{Advanced}} 
        & \multicolumn{2}{c}{\textbf{Local}} 
        & \multicolumn{2}{c}{\textbf{Deceptive}} \\
    \cmidrule(lr){2-3}
    \cmidrule(lr){4-5}
    \cmidrule(lr){6-7}
    \cmidrule(lr){8-9}
    & $\mu$ & SE
    & $\mu$ & SE
    & $\mu$ & SE
    & $\mu$ & SE
    \\
    \midrule
    Logarithmic Decay      & 0.022 & 0.000 & 0.041 & 0.001 & 0.056 & 0.001 & 0.107 & 0.001 \\
    Linear Decay           & 0.025 & 0.001 & 0.040 & 0.001 & 0.056 & 0.001 & 0.102 & 0.001 \\
    Quadratic Decay        & 0.029 & 0.001 & 0.041 & 0.001 & 0.053 & 0.001 & 0.092 & 0.001 \\
    Cubic Decay            & 0.029 & 0.001 & 0.045 & 0.001 & 0.056 & 0.001 & 0.086 & 0.001 \\
    To the Power of 4 Decay & 0.032 & 0.001 & 0.049 & 0.001 & 0.054 & 0.001 & 0.088 & 0.001 \\
    To the Power of 5 Decay & 0.033 & 0.001 & 0.050 & 0.001 & 0.053 & 0.001 & 0.084 & 0.001 \\
    \bottomrule
    \end{tabular}
    \label{tab:prior_decay_regret}
\end{table}

\begin{table}[ht]
    \centering
    \caption{Mean regret ($\mu$) and standard error (SE) for each prior type across decay configurations. All values are rounded to four decimal places.}
    \begin{tabular}{lcccccccc}
        \toprule
        \textbf{Config} 
            & \multicolumn{2}{c}{\textbf{Expert}} 
            & \multicolumn{2}{c}{\textbf{Advanced}} 
            & \multicolumn{2}{c}{\textbf{Local}} 
            & \multicolumn{2}{c}{\textbf{Deceptive}} \\
        \cmidrule(lr){2-3}
        \cmidrule(lr){4-5}
        \cmidrule(lr){6-7}
        \cmidrule(lr){8-9}
        & $\mu$ & SE
        & $\mu$ & SE
        & $\mu$ & SE
        & $\mu$ & SE
        \\
        \midrule
        Logarithmic Decay       & {0.0698} & {0.0017} & {0.0830} & {0.0018} & {0.0845} & {0.0020} & 0.1334 & 0.0033 \\
        Linear Decay            & 0.0756 & 0.0018 & 0.0878 & 0.0019 & 0.0862 & 0.0020 & 0.1296 & 0.0031 \\
        Quadratic Decay         & 0.0785 & 0.0019 & 0.0866 & 0.0019 & 0.0860 & 0.0020 & 0.1140 & 0.0026 \\
        Cubic Decay             & 0.0784 & 0.0019 & 0.0836 & 0.0019 & 0.0863 & 0.0020 & 0.1089 & 0.0024 \\
        To the Power of 4 Decay & 0.0780 & 0.0019 & 0.0856 & 0.0019 & 0.0860 & 0.0021 & 0.1057 & 0.0024 \\
        To the Power of 5 Decay & 0.0795 & 0.0020 & 0.0842 & 0.0019 & 0.0849 & 0.0020 & {0.1049} & 0.0024 \\
        \bottomrule
    \end{tabular}
    \label{tab:prior_decay_regret_yahpo}
\end{table}

\subsection{Investigating the Impact of $\beta$}\label{beta}

As proposed by \citet{hvarfner-iclr22a}, we utilize $\beta = \nicefrac{N}{10}$ in our paper. Below, we ablated this choice, focusing on the PD1 benchmark. In this ablation, we consider both with and without prior rejection in \cref{tab:with-prior-rejection} and \cref{tab:without-prior-rejection}, respectively. We evaluate the same beta configurations as in \pibo on PD1. Our results indicate that one can make a case for multiple values, but $\beta=\nicefrac{N}{10}$ remains a reasonable choice, as it performs adequately for both Expert and Deceptive priors, and the best for Local priors with rejection.

\begin{table}[H]
  \centering
  \caption{With prior rejection (30 seeds, 4 scenarios, PD1)}
  \label{tab:with-prior-rejection}
  \begin{tabular}{rccccc}
    \toprule
    $\beta=\frac{N}{.}$ & Expert & Advanced & Local & Deceptive & All \\
    \midrule
    50  & $0.0329 \pm 0.0016$ & $0.0500 \pm 0.0022$ & $0.0584 \pm 0.0030$ & $0.0749 \pm 0.0018$ & $0.0540 \pm 0.0013$ \\
    25  & $0.0293 \pm 0.0018$ & $0.0499 \pm 0.0021$ & $0.0560 \pm 0.0030$ & $0.0755 \pm 0.0019$ & $0.0527 \pm 0.0014$ \\
    10  & $0.0270 \pm 0.0017$ & $0.0443 \pm 0.0021$ & $0.0565 \pm 0.0031$ & $0.0768 \pm 0.0019$ & $0.0511 \pm 0.0014$ \\
    5 & $0.0236 \pm 0.0018$ & $0.0431 \pm 0.0021$ & $0.0553 \pm 0.0033$ & $0.0770 \pm 0.0019$ & $0.0497 \pm 0.0015$ \\
    2.5 & $0.0208 \pm 0.0016$ & $0.0465 \pm 0.0023$ & $0.0589 \pm 0.0034$ & $0.0771 \pm 0.0020$ & $0.0509 \pm 0.0015$ \\
    \bottomrule
  \end{tabular}
\end{table}

\begin{table}[H]
  \centering
  \caption{Without prior rejection (30 seeds, 4 scenarios, PD1)}
  \label{tab:without-prior-rejection}
  \begin{tabular}{rccccc}
    \toprule
    $\beta=\frac{N}{.}$ & Expert & Advanced & Local & Deceptive & All \\
    \midrule
    50  & $0.0330 \pm 0.0015$ & $0.0517 \pm 0.0021$ & $0.0587 \pm 0.0030$ & $0.0957 \pm 0.0023$ & $0.0598 \pm 0.0015$ \\
    25  & $0.0287 \pm 0.0016$ & $0.0452 \pm 0.0019$ & $0.0569 \pm 0.0030$ & $0.0989 \pm 0.0023$ & $0.0574 \pm 0.0016$ \\
    10  & $0.0261 \pm 0.0017$ & $0.0411 \pm 0.0021$ & $0.0589 \pm 0.0034$ & $0.1027 \pm 0.0024$ & $0.0572 \pm 0.0018$ \\
    5 & $0.0221 \pm 0.0014$ & $0.0401 \pm 0.0022$ & $0.0599 \pm 0.0034$ & $0.1052 \pm 0.0024$ & $0.0568 \pm 0.0019$ \\
    2.5 & $0.0211 \pm 0.0015$ & $0.0385 \pm 0.0021$ & $0.0583 \pm 0.0033$ & $0.1071 \pm 0.0024$ & $0.0563 \pm 0.0019$ \\
    \bottomrule
  \end{tabular}
\end{table}
 
\subsection{Prior Rejection Sampling Budget Ablation} \label{app:further-results:rior_rejection_sampling_budget_ablation}

We ablate the number of prior rejection samples to assess the sensitivity of our method to this hyperparameter. Specifically, we compare sample budgets of 100, 500, and 1000 across all four prior types (Expert, Advanced, Local, and Deceptive) on both the \yahpo and PD1 benchmarks. We use 3 seeds for \yahpo while retaining 30 seeds for PD1. The results, reported in Tables~\ref{tab:ablation_budget_yahpo} and~\ref{tab:ablation_budget_pd1}, show that rejection decisions remain consistent across all three budgets, indicating that the exact sample count has a negligible impact on performance in practice.

\begin{table}[h]
\centering
\caption{Sensitivity analysis of the prior rejection sample budget on \yahpo (3 seeds). Values are mean regret $\pm$ standard error.}
\label{tab:ablation_budget_yahpo}
\begin{tabular}{lccc}
\toprule
\textbf{Prior Kind} & \textbf{100} & \textbf{500} & \textbf{1000} \\
\midrule
Expert     & $0.0765 \pm 0.0018$ & $0.0756 \pm 0.0018$ & $0.0752 \pm 0.0018$ \\
Advanced   & $0.0868 \pm 0.0019$ & $0.0878 \pm 0.0019$ & $0.0879 \pm 0.0019$ \\
Local & $0.0860 \pm 0.0020$ & $0.0862 \pm 0.0020$ & $0.0866 \pm 0.0020$ \\
Deceptive  & $0.1289 \pm 0.0031$ & $0.1295 \pm 0.0031$ & $0.1263 \pm 0.0030$ \\
\bottomrule
\end{tabular}
\end{table}

\begin{table}[h]
\centering
\caption{Sensitivity analysis of the prior rejection sample budget on PD1 (30 seeds). Values are mean regret $\pm$ standard error.}
\label{tab:ablation_budget_pd1}
\begin{tabular}{lccc}
\toprule
\textbf{Prior Kind} & \textbf{100} & \textbf{500} & \textbf{1000} \\
\midrule
Expert     & $0.1181 \pm 0.0048$ & $0.1174 \pm 0.0048$ & $0.1188 \pm 0.0049$ \\
Advanced   & $0.1272 \pm 0.0044$ & $0.1273 \pm 0.0043$ & $0.1232 \pm 0.0042$ \\
Local & $0.1349 \pm 0.0049$ & $0.1349 \pm 0.0048$ & $0.1337 \pm 0.0047$ \\
Deceptive  & $0.1528 \pm 0.0052$ & $0.1526 \pm 0.0052$ & $0.1528 \pm 0.0052$ \\
\bottomrule
\end{tabular}
\end{table}

\newpage
\section{Declaration of LLM Usage}
\label{app:llm-usage}
Throughout this submission, we made limited use of Large Language Models (LLMs) in the following ways:  
\begin{itemize}
    \item Code generation from specific instructions, primarily for producing plots and tables.  
    \item Writing support, including translation and alternative phrasings.  
    \item Assistance in locating related research.  
\end{itemize}  
All conceptual contributions, methodological developments, experimental designs, and analyses were carried out solely by the authors.  

\end{document}